\documentclass[12pt]{dualexamreport}
\usepackage{amsmath}
\usepackage{float}
\usepackage{pifont}
\usepackage{url}
\usepackage{tikz}
\usetikzlibrary{arrows.meta,positioning,fit,calc}
\AtBeginDocument{%
  \pretocmd{\section}{\FloatBarrier}{}{}%

}

\colorlet{agentblue}{metablue}
\colorlet{reportink}{metafg}
\colorlet{reportpanel}{metabg}
\definecolor{enginegreen}{HTML}{2A9D8F}
\definecolor{strictorange}{HTML}{E76F51}
\definecolor{light}{gray}{0.9}
\definecolor{leaderphysical}{HTML}{3568A8}
\definecolor{leaderdigital}{HTML}{7257B8}
\definecolor{leadersocial}{HTML}{C96B19}
\definecolor{leaderscience}{HTML}{238678}
\newcommand{\shortbench}{Dual Exam}
\newcolumntype{L}[1]{>{\raggedright\arraybackslash}p{#1}}
\newcolumntype{C}[1]{>{\centering\arraybackslash}p{#1}}
\newif\iftechnicalreport
\technicalreporttrue
\newcommand{\equalcontributionmark}{%
  \textsuperscript{\textcolor{agentblue}{*}}}
\newcommand{\correspondingmark}{%
  \textsuperscript{\textcolor{agentblue}{\textdagger}}}
\newcommand{\authorentry}[1]{%
  \mbox{#1}}
\newcommand{\authorgroupheading}[1]{%
  {\bfseries\textcolor{agentblue}{#1:}}\enspace}

\newcommand{\reportauthors}{%
  \begin{minipage}{0.97\textwidth}
  \raggedright
  \normalfont\sffamily\footnotesize\color{reportink}
  \linespread{1.10}\selectfont
  \authorgroupheading{Organizing Team}
  \authorentry{Zhefan~Rao\equalcontributionmark},
  \authorentry{Meng~Chu\equalcontributionmark},
  \authorentry{Xuan~Billy~Zhang\equalcontributionmark},
  \authorentry{Jingfan~Chen\equalcontributionmark},
  \authorentry{Guohao~Cai\equalcontributionmark},
  \authorentry{Runyi~Li\equalcontributionmark},
\authorentry{Shiyi~Du\equalcontributionmark},
\authorentry{Xu~Huang\equalcontributionmark},
  \authorentry{Ning~Lu},
  \authorentry{Trusure~Chen},
  \authorentry{Feng~Nie},
  \authorentry{Zhenhao~Liang},
  \authorentry{Haodong~Chen},
  \authorentry{Hanwen~Wang},
  \authorentry{Wenjie~Shu},
  \authorentry{Yajing~Bai},
  \authorentry{Xianfeng~Wu},
  \authorentry{Xuanhua~He},
  \authorentry{An~Wang},
  \authorentry{Shuhan~Guo},
  \authorentry{Ziyue~Ma}
  \par\vspace{0.65em}
  \authorgroupheading{Contributors}
  \authorentry{Ruichuan~An},
  \authorentry{Long~Bai},
  \authorentry{Miaosen~Chai},
  \authorentry{Wenhao~Chai},
  \authorentry{Jingye~Chen},
  \authorentry{Qi~Chen},
  \authorentry{Xuhang~Chen},
  \authorentry{Wei~Chow},
  \authorentry{Minghao~Fu},
  \authorentry{Shenyuan~Gao},
  \authorentry{Yunhao~Gou},
  \authorentry{Haokun~Gui},
  \authorentry{Dong~Huang},
  \authorentry{Wei~Huang},
  \authorentry{Zi~Cheng~Huang},
  \authorentry{Ziyan~Jiang},
  \authorentry{Haibo~Jin},
  \authorentry{Lingdong~Kong},
  \authorentry{Leung~Chun~Ho},
  \authorentry{Tianqin~Li},
  \authorentry{Yanheng~Li},
  \authorentry{Jiayuan~Liu},
  \authorentry{Jiyang~Liu},
  \authorentry{Mingyu~Liu},
  \authorentry{Shaoteng~Liu},
  \authorentry{Xiaocheng~Lu},
  \authorentry{Yawen~Luo},
  \authorentry{Zeyu~Qin},
  \authorentry{Tiancheng~Shen},
  \authorentry{Yueqi~Song},
  \authorentry{Zhaolong~Su},
  \authorentry{Hengkai~Tan},
  \authorentry{Tianhao~Tang},
  \authorentry{Yiran~Tao},
  \authorentry{Juncheng~Wang},
  \authorentry{Qinghe~Wang},
  \authorentry{Shangshang~Wang},
  \authorentry{Yingying~Wang},
  \authorentry{Fang~Wu},
  \authorentry{Tiange~Xiang},
  \authorentry{Qiushi~Yang},
  \authorentry{Weicai~Ye},
  \authorentry{Weichen~Yu},
  \authorentry{Jiarui~Zhang},
  \authorentry{Jize~Zhang},
  \authorentry{Liuzhou~Zhang},
  \authorentry{Renrui~Zhang},
  \authorentry{Sheng~Zhang},
  \authorentry{Wenhu~Zhang},
  \authorentry{Junru~Zhao},
  \authorentry{Lingting~Zhu},
  \authorentry{Xi~Zhu},
  \authorentry{Yuxin~Zuo}.
  \par\vspace{0.65em}
  \authorgroupheading{Advisory Committee}
  \authorentry{Yutong~Bai},
  \authorentry{Haoxuan~Che\correspondingmark},
  \authorentry{Qifeng~Chen},
  \authorentry{Long~Chen},
  \authorentry{Yang~Deng},
  \authorentry{Ziqi~Huang},
  \authorentry{Yeying~Jin},
  \authorentry{Weijia~Li},
  \authorentry{Kevin~Qinghong~Lin},
  \authorentry{Quande~Liu},
  \authorentry{Dimitris~Metaxas},
  \authorentry{See-Kiong~Ng},
  \authorentry{Hongliang~Ren},
  \authorentry{Hongru~Wang},
  \authorentry{Gerbrand~Ceder},
  \authorentry{Shujun~Wang},
  \authorentry{Ming-Hsuan~Yang},
  \authorentry{Alan~Yuille}.
  \par\vspace{0.45em}
  {\scriptsize
    \textcolor{agentblue}{*}~Equal contribution.\quad
    \textcolor{agentblue}{\textdagger}~Corresponding author.\par
    \smallskip
    Contributors and Advisory Committee members are ordered alphabetically by surname.\par
  }
  \end{minipage}%
}

\patchcmd{\reportauthors}
  {\normalfont\sffamily\footnotesize\color{reportink}}
  {\normalfont\rmfamily\normalsize\color{metafg}}{}{}
\renewcommand{\authorgroupheading}[1]{%
  {\sffamily\bfseries\color{black}#1:}\enspace}

\hypersetup{
  bookmarksdepth=2,
  pdftitle={Dual Exam: Controlled Evaluation of Multimodal Agents and World Models},
  pdfauthor={Dual Exam Consortium},
  pdfsubject={Technical report on dual-role evaluation across physical, digital, social, and scientific worlds}
}
\title{Abstract}
\patchcmd{\mymaketitle}{\abstractlist\par}{%
  {\centering
    \includegraphics[width=\linewidth]{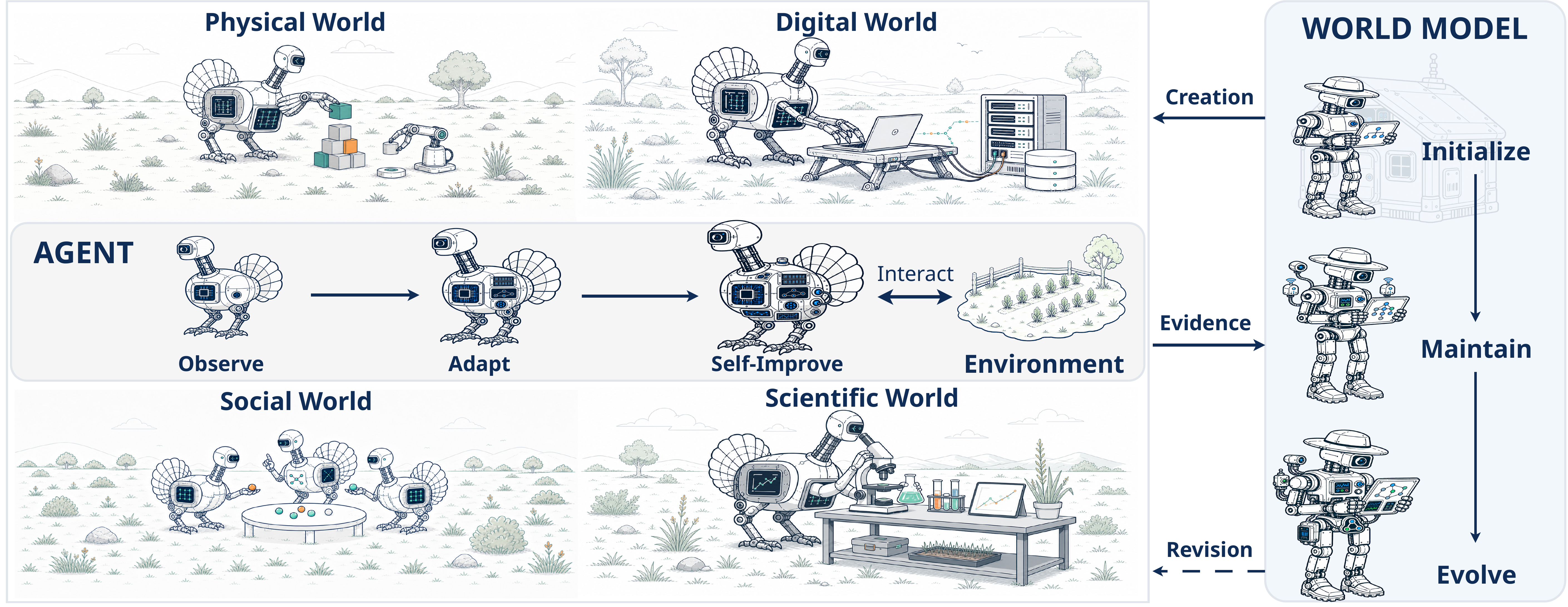}\par}%
  \vspace{0.15cm}%
  \abstractlist\par}{}{%
  \PackageError{report_metadata}{Could not insert abstract overview figure}{}}
\patchcmd{\mymaketitle}{\vskip 0.5cm}{\vskip 1cm}{}{%
  \PackageError{report_metadata}{Could not adjust abstract metadata spacing}{}}
\DeclareRobustCommand{\reportmetablogicon}{%
  \makebox[13bp][c]{\raisebox{-0.18\height}{\includegraphics[width=11bp,height=11bp,keepaspectratio]{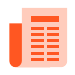}}}\hspace{0.4em}}
\DeclareRobustCommand{\reportmetacorrespondenceicon}{%
  \makebox[13bp][c]{\raisebox{-0.18\height}{\includegraphics[width=11bp,height=11bp,keepaspectratio]{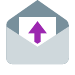}}}\hspace{0.4em}}

\metadata[\reportmetablogicon{}Blog]{\href{https://xgenlabs.ai/research/dual-exam/}{xgenlabs.ai/research/dual-exam/}}

\metadata[\reportmetacorrespondenceicon{}Correspondence]{Haoxuan Che \href{mailto:hche@connect.ust.hk}{(hche@connect.ust.hk)}}

\usepackage{xcolor}
\definecolor{darkgreen}{RGB}{0,100,0}

\defcitealias{team2026alayaworldfull}{AlayaWorld}

\abstract{
World Models (WMs) can support Multimodal Agents (MAs) through simulated experience, while Agent trajectories can inform World Model refinement. Evaluating this interaction requires separate evidence about action selection and state evolution. We introduce \shortbench{}, which tests Agents under fixed environment response mechanisms and World Models under fixed action interventions across physical, digital, social, and scientific domains. The Agent track compares 10 configurations on 1,088 items in 15 reporting categories. The World Model track contains 400 cases across eight scenarios; 15 configurations have full coverage and one covers six scenarios. GPT 6 Astra leads the case-weighted Agent Index at 31.3, and GPT-Image-2 leads the full-coverage WM Overall at 21.7. These scores use different tasks and metrics and are not directly comparable. A 150-case pilot of one coupled pairing records zero task successes and motivates component-level diagnosis. Results reveal uneven task performance and failures to preserve task-relevant state. Their value for predicting coupled success or sustained reciprocal learning remains to be established.
}

\begin{document}
\arrayrulecolor{black} 
  \begin{titlepage}
    \raggedright
    \vspace*{4cm}
    {\fontsize{60}{64}\selectfont\sffamily\bfseries
      \textcolor{metablue}{Dual Exam:}\par}
    {\fontsize{40}{44}\selectfont\sffamily
      \textcolor{black!60}{Controlled Evaluation of\\
      Multimodal Agents and\\World Models}\par}
    \vspace{2cm}
    \vfill
    \vspace{1cm}
    \begin{tikzpicture}[remember picture,overlay]
\node[inner sep=0pt] at ([xshift=108.0bp,yshift=-468bp]current page.north west) {\includegraphics[width=79.967bp,height=21.884bp]{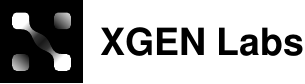}};
\node[inner sep=0pt] at ([xshift=207.0bp,yshift=-468bp]current page.north west) {\includegraphics[width=74.173bp,height=23.594bp]{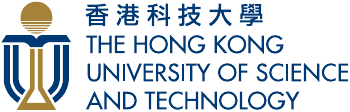}};
\node[inner sep=0pt] at ([xshift=306.0bp,yshift=-468bp]current page.north west) {\includegraphics[width=61.854bp,height=28.292bp]{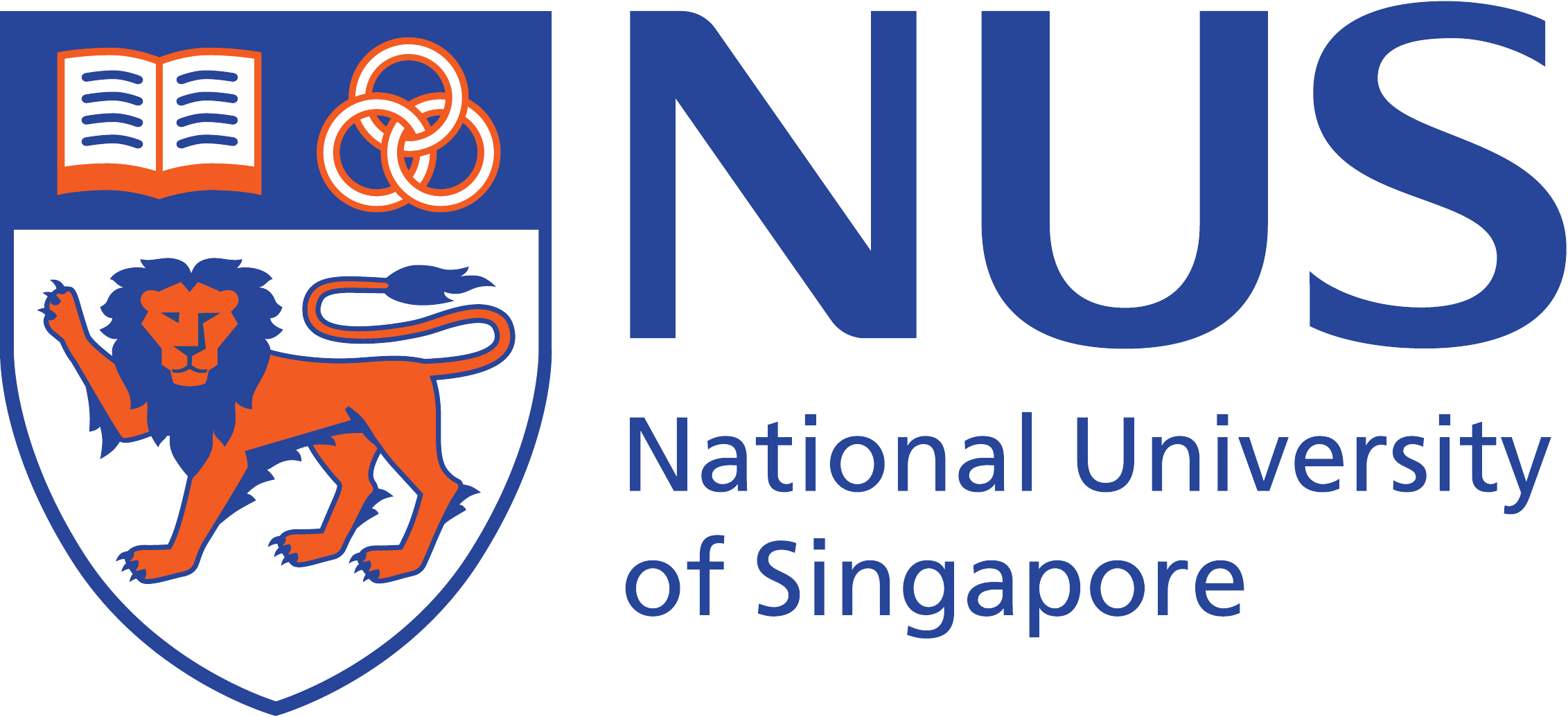}};
\node[inner sep=0pt] at ([xshift=405.0bp,yshift=-468bp]current page.north west) {\includegraphics[width=77.387bp,height=22.614bp]{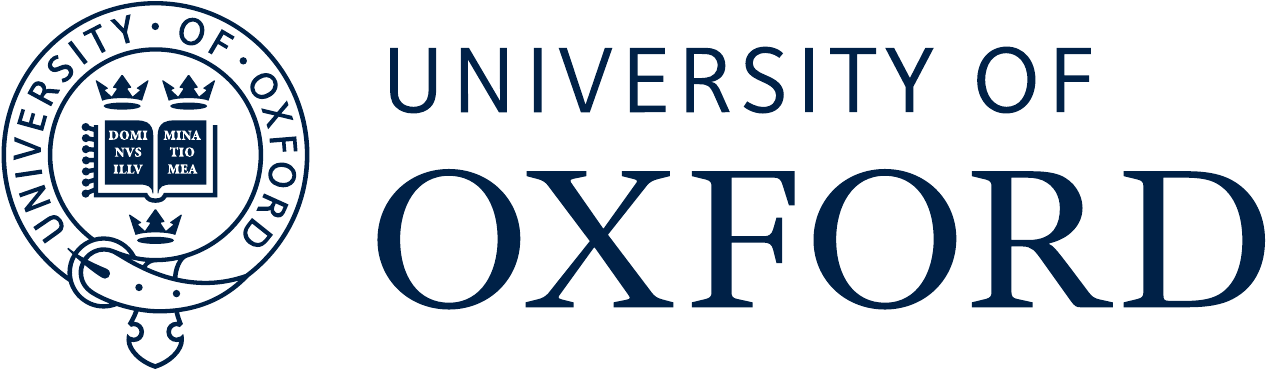}};
\node[inner sep=0pt] at ([xshift=504.0bp,yshift=-468bp]current page.north west) {\includegraphics[width=76.678bp,height=22.823bp]{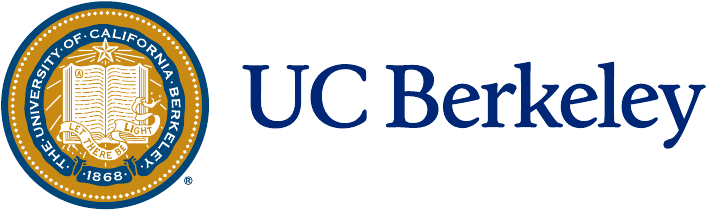}};
\node[inner sep=0pt] at ([xshift=108.0bp,yshift=-517bp]current page.north west) {\includegraphics[width=86.000bp,height=13.226bp]{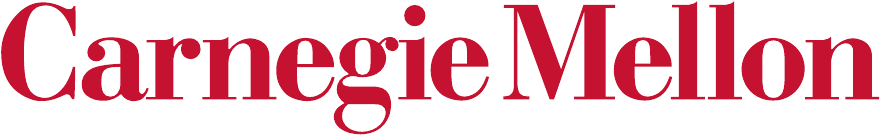}};
\node[inner sep=0pt] at ([xshift=207.0bp,yshift=-517bp]current page.north west) {\includegraphics[width=74.601bp,height=23.458bp]{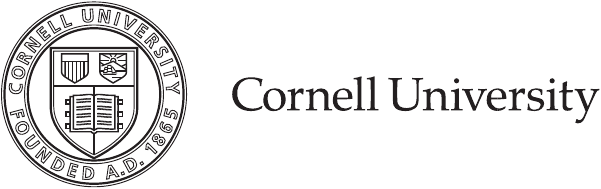}};
\node[inner sep=0pt] at ([xshift=306.0bp,yshift=-517bp]current page.north west) {\includegraphics[width=86.000bp,height=16.538bp]{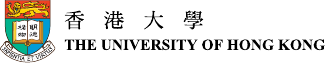}};
\node[inner sep=0pt] at ([xshift=405.0bp,yshift=-517bp]current page.north west) {\includegraphics[width=53.067bp,height=30.000bp]{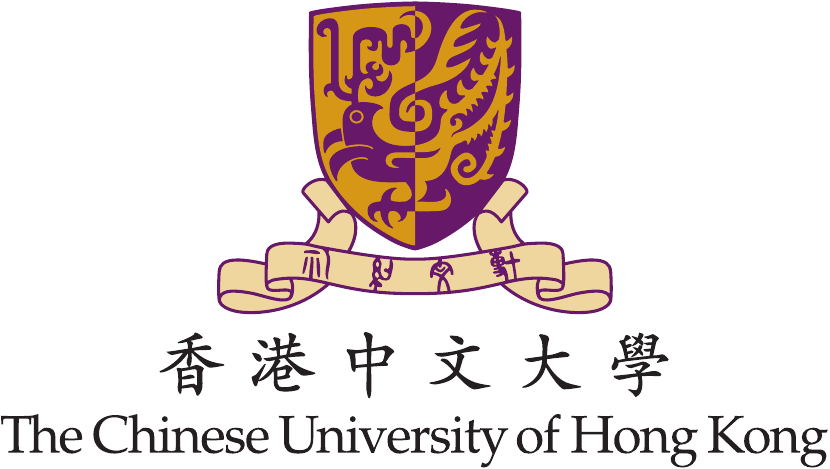}};
\node[inner sep=0pt] at ([xshift=504.0bp,yshift=-517bp]current page.north west) {\includegraphics[width=53.290bp,height=30.000bp]{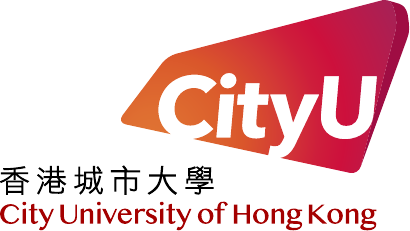}};
\node[inner sep=0pt] at ([xshift=108.0bp,yshift=-566bp]current page.north west) {\includegraphics[width=86.000bp,height=16.426bp]{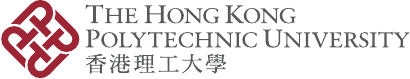}};
\node[inner sep=0pt] at ([xshift=207.0bp,yshift=-566bp]current page.north west) {\includegraphics[width=80.584bp,height=21.716bp]{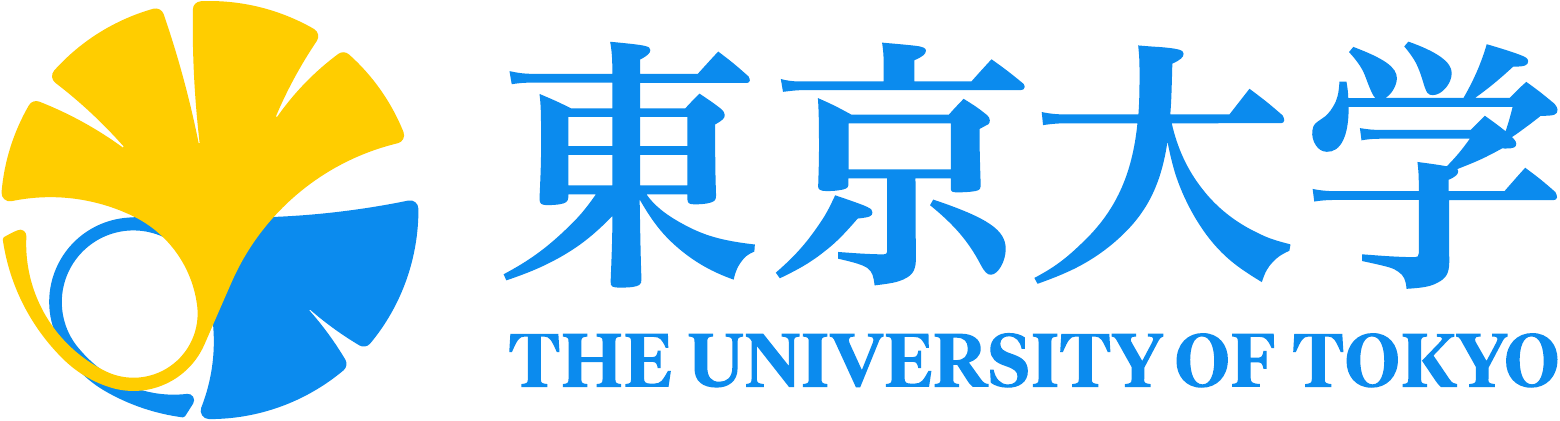}};
\node[inner sep=0pt] at ([xshift=306.0bp,yshift=-566bp]current page.north west) {\includegraphics[width=79.157bp,height=22.108bp]{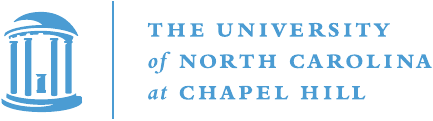}};
\node[inner sep=0pt] at ([xshift=405.0bp,yshift=-566bp]current page.north west) {\includegraphics[width=86.000bp,height=17.936bp]{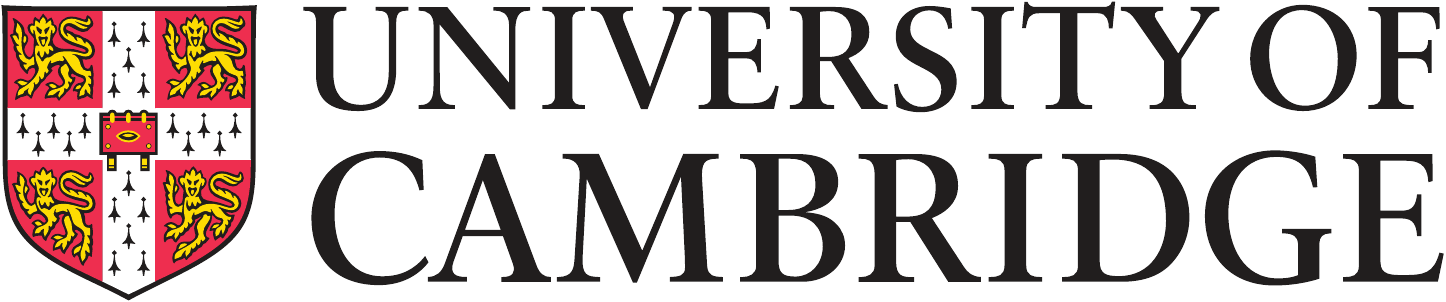}};
\node[inner sep=0pt] at ([xshift=504.0bp,yshift=-566bp]current page.north west) {\includegraphics[width=85.548bp,height=20.456bp]{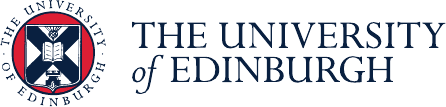}};
\node[inner sep=0pt] at ([xshift=108.0bp,yshift=-615bp]current page.north west) {\includegraphics[width=69.842bp,height=25.056bp]{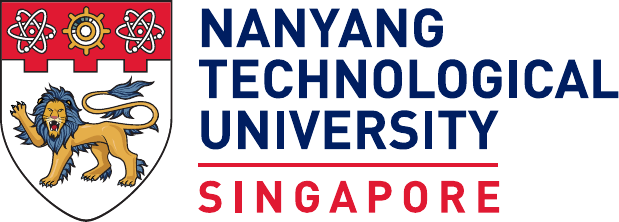}};
\node[inner sep=0pt] at ([xshift=207.0bp,yshift=-615bp]current page.north west) {\includegraphics[width=64.452bp,height=27.152bp]{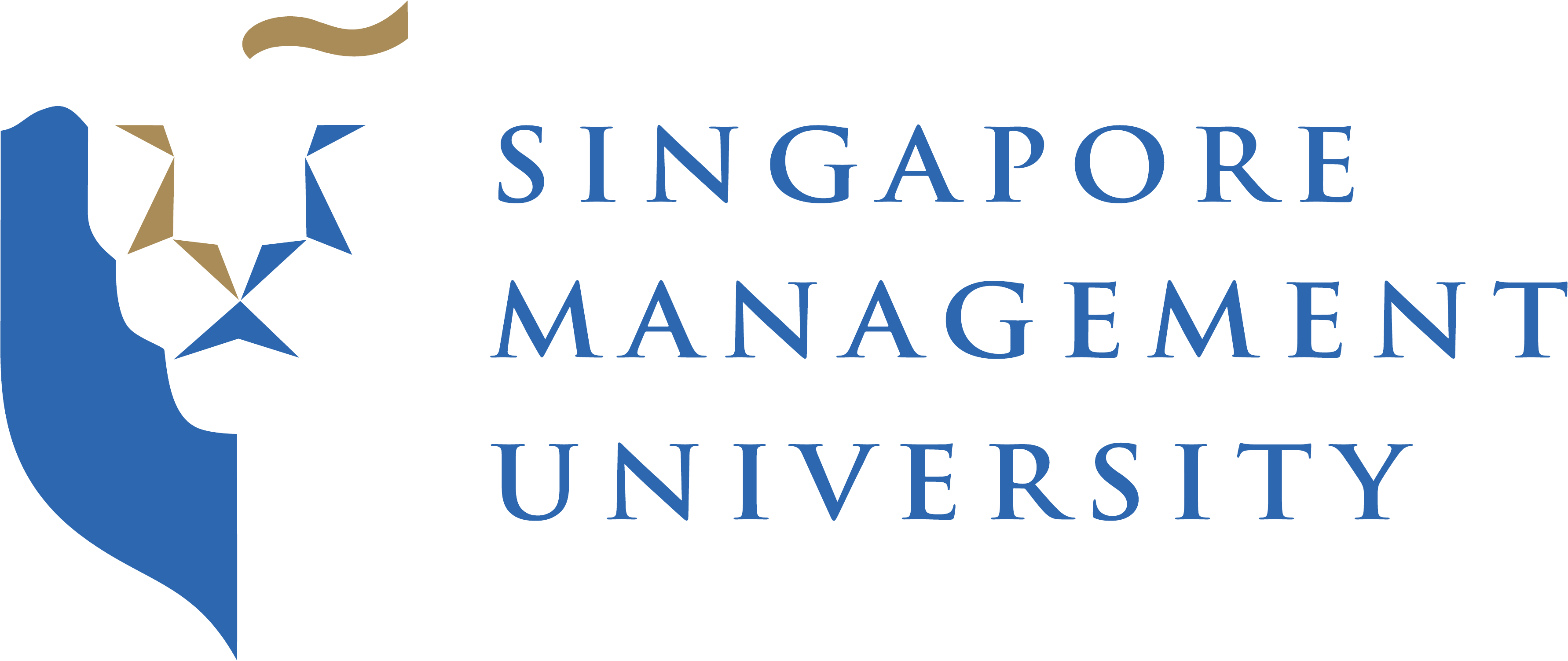}};
\node[inner sep=0pt] at ([xshift=306.0bp,yshift=-615bp]current page.north west) {\includegraphics[width=78.949bp,height=22.166bp]{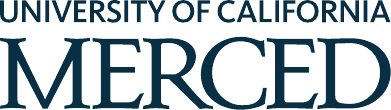}};
\node[inner sep=0pt] at ([xshift=405.0bp,yshift=-615bp]current page.north west) {\includegraphics[width=86.000bp,height=16.367bp]{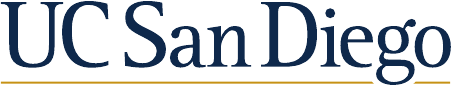}};
\node[inner sep=0pt] at ([xshift=504.0bp,yshift=-615bp]current page.north west) {\includegraphics[width=86.000bp,height=18.985bp]{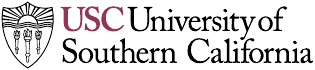}};
\node[inner sep=0pt] at ([xshift=108.0bp,yshift=-664bp]current page.north west) {\includegraphics[width=86.000bp,height=11.594bp]{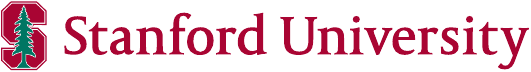}};
\node[inner sep=0pt] at ([xshift=207.0bp,yshift=-664bp]current page.north west) {\includegraphics[width=86.000bp,height=19.834bp]{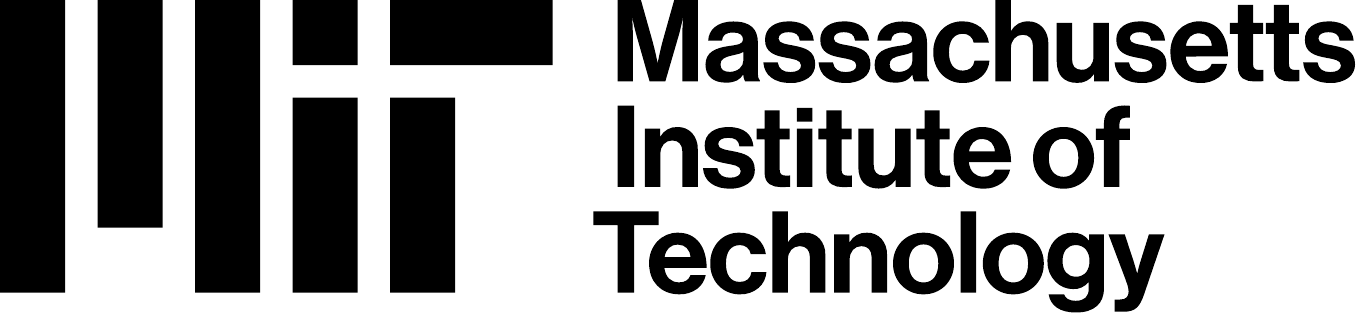}};
\node[inner sep=0pt] at ([xshift=306.0bp,yshift=-664bp]current page.north west) {\includegraphics[width=67bp,trim=282bp 254bp 197bp 180bp,clip]{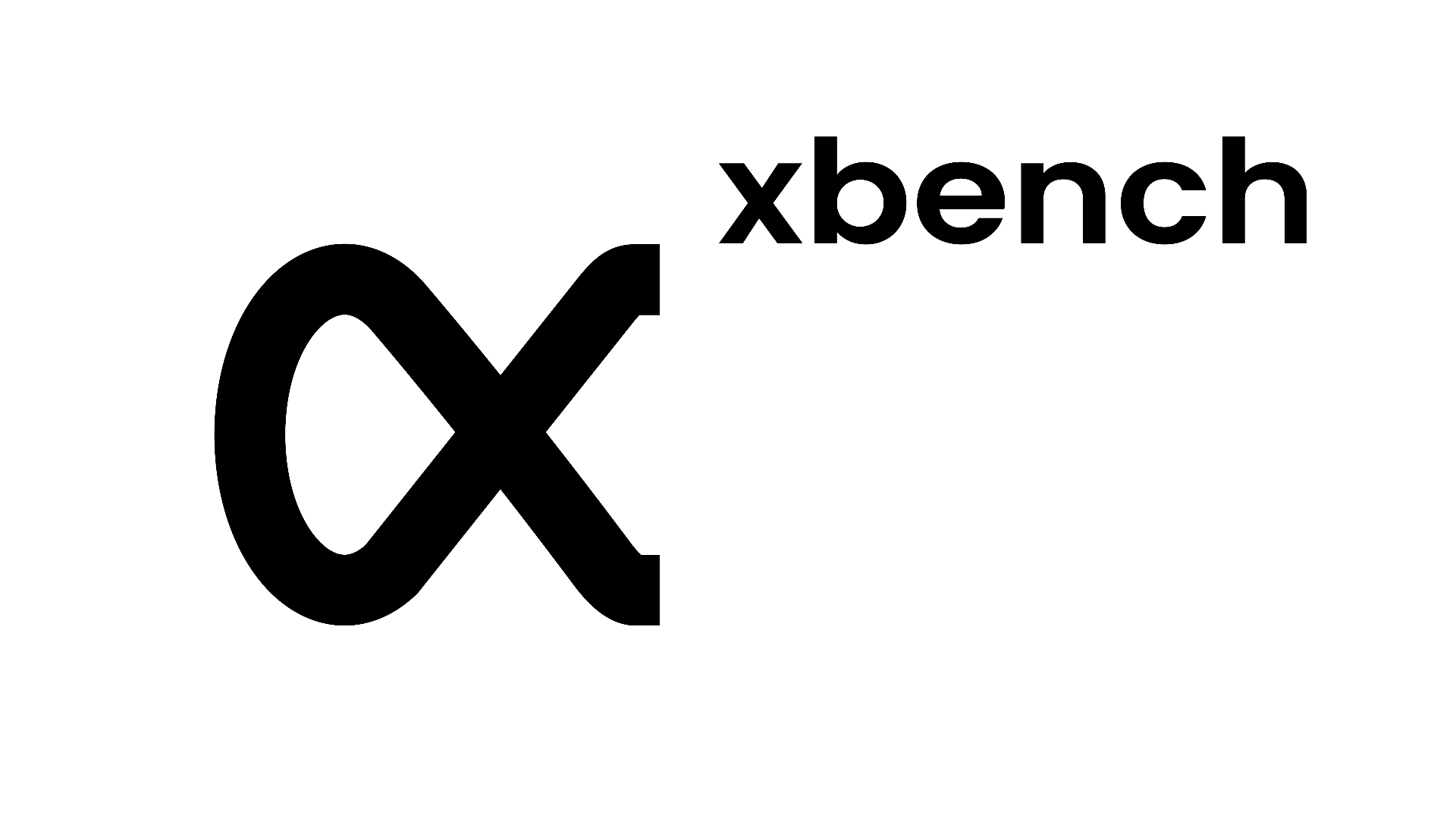}};
\node[inner sep=0pt] at ([xshift=405.0bp,yshift=-664bp]current page.north west) {\includegraphics[width=86.000bp,height=6.525bp]{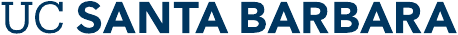}};
\node[inner sep=0pt] at ([xshift=504.0bp,yshift=-664bp]current page.north west) {\includegraphics[width=85.391bp,height=20.494bp]{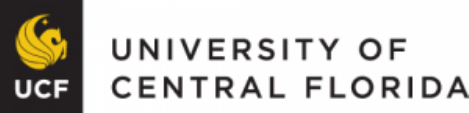}};
\node[inner sep=0pt] at ([xshift=157.5bp,yshift=-713bp]current page.north west) {\includegraphics[width=86.000bp,height=16.760bp]{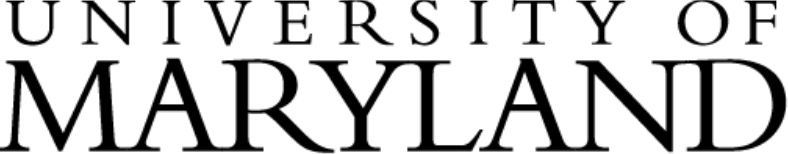}};
\node[inner sep=0pt] at ([xshift=256.5bp,yshift=-713bp]current page.north west) {\includegraphics[width=86.000bp,height=14.538bp]{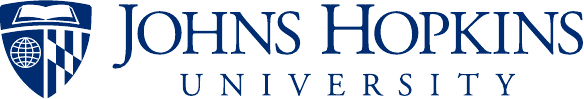}};
\node[inner sep=0pt] at ([xshift=355.5bp,yshift=-713bp]current page.north west) {\includegraphics[width=83.294bp,height=21.010bp]{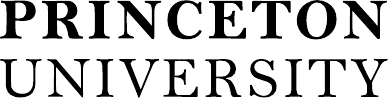}};
\node[inner sep=0pt] at ([xshift=454.5bp,yshift=-713bp]current page.north west) {\includegraphics[width=86.000bp,height=17.449bp]{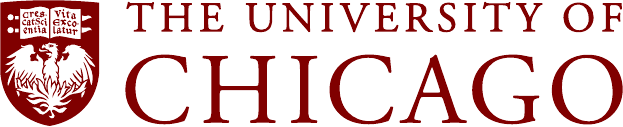}};

    \end{tikzpicture}
  \end{titlepage}

\maketitle
\thispagestyle{empty}
\clearpage
\begingroup
\small
\tableofcontents
\endgroup
\clearpage

\begin{figure}[h]
    \centering
    \includegraphics[width=\linewidth]{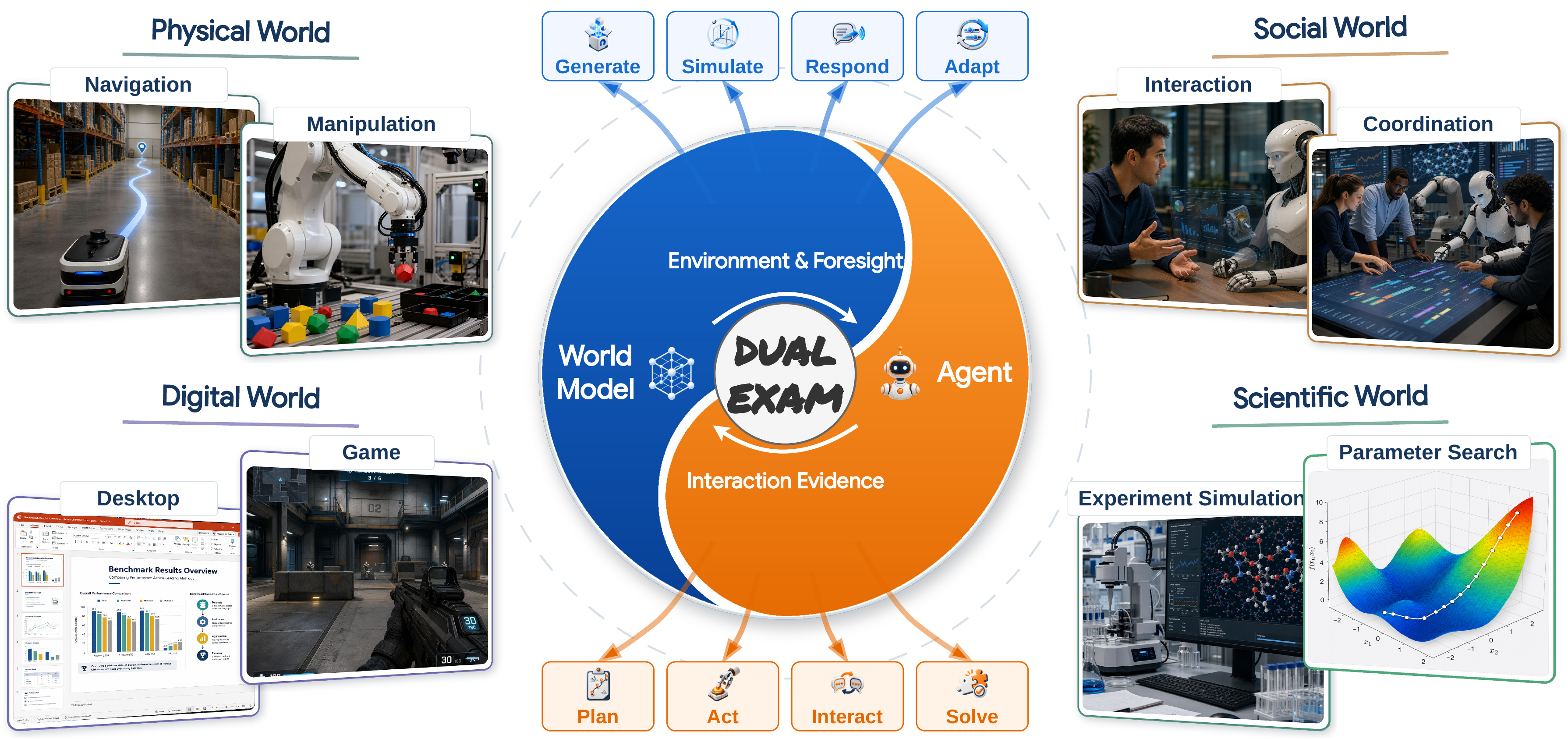}
    \caption{\textbf{Agent--World Model interaction and the scope of Dual Exam.} World Models supply simulated observations and predictions; Agent trajectories supply interaction evidence. Arrows indicate potential uses, not demonstrated joint-training gains. Task labels are illustrative; Table~\ref{tab:agent-task-taxonomy} defines the evaluated inventory.}
    \label{fig:dual-exam-overview}
\end{figure}

\section{Introduction}

World Models (WMs) support Multimodal Agents (MAs) through simulated experience and action-conditioned predictions~\citep{ha2018worldmodels,hafner2025dreamerv3,gu2025webdreamer}; Agent trajectories can in turn inform WM refinement~\citep{worldevolver,yu2026reinforcementworldmodel}. Recent systems couple these roles in specific settings~\citep{fang2025webevolver,guo2025genenv,liu2026comap}. Extending such interaction across physical, digital, social, and scientific domains requires both reliable action selection and faithful state evolution~\citep{chu2026agenticworldmodeling}.

Both components remain unreliable over extended interaction. Agents struggle with grounding, sequencing, and completion~\citep{xie2024osworld,chen2025worldprediction,sun2026agentslastexam}; WMs struggle with memory and action control~\citep{ye2026mind}. Visual fidelity alone does not guarantee downstream utility~\citep{shang2026worldarena}. Planning benefits depend on the simulated states and actions, and on the Agent's ability to use them~\citep{li2025wordtoworld,qian2026foresight}; model bias can also accumulate over rollouts~\citep{janner2019mbpo}. Because each component consumes the other's outputs, an unsuccessful coupled run cannot by itself distinguish decision errors from simulation errors.

Our preliminary study tests one MA--WM pairing on 150 cases across ten scenarios, with 0\% task success under the stated protocol (Section~\ref{sec:preliminary}). The trajectories reveal state discrepancies and limited responses, but do not identify which component caused each failure. This motivates \textbf{Dual Exam}: we evaluate MAs under a fixed environment response mechanism and WMs under fixed action interventions.

\begin{figure}[!t]
  \centering
  \includegraphics[width=\linewidth]{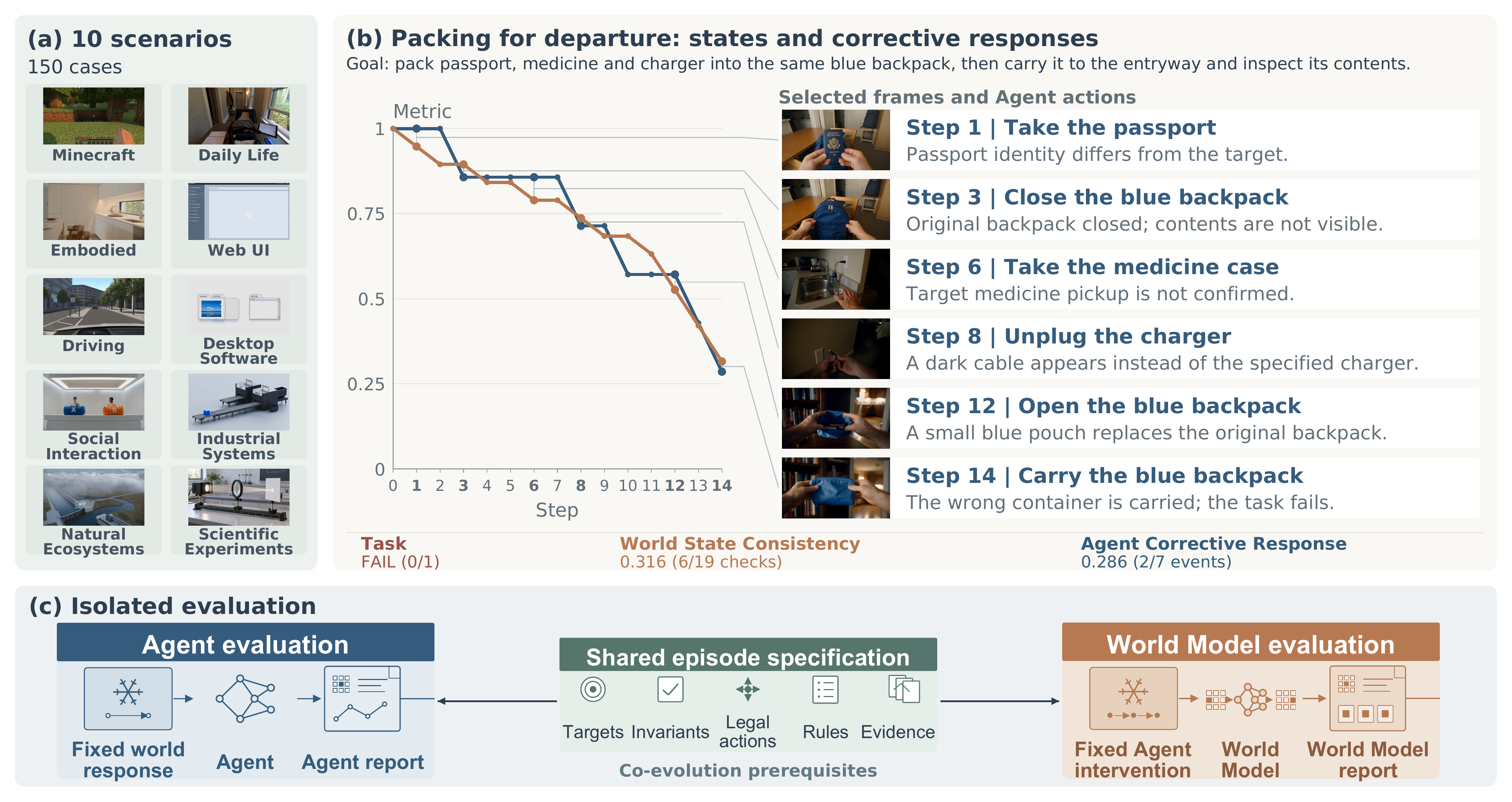}
  \caption{\textbf{From a failed packing rollout to controlled evaluation.} (a) Coverage of 150 cases across ten scenarios. (b) Six selected frames and diagnostic curves covering all 14 steps of a GPT-5.6 Sol--Seedance 2.5 rollout. Blue text gives MA commands; images show WM responses. Gray lines mark same-step correspondence, not causal attribution. World State Consistency (orange) and Agent Corrective Response (blue) use cumulative deductions from an assigned unit baseline with fixed full-trajectory denominators. Later packing and carrying use a different blue pouch, violating the same-backpack requirement. (c) Dual Exam holds the environment response mechanism fixed to evaluate MAs and fixes action interventions to evaluate WMs under explicit task specifications; the two tracks need not use matched episodes.}
  \label{fig:decoupled-evaluation}
\end{figure}

Dual Exam measures these components against explicit goals and state constraints. GPT 6 Astra leads the Agent Index at 31.3, while GPT-Image-2 leads the full-coverage WM Overall at 21.7. Agent case studies show that correct local judgments can still yield invalid routes or action sequences. WM cases distinguish responding to an instruction from preserving state: models may retain the required object count yet alter unrelated scene attributes. Category-specific performance and sensitivity to task weights limit interpretation of either scalar ranking. The results describe the evaluated model--harness configurations; execution settings are not fully matched, and component scores have not been validated as predictors of coupled success.

Our contributions are:
\begin{itemize}[leftmargin=*, itemsep=0.1em]
\item We introduce \textbf{Dual Exam}: separate evaluations of Agent action selection and World Model state evolution under controlled reference conditions.
\item We compare 10 Agent configurations on 1,088 items in 15 reporting categories across four domains, and organize the World Model evaluation into 400 cases across eight scenarios. The results provide complementary task and scenario profiles, with explicit aggregation rules and configuration-specific coverage.
\item A 150-case coupled pilot and source-linked case studies illustrate the distinction between local progress and verified task completion.
\end{itemize}

\begin{table}[!t]
  \centering
  \caption{\textbf{Coupled MA--WM evaluation across ten scenarios.} GPT-5.6 Sol interacts with Seedance 2.5 on 15 cases per scenario (150 total), with one attempt of at most 20 steps and no episode-level retries. Initialization and technical failures are excluded. Scenario rates are case-averaged; Overall averages the ten displayed scenario rates equally and is rounded to one decimal place. Section~\ref{sec:preliminary} defines the metrics.}
  \label{tab:preliminary-domain-results}
  \fontsize{7.5}{9}\selectfont
  \setlength{\tabcolsep}{1.5pt}
  \renewcommand{\arraystretch}{1.15}
  \newsavebox{\preliminarytablebox}
  \newsavebox{\preliminaryheaderbox}
  \newlength{\preliminaryoverallwidth}
  \settowidth{\preliminaryoverallwidth}{\textbf{Overall}}
  \addtolength{\preliminaryoverallwidth}{\tabcolsep}
  \sbox{\preliminaryheaderbox}{%
    \begin{tabular}{@{}l@{}}
      \toprule
      \makecell{Daily\\Life}\\
      \midrule
    \end{tabular}%
  }
  \begin{lrbox}{\preliminarytablebox}
  \begin{tabular*}{\linewidth}{@{\extracolsep{\fill}}l*{11}{c}@{}}
    \toprule
    Metric (\%) & Minecraft & \makecell{Daily\\Life} & \makecell{Embodied\\Tasks} & \makecell{Web\\UI} & Driving & \makecell{Desktop\\Software} & \makecell{Industrial\\Systems} & \makecell{Natural\\Ecosystems} & \makecell{Scientific\\Experiments} & \makecell{Social\\Interaction} & \textbf{Overall} \\
    \midrule
    \makecell[l]{World State\\Consistency} & 12.7 & 37.9 & 16.5 & 18.7 & 14.7 & 22.9 & 34.4 & 39.5 & 29.5 & 38.9 & 26.6 \\
    \addlinespace[2pt]
    \makecell[l]{Agent Corrective\\Response} & 0.0 & 13.3 & 16.7 & 6.7 & 0.0 & 0.0 & 20.0 & 10.0 & 0.0 & 25.0 & 9.2 \\
    \addlinespace[2pt]
    Task success & 0.0 & 0.0 & 0.0 & 0.0 & 0.0 & 0.0 & 0.0 & 0.0 & 0.0 & 0.0 & 0.0 \\
    \bottomrule
  \end{tabular*}
  \end{lrbox}
  \noindent\makebox[0pt][l]{%
    \hspace*{\dimexpr\linewidth-\preliminaryoverallwidth\relax}%
    {\color{black!6}\rule[-\dp\preliminarytablebox]{\preliminaryoverallwidth}{\dimexpr\ht\preliminarytablebox+\dp\preliminarytablebox-\ht\preliminaryheaderbox-\dp\preliminaryheaderbox+\belowrulesep\relax}}%
  }\usebox{\preliminarytablebox}
\end{table}

\raggedbottom
\section{Dual Exam}
\label{sec:dual-exam}
\label{sec:benchmark}
\label{sec:schema}
\label{sec:protocol}

Dual Exam evaluates Agents against reference environment responses and World Models under fixed action interventions. Both tracks verify recorded evidence against explicit task requirements, but use separate cases, metrics, and interaction protocols. The inventory comprises 1,088 Agent evaluation items and 400 World Model cases across four domains.

\begin{figure}[!tbp]
  \centering
  \includegraphics[width=\linewidth]{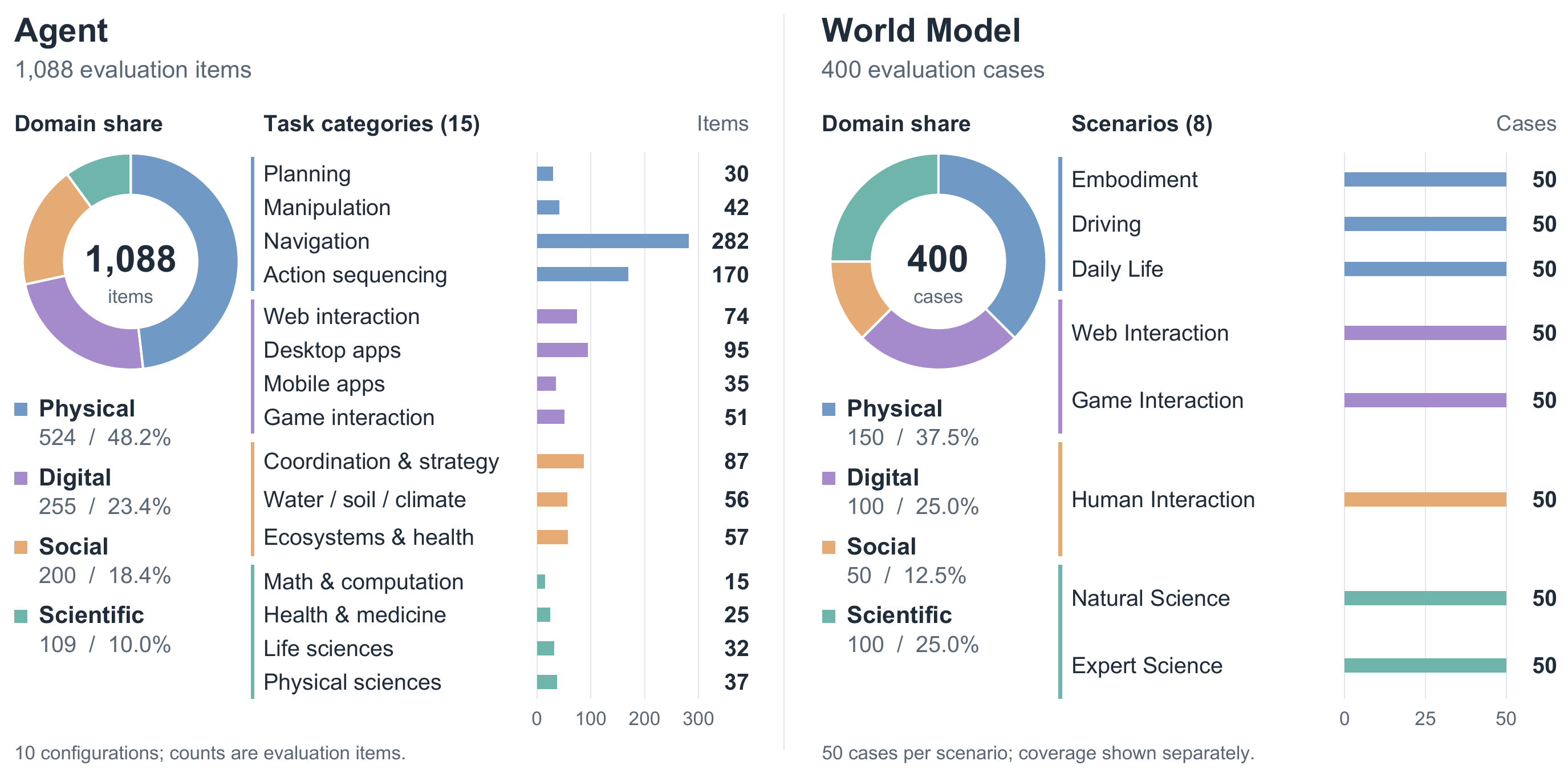}
  \caption{\textbf{Evaluation inventory across four domains.} Rings show domain shares; bars show counts within each track, with separate horizontal scales. Agent labels are shortened reporting categories, not independent abilities; Table~\ref{tab:agent-task-taxonomy} gives their full names. The 1,088 items include ten derived ToolChoice checkpoints. World Model evaluation uses 50 cases in each of eight scenarios: 15 configurations cover all 400 cases; Seedance 2.5 covers six scenarios (300 cases), excluding Daily Life and Human Interaction. Counts are not multiplied by model configurations.}
  \label{fig:case-distribution}
\end{figure}

Figure~\ref{fig:case-distribution} summarizes the inventory by domain and reporting category.

\subsection{Preliminary Study: Motivation for Dual Exam}
\label{sec:preliminary}

We begin with a pilot study of coupled interaction: GPT-5.6 Sol selects actions from the current observation, and Seedance 2.5 generates the next observation. The study covers 150 cases across ten scenarios, with 15 cases per scenario. Each case has one attempt of at most 20 interaction steps, with no episode-level retries. Initialization and technical failures are excluded from the 150 cases and the task-success denominator; the excluded count and complete runtime version and date records are unavailable. Table~\ref{tab:preliminary-domain-results} reports a retrospective trajectory audit.

\emph{World State Consistency} measures preservation of task-relevant identity and state and realization of requested action effects, averaging supported checks over applicable checks, then over steps and cases. \emph{Agent Corrective Response} measures the fraction of eligible visible discrepancies receiving targeted verification, repositioning, or correction within two actions; a response need not repair the discrepancy. \emph{Task Success} requires satisfying the full goal and all constraints. Scenario rates are case-averaged. Overall averages the ten displayed scenario rates equally, giving 26.6\%, 9.2\%, and 0.0\%, respectively. These rates describe this pairing and protocol, not either component's isolated ability.

The packing rollout shows how continued action can diverge from the task requirements (Figure~\ref{fig:decoupled-evaluation}(b)). The Agent must pack a passport, medicine, and a charger into the same blue backpack, then carry it to the entryway for inspection. Commands continue despite unconfirmed item acquisition, and later packing and carrying use a smaller blue pouch. This container substitution violates the task regardless of whether occluded items are present. Six selected frames accompany curves over all 14 steps, ending at 0.316 (6/19 state checks) and 0.286 (2/7 response events). Both responses are navigation retries; none of the four identity discrepancies receives a targeted response.

The curves start at an assigned score of 1 and deduct $1/19$ per failed state check and $1/7$ per eligible discrepancy left unanswered after two actions. Denominators are fixed from the complete trajectory; declining curves therefore do not measure capability degradation. This illustrative curve pools state checks within one trajectory; Table~\ref{tab:preliminary-domain-results} instead averages over steps and cases. The state audit records six passes and 13 failures, excluding 22 unknown and one inapplicable check. This single automated audit used a retrospective rubric and has not been independently adjudicated.

Coupled failure leaves a diagnostic ambiguity: subsequent actions depend on generated observations, so the outcome alone cannot separate Agent errors from World Model errors or their interaction. Dual Exam therefore holds the reference environment response mechanism fixed to test MAs and fixes action interventions to test WMs against specified goals and constraints (Figure~\ref{fig:decoupled-evaluation}(c)). These tests measure role-specific performance rather than retrospectively assigning causes to the pilot failures. Table~\ref{tab:benchmark_comparison} compares this scope with existing benchmarks.


\providecommand{\cmark}{\ding{51}}
\providecommand{\xmark}{\ding{55}}

\begin{table*}[t]
\centering

\caption{
\textbf{Comparison of benchmark scope.}
Representative Agent and World Model benchmarks are compared
along interaction, diagnostic, and evaluation-scope dimensions.
}
\label{tab:benchmark_comparison}

\renewcommand{\arraystretch}{1.10}
\scriptsize
\setlength{\tabcolsep}{4pt}

\begin{tabular*}{\textwidth}{
@{\extracolsep{\fill}}
lccccccr
@{}
}
\toprule

&
\multicolumn{2}{c}{\textbf{Interaction}}
&
\multicolumn{2}{c}{\textbf{Diagnosis}}
&
\multicolumn{2}{c}{\textbf{Scope}}
& \\[-1mm]

\cmidrule(lr){2-3}
\cmidrule(lr){4-5}
\cmidrule(lr){6-7}

\textbf{Benchmark}
& \makecell{\textbf{Multimodal}\\\textbf{Input}}
& \makecell{\textbf{Interactive}\\\textbf{Rollout}}
& \makecell{\textbf{Environment-}\\\textbf{grounded}}
& \makecell{\textbf{Component}\\\textbf{Isolation}}
& \makecell{\textbf{Persistent}\\\textbf{State}}
& \makecell{\textbf{Four-domain}\\\textbf{Coverage}}
& \makecell{\textbf{\# Eval.}\\\textbf{Units}}
\\

\midrule


\rowcolor{black!5}
\multicolumn{8}{c}{\textbf{Agent Benchmarks}} \\

GAIA~\citep{mialon2024gaia}
& \cmark & \xmark
& \xmark & \xmark
& \xmark & \xmark
& 466 \\

AgentBench~\citep{liu2024agentbench}
& \xmark & \cmark
& \cmark & \xmark
& \xmark & \xmark
& 1,141 \\

WebArena~\citep{zhou2024webarena}
& \xmark & \cmark
& \cmark & \xmark
& \xmark & \xmark
& 812 \\

VisualWebArena~\citep{koh2024visualwebarena}
& \cmark & \cmark
& \cmark & \xmark
& \xmark & \xmark
& 910 \\

OSWorld~2.0~\citep{yuan2026osworld2}
& \cmark & \cmark
& \cmark & \xmark
& \xmark & \xmark
& 108 \\

Agents' Last Exam~\citep{sun2026agentslastexam}
& \cmark & \cmark
& \cmark & \xmark
& \xmark & \xmark
& 1,500+ \\

\rowcolor{black!3}
\textbf{Dual Exam (Agent)}
& \cmark & \cmark
& \cmark & \cmark
& \cmark & \cmark
& \textbf{1,088} \\


\midrule

\rowcolor{black!5}
\multicolumn{8}{c}{\textbf{World Model Benchmarks}} \\

VBench~\citep{huang2024vbench}
& \xmark & \xmark
& \xmark & \xmark
& \xmark & \xmark
& 946 \\

WorldScore~\citep{duan2025worldscore}
& \cmark & \cmark
& \xmark & \xmark
& \xmark & \xmark
& 3,000 \\

WorldModelBench~\citep{li2025worldmodelbench}
& \cmark & \xmark
& \xmark & \xmark
& \xmark & \xmark
& 350 \\

MIND~\citep{ye2026mind}
& \cmark & \cmark
& \xmark & \xmark
& \cmark & \xmark
& 250 \\

WorldArena~\citep{shang2026worldarena}
& \cmark & \cmark
& \cmark & \xmark
& \xmark & \xmark
& 500\\

Omni-WorldBench~\citep{wu2026omniworldbench}
& \cmark & \cmark
& \xmark & \xmark
& \cmark & \xmark
& 1,068 \\

WorldMark~\citep{xu2026worldmark}
& \cmark & \cmark
& \xmark & \xmark
& \cmark & \xmark
& 500 \\

WBench~\citep{ying2026wbench}
& \cmark & \cmark
& \xmark & \xmark
& \cmark & \xmark
& 289 \\

WorldExam~\citep{yang2026worldexam}
& \cmark & \cmark
& \xmark & \xmark
& \cmark & \xmark
& 1,474 \\

PlayWorld~\citep{ding2026playworld}
& \cmark & \cmark
& \xmark & \xmark
& \cmark & \xmark
& 171 \\

\rowcolor{black!3}
\textbf{Dual Exam (World Model)}
& \cmark & \cmark
& \cmark & \cmark
& \cmark & \cmark
& \textbf{400} \\

\bottomrule
\end{tabular*}

\vspace{1mm}

\begin{minipage}{\textwidth}
\scriptsize
\textit{Notation.}
\cmark{} / \xmark{} indicate explicit / no explicit coverage,
not model capability; coverage need not span all tasks.
\textbf{Multimodal Input}: visual inputs paired with language or
action/control inputs, initially or during interaction.
\textbf{Interactive Rollout}: actions or controls affect subsequent
observations or states, using predefined or adaptive sequences.
\textbf{Environment-grounded}: evaluation against execution outcomes
or independently verifiable environment states.
\textbf{Component Isolation}: paired Agent and World Model evaluation tracks, using reference environment responses for Agents and fixed action interventions for World Models. This denotes dual-track coverage, not an absence of controls in other benchmarks or causal attribution in coupled runs.
\textbf{Persistent State}: dedicated evaluation of state preservation
or evolution across revisits, occlusion, or interaction turns,
beyond generic temporal video quality.
\textbf{Four-domain Coverage}: coverage of physical, digital, social,
and scientific domains.
\textbf{\# Eval. Units}: benchmark-specific counts of questions,
tasks, prompts, videos, or scenarios, not directly comparable.

\end{minipage}

\vspace{-2mm}
\end{table*}

\paragraph{Research questions.}
We ask (RQ1) how Agent task performance varies under a fixed environment response mechanism, and (RQ2) whether WMs realize prescribed actions while preserving task-relevant state. The four domains and their reporting categories organize the inventory; they are not independent dimensions of intelligence. Table~\ref{tab:agent-task-taxonomy} lists the task categories.

\subsection{Dual Examination Framework}
\label{sec:dual-design}
The Agent selects actions from its observation history and is evaluated against a fixed reference environment response mechanism. The World Model instead receives a predefined action sequence and generates the resulting observations. Each track specifies an initial state, permitted actions, required changes, protected state, and evidence rules. Reference mechanisms need not expose their underlying state to the Agent and are not assumed error-free (Section~\ref{sec:comparison-limits}).

These controls support separate assessments of action selection and state evolution. They do not imply matched episodes, interchangeable scores, or causal attribution of coupled failures.


\subsection{Agent Evaluation}
\label{sec:agent-exam}

\begin{figure}[t]
  \centering
  \includegraphics[width=\linewidth]{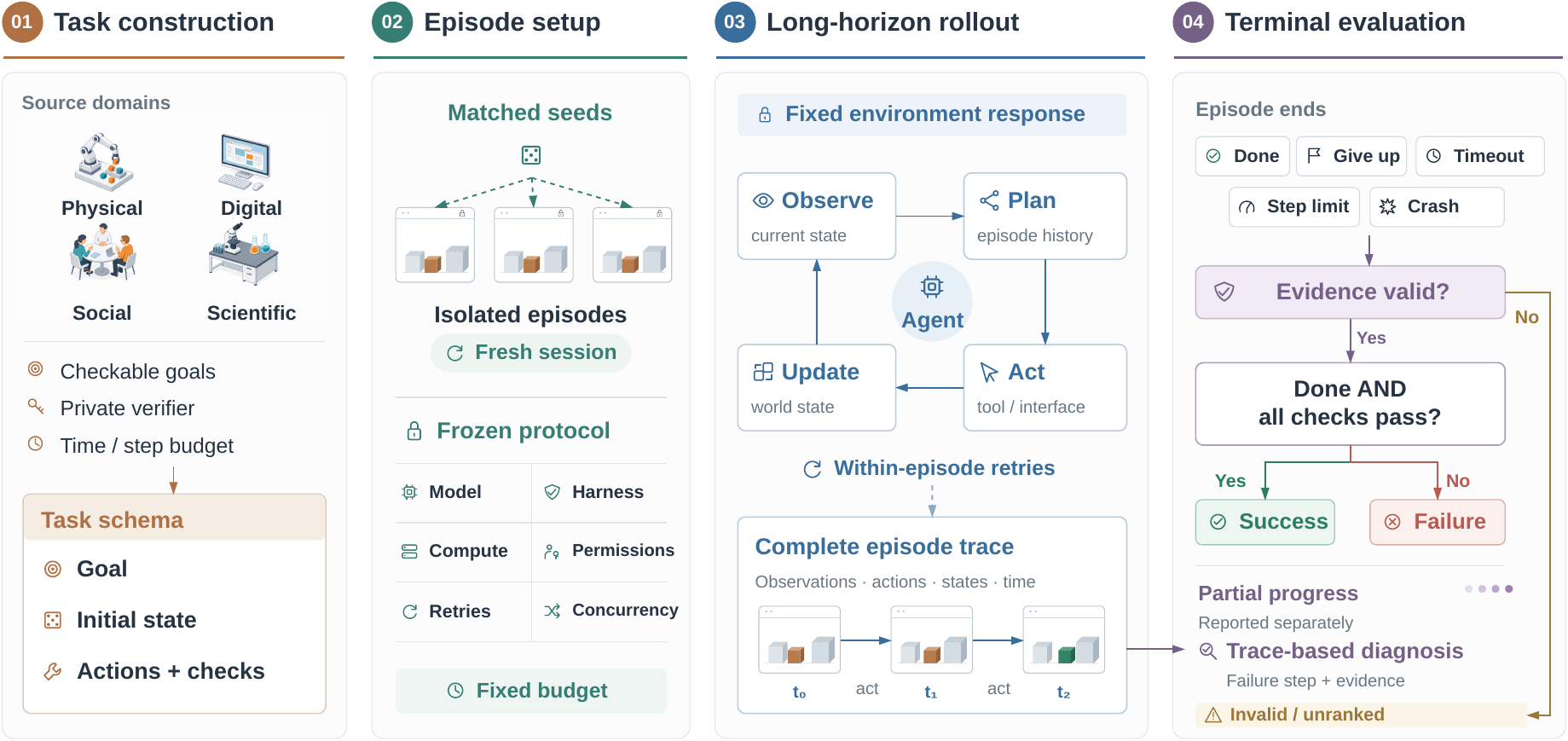}
  \caption{
  \textbf{Agent task construction and evaluation.}
  Tasks specify goals, initial states, allowed actions, rules, and required
  checks. Candidate Agents interact with controlled environments under
  predefined settings and budgets, with observations, actions, states,
  and artifacts recorded throughout. The diagram illustrates strict-success evaluation and the intended controlled protocol. The leaderboard uses task-specific scoring, including graded outcomes; partial progress is not universally excluded from scores. Historical configuration differences and evidence limitations are documented in the evaluation notes.
  }
  \label{fig:agent-task-pipeline}
\end{figure}

The Agent track tests task completion through interaction with a reference environment. Figure~\ref{fig:agent-task-pipeline} summarizes benchmark construction and evaluation.


\noindent\textbf{Benchmark construction.}
The Agent inventory contains 1,088 items: 524 Physical, 255 Digital, 200 Social, and 109 Scientific (Table~\ref{tab:agent-task-taxonomy}). Scientific items comprise 63 tasks from Agents' Last Exam (ALE), 30 SWE-bench code-repair tasks~\citep{arxiv231006770}, ten ToolChoice checkpoints derived from repair workflows, and six EdgeBench optimization tasks~\citep{zhu2026edgebench}. ToolChoice inherits the parent task's subject category. These 99 full workflows and ten checkpoints are not necessarily independent observations. Table~\ref{tab:scientific-capability-counts} provides source-by-category counts.

\noindent\textbf{Evaluation and aggregation.}
Task-specific scores are reported on a 0--100 scale. A score can represent strict completion or graded performance according to the corresponding task's evaluation rule; the values should not all be interpreted as binary success rates. For model configuration $m$, category score $s_{m,c}$, and category count $n_c$, the reported Index is
\[
I_m=\frac{\sum_c n_c s_{m,c}}{1088}.
\]
Physical, Digital, Social, and Scientific tasks contribute 48.16\%, 23.44\%, 18.38\%, and 10.02\% of Index, respectively. These weights reflect inventory counts, not domain importance.

For each Scientific category, we multiply the original ALE mean by its item count, add 100 per strict SWE or ToolChoice pass and the graded Edge scores, then divide by the total category count. ToolChoice partial credit is not used. Borden and DABIC scores are clipped to 0--100 by our convention, not an official EdgeBench rule (Section~\ref{app:edge-aggregation}). Original category means retain their published one-decimal precision; derived scores are rounded only after aggregation. Section~\ref{app:agent-weight-sensitivity} compares equal-domain weighting.

\subsection{World Model Evaluation}
\label{sec:world-model-exam}

\begin{figure}[t]
  \centering
  \includegraphics[width=\linewidth]{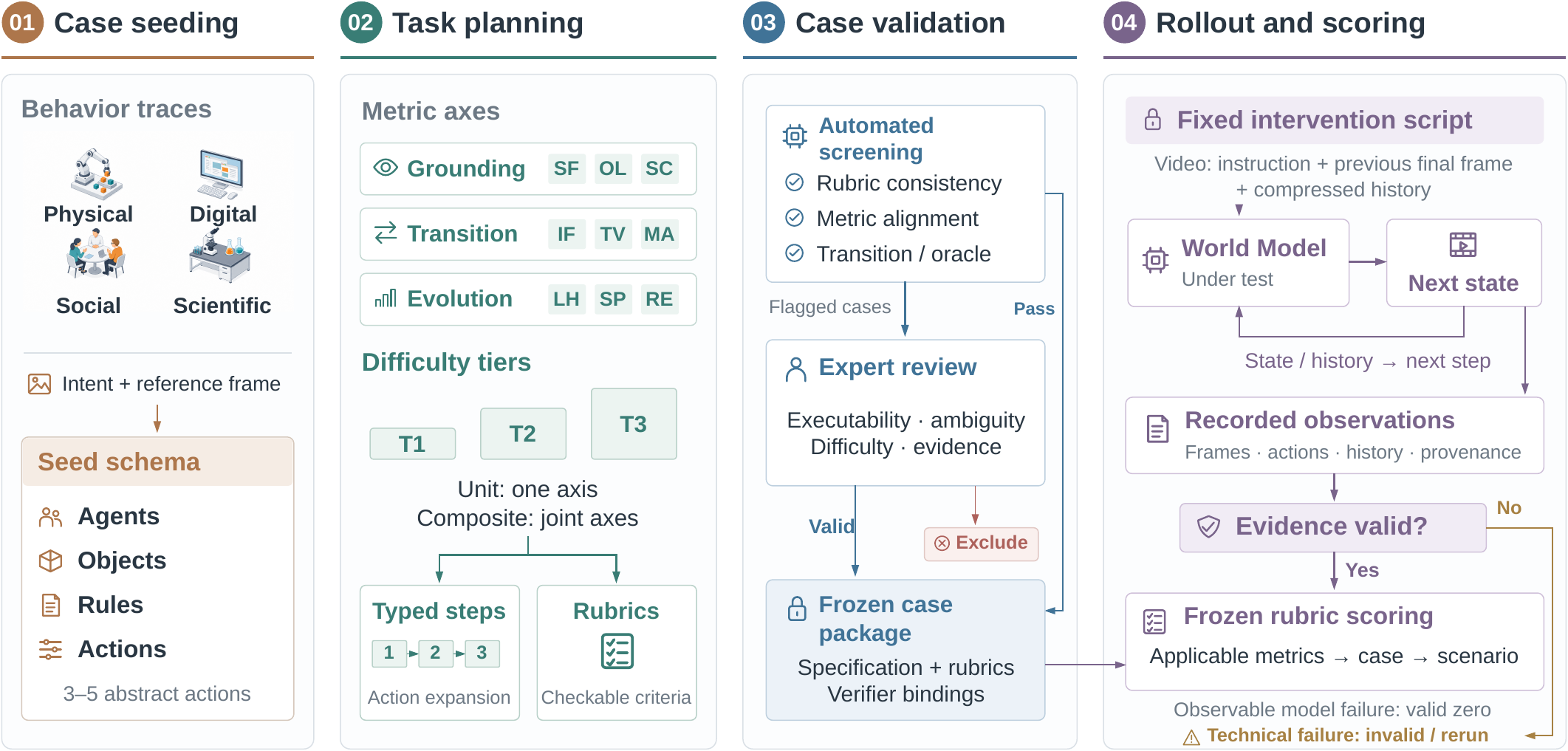}
  \caption{
  \textbf{World Model task construction and evaluation.}
  Behavior traces provide case seeds, which are expanded into intervention
  sequences and case-specific rubrics. Automated screening and expert
  review are construction checks. The video path is illustrated; input and history access depend on the evaluated configuration. Fixed-intervention rollouts are scored against the case rubrics; construction review does not constitute independent agreement testing of the resulting scores.
  }
  \label{fig:world-model-task-pipeline}
\end{figure}

The World Model track tests prescribed action effects and preservation of entities, state, and rules (Figure~\ref{fig:world-model-task-pipeline}).


\noindent\textbf{Benchmark construction.}
The World Model evaluation comprises 400 cases across eight scenarios, with 50 cases per scenario: Embodiment, Driving, and Daily Life in the Physical domain; Web and Game in the Digital domain; Human Interaction in the Social domain; and Natural Science and Expert Science in the Scientific domain. Cases specify interventions and checks for requested action effects, preservation of prior state and identities, shared-state consistency, and rule adherence. This allocation gives 150 Physical, 100 Digital, 50 Social, and 100 Scientific cases. Construction review is not an independent audit of output scores.

\noindent\textbf{Evaluation and aggregation.}
Each rollout uses fixed interventions and the observations and history permitted by its configuration. Case-specific checks assess action effects, state and identity preservation, shared-state consistency, and rule adherence. Observable model errors remain evaluation outcomes; technical failures and unsupported scenarios must be recorded separately. WM Overall is the mean of the eight scenario scores on a 0--100 scale. With 50 cases per scenario, this is also a case-balanced aggregate when each scenario score averages the same number of equally weighted case scores. It is not an equal-domain average: the scenario allocation gives Physical, Digital, Social, and Scientific weights of 37.5\%, 25\%, 12.5\%, and 25\%. Missing scenario scores are not zeros. A mean over fewer scenarios must be labeled as partial coverage and must not enter the full-coverage Overall ranking.

\paragraph{Modality and execution scope.}
Table~\ref{tab:world-model-results} separates image and video configurations. The shared target is observable task-consistent state evolution; video additionally supplies temporal evidence unavailable from a static checkpoint. A plausible still image cannot establish timing or an oscillation period. Text-only World Models are outside this visual-observation scope and would require a separately specified observation interface and rubric. Per-model endpoint snapshots, generation duration, complete histories, budgets, and retry and invalid-run counts are not consolidated, limiting cross-modality interpretation.

\section{Experiments}
\label{sec:experiments}
\label{sec:results}

We evaluate 10 Agent configurations on 1,088 items in 15 categories (Table~\ref{tab:agent-results}) and 16 World Model configurations with reported coverage of up to 400 cases in eight scenarios (Table~\ref{tab:world-model-results}). Agent Index and WM Overall use different tasks, metrics, and aggregation rules and are not directly comparable. Sections~\ref{app:agent-evaluation} and~\ref{app:world-model-evaluation} present illustrative case evidence.

\subsection{Agent Results}



\paragraph{Execution scores vary across task settings.}
The highest displayed scores are 9.5 for Manipulation, 11.2 for Action Sequencing, and 5.6 for Desktop Application Interaction, compared with 29.6 for Web, 42.9 for Mobile, and 47.1 for Game Environment Interaction. A high Index can therefore coexist with low performance on specific tasks. These contrasts depend on task requirements and scoring rules and do not establish equal difficulty across categories.

The Navigation and Action Sequencing examples distinguish correct local judgments from valid complete solutions: feasible destinations can be paired with nonexistent route edges, and correct clip membership with incorrect ordering. These cases illustrate constraint violations, not their frequency or contribution to the rankings.

\paragraph{Scientific scores combine distinct evaluation units.}
GPT 6 Astra leads the within-domain case-weighted score at 50.6, followed by DeepSeek V4.1 Flash at 44.0 and GPT 5.6 Sol at 42.2. Astra leads Mathematics \& Computation (60.8) and Physical Sciences (52.2), Sol leads Health \& Medicine (45.4), and DeepSeek leads Life Sciences (53.7). The aggregate combines ALE task scores, strict SWE and ToolChoice outcomes, and graded Edge results, including our Borden/DABIC clipping convention (Section~\ref{app:edge-aggregation}). It is not a uniform success rate.

\paragraph{Aggregate rank depends on task composition.}
GPT 6 Astra leads Index at 31.3, followed by GPT 5.6 Sol (20.7), Gemini 3.6 Flash (20.1), Qwen 3.8 Max (20.0), and Kimi K3 (19.8). Astra leads nine categories, DeepSeek three, Sol two, and Seed one. Under equal-domain weighting, DeepSeek moves from sixth to second and Kimi from fifth to fourth; Astra remains first (Section~\ref{app:agent-weight-sensitivity}). The primary Index assigns 48.16\% weight to Physical and 10.02\% to Scientific tasks, reflecting inventory counts. Rankings describe this weighting and the evaluated configurations; small gaps derived from rounded baseline means do not establish statistical superiority.


\begin{table}[!t]
\centering
\caption{\textbf{Agent leaderboard by overall Index.} 10 model configurations are compared across 15 task categories and 1,088 evaluation items on a 0--100 scale. $N$ is the number of items. Index uses case-count weights (524 Physical, 255 Digital, 200 Social, and 109 Scientific); it is not an equal-domain average. Bold and underline indicate the highest and second-highest distinct scores, including ties. Model columns are ordered by Index from highest to lowest, using the rounded baseline means and added item-level scores before final rounding for ordering and Index highlights.}
\label{tab:agent-results}
\label{tab:cross-benchmark-performance}
\scriptsize
\setlength{\tabcolsep}{1pt}
\renewcommand{\arraystretch}{1.45}
\definecolor{agentBarA}{HTML}{A892CE}
\definecolor{agentBarB}{HTML}{B194CA}
\definecolor{agentBarC}{HTML}{BA97C5}
\definecolor{agentBarD}{HTML}{C499BF}
\definecolor{agentBarE}{HTML}{CC9AB6}
\definecolor{agentBarF}{HTML}{D49CAC}
\definecolor{agentBarG}{HTML}{DA9EA1}
\definecolor{agentBarH}{HTML}{E2A396}
\definecolor{agentBarI}{HTML}{E8A98C}
\definecolor{agentBarJ}{HTML}{ECB082}
\definecolor{agentBarK}{HTML}{EEB679}
\definecolor{agentBarL}{HTML}{F0BD73}
\resizebox{\textwidth}{!}{%
\begin{tabular}{l r *{10}{c}}
\toprule
Category & $N$ &  \raisebox{-1.6pt}{\includegraphics[height=8pt]{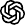}}\,GPT & \raisebox{-1.6pt}{\includegraphics[height=8pt]{assets/model_logos/openai.pdf}}\,GPT & \raisebox{-1.6pt}{\includegraphics[height=8pt]{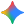}}\,Gemini & \raisebox{-1.6pt}{\includegraphics[height=8pt]{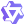}}\,Qwen & \raisebox{-1.6pt}{\includegraphics[height=8pt]{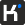}}\,Kimi & \raisebox{-1.6pt}{\includegraphics[height=8pt]{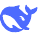}}\,DeepSeek & \raisebox{-1.6pt}{\includegraphics[height=8pt]{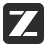}}\,GLM & \raisebox{-1.6pt}{\includegraphics[height=8pt]{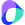}}\,Seed & \raisebox{-1.6pt}{\includegraphics[height=8pt]{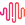}}\,MiniMax & \raisebox{-1.6pt}{\includegraphics[height=8pt]{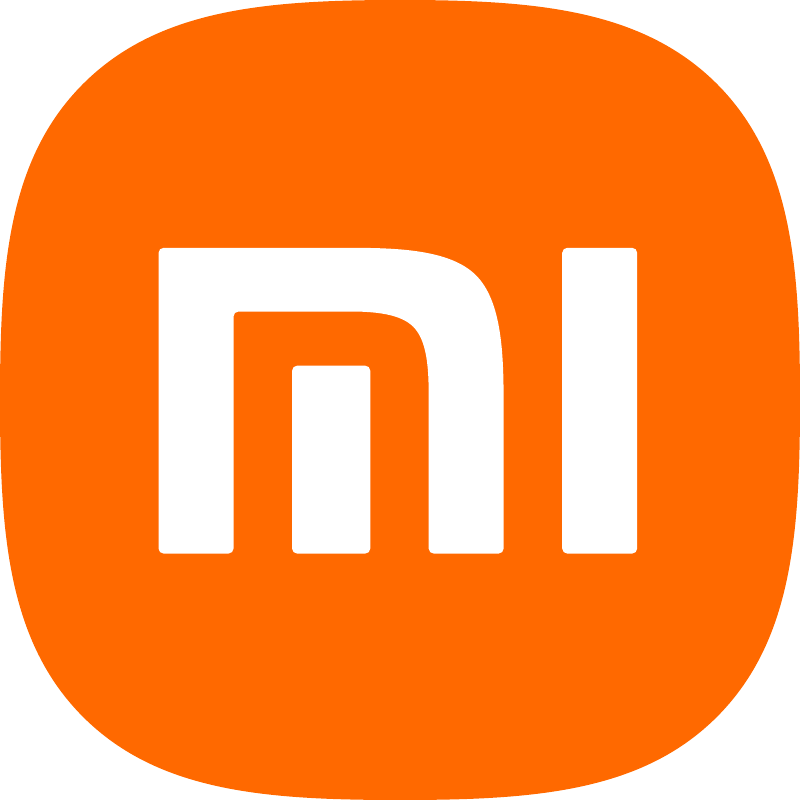}}\,MiMo \\
 & & 6 Astra & 5.6 Sol & 3.6 Flash & 3.8 Max & K3 & V4.1 Flash & 5.3 Flash & 2.1 Pro & M3 & V2.5 \\
\rule{0pt}{10pt} &
& {\tiny\cite{openai2026gpt6}} 
& {\tiny\cite{openai2026gpt56}} 
& {\tiny\cite{google2026gemini36flash}} 
& {\tiny\cite{alibaba2026qwen38max}} 
& {\tiny\cite{moonshot2026kimik3}} 
& {\tiny\cite{deepseekai2026deepseekv41flash}} 
& {\tiny\cite{zhipu2026glm53}} 
& {\tiny\cite{bytedance2026seed21}} 
& {\tiny\cite{minimax2026m3}} 
& {\tiny\cite{xiaomi2026mimo25}} 
 \\
\midrule
\rowcolor{leaderphysical!10}\multicolumn{12}{c}{\textcolor{leaderphysical}{\textbf{Physical World}}} \\
\rowcolor{leaderphysical!5}
PL & 30 & \textbf{63.3} & 16.7 & 10.0 & 20.0 & 16.7 & \underline{23.2} & 10.0 & 3.3 & 3.3 & 6.7 \\
\rowcolor{leaderphysical!5}
MAN & 42 & \underline{7.1} & \textbf{9.5} & 4.8 & \underline{7.1} & 4.8 & 4.8 & \underline{7.1} & 0.0 & 0.0 & \underline{7.1} \\
\rowcolor{leaderphysical!5}
NAV & 282 & \textbf{49.7} & 36.2 & \underline{37.6} & 36.2 & 33.3 & 23.1 & 23.1 & 8.9 & 22.0 & 15.6 \\
\rowcolor{leaderphysical!5}
AS & 170 & \textbf{11.2} & \underline{6.5} & 0.0 & \underline{6.5} & 4.7 & 2.4 & 4.1 & 3.5 & 0.6 & 0.0 \\
\rowcolor{leaderdigital!10}\multicolumn{12}{c}{\textcolor{leaderdigital}{\textbf{Digital World}}} \\
\rowcolor{leaderdigital!5}
WI & 74 & \textbf{29.6} & \underline{24.9} & 7.8 & 20.1 & 14.3 & 14.1 & 20.7 & 18.5 & 22.0 & 11.8 \\
\rowcolor{leaderdigital!5}
DAI & 95 & \textbf{5.6} & 2.1 & 2.1 & 3.2 & 1.1 & \underline{4.5} & 1.6 & 1.1 & 2.1 & 1.6 \\
\rowcolor{leaderdigital!5}
MAI & 35 & \textbf{42.9} & 8.6 & 22.9 & 20.0 & \underline{28.6} & 8.6 & 8.6 & 2.9 & 8.6 & 14.3 \\
\rowcolor{leaderdigital!5}
GEI & 51 & \textbf{47.1} & 13.3 & 25.4 & \underline{27.1} & 15.9 & 20.3 & 20.0 & 15.1 & 10.4 & 18.0 \\
\rowcolor{leadersocial!10}\multicolumn{12}{c}{\textcolor{leadersocial}{\textbf{Social World}}} \\
\rowcolor{leadersocial!5}
CSI & 87 & 31.5 & 15.7 & \underline{48.0} & 18.1 & 32.5 & 10.1 & 14.2 & \textbf{50.6} & 13.5 & 13.5 \\
\rowcolor{leadersocial!5}
WSC-RG & 56 & 8.3 & 10.9 & 8.9 & 11.7 & 9.3 & \textbf{43.9} & 9.5 & 11.0 & 10.2 & \underline{12.9} \\
\rowcolor{leadersocial!5}
TEH-RG & 57 & 9.6 & 13.1 & 10.4 & 11.2 & 10.4 & \textbf{21.1} & 11.5 & 12.8 & 12.0 & \underline{14.9} \\
\rowcolor{leaderscience!10}\multicolumn{12}{c}{\textcolor{leaderscience}{\textbf{Scientific World}}} \\
\rowcolor{leaderscience!5}
MC & 15 & \textbf{60.8} & 45.2 & 38.5 & 37.1 & \underline{49.5} & 45.5 & 40.0 & 32.7 & 10.3 & 31.9 \\
\rowcolor{leaderscience!5}
HM & 25 & \underline{44.3} & \textbf{45.4} & 10.2 & 22.8 & 39.1 & 37.8 & 41.8 & 22.8 & 8.6 & 18.6 \\
\rowcolor{leaderscience!5}
LS & 32 & \underline{48.9} & 46.6 & 28.2 & 30.1 & 37.8 & \textbf{53.7} & 36.5 & 29.5 & 19.2 & 25.9 \\
\rowcolor{leaderscience!5}
PS & 37 & \textbf{52.2} & 35.0 & 25.3 & 19.7 & 20.9 & \underline{39.2} & 39.1 & 14.5 & 7.9 & 15.2 \\

\midrule
\textcolor{strictorange}{\textbf{Index $\uparrow$}} & 1,088 & \textbf{31.3} & \underline{20.7} & 20.1 & 20.0 & 19.8 & 18.3 & 16.1 & 12.7 & 11.7 & 11.4 \\
\addlinespace[3pt]
\shortstack[l]{Index\\comparison} & {\setlength{\unitlength}{1mm}\begin{picture}(9,13)(0,-0.5)\put(8,12){\makebox(0,0)[r]{35}}\put(8,6.85714){\makebox(0,0)[r]{20}}\put(8,0){\makebox(0,0)[r]{0}}\end{picture}} & {\setlength{\unitlength}{1mm}\begin{picture}(6,13)(0,-0.5)\put(0,6.85714){\color{black!10}\line(1,0){6}}\put(1.6,0){\color{agentBarA}\rule{2.8mm}{10.71788mm}}\put(0,0){\color{black!35}\line(1,0){6}}\end{picture}} & {\setlength{\unitlength}{1mm}\begin{picture}(6,13)(0,-0.5)\put(0,6.85714){\color{black!10}\line(1,0){6}}\put(1.6,0){\color{agentBarB}\rule{2.8mm}{7.10796mm}}\put(0,0){\color{black!35}\line(1,0){6}}\end{picture}} & {\setlength{\unitlength}{1mm}\begin{picture}(6,13)(0,-0.5)\put(0,6.85714){\color{black!10}\line(1,0){6}}\put(1.6,0){\color{agentBarC}\rule{2.8mm}{6.90796mm}}\put(0,0){\color{black!35}\line(1,0){6}}\end{picture}} & {\setlength{\unitlength}{1mm}\begin{picture}(6,13)(0,-0.5)\put(0,6.85714){\color{black!10}\line(1,0){6}}\put(1.6,0){\color{agentBarD}\rule{2.8mm}{6.86063mm}}\put(0,0){\color{black!35}\line(1,0){6}}\end{picture}} & {\setlength{\unitlength}{1mm}\begin{picture}(6,13)(0,-0.5)\put(0,6.85714){\color{black!10}\line(1,0){6}}\put(1.6,0){\color{agentBarE}\rule{2.8mm}{6.77866mm}}\put(0,0){\color{black!35}\line(1,0){6}}\end{picture}} & {\setlength{\unitlength}{1mm}\begin{picture}(6,13)(0,-0.5)\put(0,6.85714){\color{black!10}\line(1,0){6}}\put(1.6,0){\color{agentBarF}\rule{2.8mm}{6.29088mm}}\put(0,0){\color{black!35}\line(1,0){6}}\end{picture}} & {\setlength{\unitlength}{1mm}\begin{picture}(6,13)(0,-0.5)\put(0,6.85714){\color{black!10}\line(1,0){6}}\put(1.6,0){\color{agentBarH}\rule{2.8mm}{5.51353mm}}\put(0,0){\color{black!35}\line(1,0){6}}\end{picture}} & {\setlength{\unitlength}{1mm}\begin{picture}(6,13)(0,-0.5)\put(0,6.85714){\color{black!10}\line(1,0){6}}\put(1.6,0){\color{agentBarI}\rule{2.8mm}{4.35968mm}}\put(0,0){\color{black!35}\line(1,0){6}}\end{picture}} & {\setlength{\unitlength}{1mm}\begin{picture}(6,13)(0,-0.5)\put(0,6.85714){\color{black!10}\line(1,0){6}}\put(1.6,0){\color{agentBarJ}\rule{2.8mm}{4.02503mm}}\put(0,0){\color{black!35}\line(1,0){6}}\end{picture}} & {\setlength{\unitlength}{1mm}\begin{picture}(6,13)(0,-0.5)\put(0,6.85714){\color{black!10}\line(1,0){6}}\put(1.6,0){\color{agentBarK}\rule{2.8mm}{3.91622mm}}\put(0,0){\color{black!35}\line(1,0){6}}\end{picture}} \\
\bottomrule
\end{tabular}}
\par\smallskip
\begin{minipage}{\textwidth}\scriptsize
Category names describe reporting groups, not independent latent abilities. Physical Sciences includes physics, chemistry, materials, and Earth/space applications. Game Environment Interaction includes Godot and GameWorld without double counting. The bars show Index on a shared zero-based 0--35 axis with a light reference at 20; the score range remains 0--100. A light purple-to-orange palette follows the displayed model order; color differences do not indicate statistical significance. Six EdgeBench optimization tasks are included using the convention in Section~\ref{app:edge-aggregation}. Scientific scores combine the original ALE category means with 30 SWE-bench repair tasks, ten ToolChoice tasks, and six EdgeBench tasks. SWE and ToolChoice receive 100 for strict success and 0 otherwise; Edge uses graded scores. Borden and DABIC are clipped to [0,100] by Dual Exam, not by an official EdgeBench normalization rule. Derived scores and Index use the original displayed one-decimal means as baseline and the added item-level scores before final rounding. These are model--harness configurations; scientific evaluation settings and limitations are described with Table~\ref{tab:scientific-source-results}.
\par\smallskip\textbf{Abbreviations:} PL = Planning; MAN = Manipulation; NAV = Navigation; AS = Action Sequencing; WI = Web Interaction; DAI = Desktop Application Interaction; MAI = Mobile Application Interaction; GEI = Game Environment Interaction; CSI = Coordination \& Strategic Interaction; WSC-RG = Water, Soil \& Climate Resource Governance; TEH-RG = Terrestrial Ecosystem \& Health Resource Governance; MC = Mathematics \& Computation; HM = Health \& Medicine; LS = Life Sciences; PS = Physical Sciences.
\end{minipage}
\end{table}

\begin{table}[!htp]
\centering
\caption{\textbf{World Model scenario scores.} Sixteen configurations are evaluated on eight scenarios (50 cases each; 400 cases for full coverage). Columns group image and video outputs, ordered by Overall within each full-coverage group. The separate final column covers six scenarios. Cross-modality comparisons are descriptive.}
\label{tab:world-model-results}
\begingroup\normalfont\fontfamily{phv}\selectfont\mdseries\fontsize{10}{12}\selectfont
\definecolor{wmBarO}{HTML}{F3C566}
\definecolor{wmBarN}{HTML}{F2C16C}
\definecolor{wmBarM}{HTML}{F0BD73}
\definecolor{wmBarL}{HTML}{EEB879}
\definecolor{wmBarK}{HTML}{EAB483}
\definecolor{wmBarJ}{HTML}{E5B08D}
\definecolor{wmBarI}{HTML}{DFAC96}
\definecolor{wmBarH}{HTML}{D9A89F}
\definecolor{wmBarG}{HTML}{D2A4A8}
\definecolor{wmBarF}{HTML}{CBA1B1}
\definecolor{wmBarE}{HTML}{C49EB9}
\definecolor{wmBarD}{HTML}{BC9BC0}
\definecolor{wmBarC}{HTML}{B598C5}
\definecolor{wmBarB}{HTML}{AE95CA}
\definecolor{wmBarA}{HTML}{A892CE}
\definecolor{wmHeader}{HTML}{E7E0EF}
\definecolor{wmPhysical}{HTML}{EEE8F5}
\definecolor{wmDigital}{HTML}{E7EDF5}
\definecolor{wmSocial}{HTML}{E8F0EB}
\definecolor{wmScience}{HTML}{F8EDDB}
\setlength{\tabcolsep}{1.5pt}
\renewcommand{\arraystretch}{1.25}
\resizebox{\linewidth}{!}{%
\begin{tabular}{@{}l*{15}{c}@{\hspace{4pt}}c@{}}
\rowcolor{wmHeader}
 & \multicolumn{4}{c}{\scriptsize\textbf{Image outputs: full coverage}} & \multicolumn{11}{c}{\scriptsize\textbf{Video outputs: full coverage}} & \cellcolor{wmScience}\makecell{\scriptsize \scriptsize\textbf{Video}\\\scriptsize Partial$^\dagger$} \\
\rowcolor{white}
 & \includegraphics[height=10pt,width=13pt,keepaspectratio]{assets/model_logos/openai.pdf} & \includegraphics[height=10pt,width=13pt,keepaspectratio]{assets/model_logos/qwen.pdf} & \includegraphics[height=10pt,width=13pt,keepaspectratio]{assets/model_logos/gemini.pdf} & \includegraphics[height=10pt,width=13pt,keepaspectratio]{assets/model_logos/doubao.pdf} & \includegraphics[height=10pt,width=13pt,keepaspectratio]{assets/model_logos/minimax.pdf} & \includegraphics[height=10pt,width=13pt,keepaspectratio]{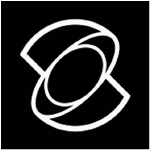} & \includegraphics[height=10pt,width=13pt,keepaspectratio]{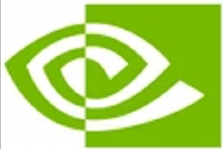} & \includegraphics[height=10pt,width=13pt,keepaspectratio]{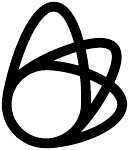} & \includegraphics[height=10pt,width=13pt,keepaspectratio]{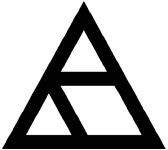} & \includegraphics[height=10pt,width=13pt,keepaspectratio]{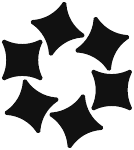} & \includegraphics[height=10pt,width=13pt,keepaspectratio]{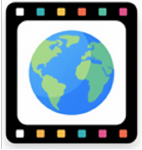} & \includegraphics[height=10pt,width=13pt,keepaspectratio]{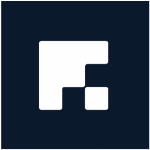} & \includegraphics[height=10pt,width=13pt,keepaspectratio]{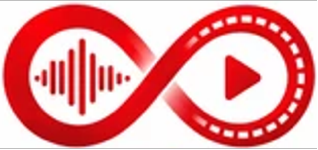} & \includegraphics[height=10pt,width=13pt,keepaspectratio]{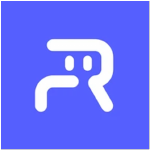} & \includegraphics[height=10pt,width=13pt,keepaspectratio]{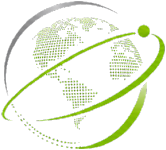} & \includegraphics[height=10pt,width=13pt,keepaspectratio]{assets/model_logos/doubao.pdf} \\
\rowcolor{white}
\textbf{Scenario} & \makecell{\scriptsize GPT\\\scriptsize Image 2} & \makecell{\scriptsize Qwen\\\scriptsize Image 3.0} & \makecell{\scriptsize Nano\\\scriptsize Banana 2} & \makecell{\scriptsize Seedream\\\scriptsize 5.0 Pro} & \makecell{\scriptsize MiniMax\\\scriptsize H3} & \makecell{\scriptsize Kling\\\scriptsize 3.0} & \makecell{\scriptsize Cosmos3\\\scriptsize Super} & \makecell{\scriptsize LTX\\\scriptsize 2.5} & \makecell{\scriptsize Alaya\\\scriptsize World} & \makecell{\scriptsize Zing\\\scriptsize 0.5} & \makecell{\scriptsize YUME\\\scriptsize 1.5} & \makecell{\scriptsize Matrix\\\scriptsize Game 3.5} & \makecell{\scriptsize Echo\\\scriptsize WM} & \makecell{\scriptsize LingBot\\\scriptsize World v2} & \makecell{\scriptsize SANA\\\scriptsize WM} & \makecell{\scriptsize Seedance\\\scriptsize 2.5$^\dagger$} \\
\rowcolor{white}
 & \makecell{\tiny\citeauthor{openai2026gptimage2}\\\tiny\citeyearpar{openai2026gptimage2}} & \makecell{\tiny\citeauthor{qwen2026image3}\\\tiny\citeyearpar{qwen2026image3}} & \makecell{\tiny\citeauthor{google2026nanobanana2}\\\tiny\citeyearpar{google2026nanobanana2}} & \makecell{\tiny\citeauthor{bytedance2026seedream5pro}\\\tiny\citeyearpar{bytedance2026seedream5pro}} & \makecell{\tiny\citeauthor{minimax2026h3}\\\tiny\citeyearpar{minimax2026h3}} & \makecell{\tiny\citeauthor{kuaishou2026kling}\\\tiny\citeyearpar{kuaishou2026kling}} & \makecell{\tiny\citeauthor{nvidia2026cosmos3}\\\tiny\citeyearpar{nvidia2026cosmos3}} & \makecell{\tiny\citeauthor{lightricks2026ltx25}\\\tiny\citeyearpar{lightricks2026ltx25}} & \makecell{\tiny\citetalias{team2026alayaworldfull}\\\tiny\citeyearpar{team2026alayaworldfull}} & \makecell{\tiny\citeauthor{seedleap2026zing05}\\\tiny\citeyearpar{seedleap2026zing05}} & \makecell{\tiny\citeauthor{mao2025yume15}\\\tiny\citeyearpar{mao2025yume15}} & \makecell{\tiny\citeauthor{qian2026matrixgame35enhancingrealtime}\\\tiny\citeyearpar{qian2026matrixgame35enhancingrealtime}} & \makecell{\tiny\citeauthor{zhang2026echowm}\\\tiny\citeyearpar{zhang2026echowm}} & \makecell{\tiny\citeauthor{lingbot-world-v2}\\\tiny\citeyearpar{lingbot-world-v2}} & \makecell{\tiny\citeauthor{zhu2026sana}\\\tiny\citeyearpar{zhu2026sana}} & \makecell{\tiny\citeauthor{bytedance2026seedance25}\\\tiny\citeyearpar{bytedance2026seedance25}} \\
\midrule
\rowcolor{wmPhysical}\multicolumn{16}{c}{\textbf{Physical World}} & \\
\rowcolor{wmPhysical!30}
\textbf{Emb.} & 20.9 & 20.5 & 25.8 & 13.7 & \textbf{53.8} & 38.7 & 44.7 & 41.9 & 21.1 & 12.8 & 27.9 & 14.3 & 14.4 & 25.9 & 21.6 & \underline{51.2} \\
\rowcolor{wmPhysical!30}
\textbf{Drive} & 15.0 & \textbf{19.8} & \underline{17.3} & 12.0 & 12.7 & 11.3 & 7.3 & 10.5 & 3.9 & 8.0 & 6.0 & 6.2 & 6.2 & 3.0 & 1.1 & 12.3 \\
\rowcolor{wmPhysical!30}
\textbf{Daily} & \underline{45.2} & \textbf{46.1} & 32.6 & 33.9 & 23.3 & 28.7 & 31.5 & 25.1 & 40.7 & 22.5 & 20.9 & 20.4 & 19.4 & 16.5 & 19.5 & -- \\
\rowcolor{wmDigital}\multicolumn{16}{c}{\textbf{Digital World}} & \\
\rowcolor{wmDigital!30}
\textbf{Web} & \textbf{16.5} & 10.7 & \underline{11.8} & 9.5 & 3.7 & 2.7 & 1.8 & 5.0 & 5.8 & 3.4 & 5.9 & 4.0 & 0.7 & 1.5 & 2.4 & 6.0 \\
\rowcolor{wmDigital!30}
\textbf{Game} & \textbf{13.3} & 11.5 & \textbf{13.3} & \underline{12.9} & 8.5 & 6.0 & 5.9 & 7.8 & 9.3 & 7.2 & 6.0 & 9.4 & 1.6 & 0.4 & 4.7 & 9.6 \\
\rowcolor{wmSocial}\multicolumn{16}{c}{\textbf{Social World}} & \\
\rowcolor{wmSocial!30}
\textbf{Human} & \textbf{22.9} & \underline{21.6} & 20.0 & 4.7 & 10.7 & 18.2 & 12.6 & 11.5 & 18.9 & 19.3 & 10.1 & 7.7 & 12.1 & 6.5 & 2.6 & -- \\
\rowcolor{wmScience}\multicolumn{16}{c}{\textbf{Scientific World}} & \\
\rowcolor{wmScience!30}
\textbf{Nat. Sci.} & \textbf{21.5} & \textbf{21.5} & \underline{19.5} & 14.3 & 14.6 & 14.0 & 11.5 & 10.9 & 9.0 & 10.5 & 10.2 & 3.6 & 8.2 & 8.6 & 7.9 & 15.1 \\
\rowcolor{wmScience!30}
\textbf{Exp. Sci.} & \textbf{18.6} & 17.4 & 15.5 & 7.6 & 11.2 & 17.3 & 11.9 & 10.7 & 14.3 & 15.7 & 5.5 & 12.9 & 8.1 & 6.7 & 2.3 & \underline{17.9} \\
\rowcolor{wmHeader!65}
\textbf{Overall $\uparrow$} & \textbf{21.7} & \underline{21.1} & 19.5 & 13.6 & 17.3 & 17.1 & 15.9 & 15.4 & 15.4 & 12.4 & 11.6 & 9.8 & 8.8 & 8.6 & 7.8 & 18.7$^\dagger$ \\
\addlinespace[4pt]
\makecell[l]{\scriptsize\textbf{Overall}\\\scriptsize\textbf{comparison}} & {\setlength{\unitlength}{1mm}\begin{picture}(7,20)(0,-1)\put(0,0){\color{black!30}\line(1,0){7}}\put(0,9.7143){\color{black!8}\line(1,0){7}}\put(0,17){\color{black!8}\line(1,0){7}}\put(1.6,0){\color{wmBarA}\rule{3.8mm}{10.54000mm}}\end{picture}} & {\setlength{\unitlength}{1mm}\begin{picture}(7,20)(0,-1)\put(0,0){\color{black!30}\line(1,0){7}}\put(0,9.7143){\color{black!8}\line(1,0){7}}\put(0,17){\color{black!8}\line(1,0){7}}\put(1.6,0){\color{wmBarB}\rule{3.8mm}{10.24857mm}}\end{picture}} & {\setlength{\unitlength}{1mm}\begin{picture}(7,20)(0,-1)\put(0,0){\color{black!30}\line(1,0){7}}\put(0,9.7143){\color{black!8}\line(1,0){7}}\put(0,17){\color{black!8}\line(1,0){7}}\put(1.6,0){\color{wmBarC}\rule{3.8mm}{9.47143mm}}\end{picture}} & {\setlength{\unitlength}{1mm}\begin{picture}(7,20)(0,-1)\put(0,0){\color{black!30}\line(1,0){7}}\put(0,9.7143){\color{black!8}\line(1,0){7}}\put(0,17){\color{black!8}\line(1,0){7}}\put(1.6,0){\color{wmBarD}\rule{3.8mm}{6.60571mm}}\end{picture}} & {\setlength{\unitlength}{1mm}\begin{picture}(7,20)(0,-1)\put(0,0){\color{black!30}\line(1,0){7}}\put(0,9.7143){\color{black!8}\line(1,0){7}}\put(0,17){\color{black!8}\line(1,0){7}}\put(1.6,0){\color{wmBarE}\rule{3.8mm}{8.40286mm}}\end{picture}} & {\setlength{\unitlength}{1mm}\begin{picture}(7,20)(0,-1)\put(0,0){\color{black!30}\line(1,0){7}}\put(0,9.7143){\color{black!8}\line(1,0){7}}\put(0,17){\color{black!8}\line(1,0){7}}\put(1.6,0){\color{wmBarF}\rule{3.8mm}{8.30571mm}}\end{picture}} & {\setlength{\unitlength}{1mm}\begin{picture}(7,20)(0,-1)\put(0,0){\color{black!30}\line(1,0){7}}\put(0,9.7143){\color{black!8}\line(1,0){7}}\put(0,17){\color{black!8}\line(1,0){7}}\put(1.6,0){\color{wmBarG}\rule{3.8mm}{7.72286mm}}\end{picture}} & {\setlength{\unitlength}{1mm}\begin{picture}(7,20)(0,-1)\put(0,0){\color{black!30}\line(1,0){7}}\put(0,9.7143){\color{black!8}\line(1,0){7}}\put(0,17){\color{black!8}\line(1,0){7}}\put(1.6,0){\color{wmBarH}\rule{3.8mm}{7.48000mm}}\end{picture}} & {\setlength{\unitlength}{1mm}\begin{picture}(7,20)(0,-1)\put(0,0){\color{black!30}\line(1,0){7}}\put(0,9.7143){\color{black!8}\line(1,0){7}}\put(0,17){\color{black!8}\line(1,0){7}}\put(1.6,0){\color{wmBarI}\rule{3.8mm}{7.48000mm}}\end{picture}} & {\setlength{\unitlength}{1mm}\begin{picture}(7,20)(0,-1)\put(0,0){\color{black!30}\line(1,0){7}}\put(0,9.7143){\color{black!8}\line(1,0){7}}\put(0,17){\color{black!8}\line(1,0){7}}\put(1.6,0){\color{wmBarJ}\rule{3.8mm}{6.02286mm}}\end{picture}} & {\setlength{\unitlength}{1mm}\begin{picture}(7,20)(0,-1)\put(0,0){\color{black!30}\line(1,0){7}}\put(0,9.7143){\color{black!8}\line(1,0){7}}\put(0,17){\color{black!8}\line(1,0){7}}\put(1.6,0){\color{wmBarK}\rule{3.8mm}{5.63429mm}}\end{picture}} & {\setlength{\unitlength}{1mm}\begin{picture}(7,20)(0,-1)\put(0,0){\color{black!30}\line(1,0){7}}\put(0,9.7143){\color{black!8}\line(1,0){7}}\put(0,17){\color{black!8}\line(1,0){7}}\put(1.6,0){\color{wmBarL}\rule{3.8mm}{4.76000mm}}\end{picture}} & {\setlength{\unitlength}{1mm}\begin{picture}(7,20)(0,-1)\put(0,0){\color{black!30}\line(1,0){7}}\put(0,9.7143){\color{black!8}\line(1,0){7}}\put(0,17){\color{black!8}\line(1,0){7}}\put(1.6,0){\color{wmBarM}\rule{3.8mm}{4.27429mm}}\end{picture}} & {\setlength{\unitlength}{1mm}\begin{picture}(7,20)(0,-1)\put(0,0){\color{black!30}\line(1,0){7}}\put(0,9.7143){\color{black!8}\line(1,0){7}}\put(0,17){\color{black!8}\line(1,0){7}}\put(1.6,0){\color{wmBarN}\rule{3.8mm}{4.17714mm}}\end{picture}} & {\setlength{\unitlength}{1mm}\begin{picture}(7,20)(0,-1)\put(0,0){\color{black!30}\line(1,0){7}}\put(0,9.7143){\color{black!8}\line(1,0){7}}\put(0,17){\color{black!8}\line(1,0){7}}\put(1.6,0){\color{wmBarO}\rule{3.8mm}{3.78857mm}}\end{picture}} & {\setlength{\unitlength}{1mm}\begin{picture}(7,20)(0,-1)\put(0,0){\color{black!30}\line(1,0){7}}\put(0,9.7143){\color{black!8}\line(1,0){7}}\put(0,17){\color{black!8}\line(1,0){7}}\put(1.6,0){\color{strictorange}\linethickness{0.25mm}\framebox(3.8,9.08286){}}\put(3.5,10.5829){\makebox(0,0){$\dagger$}}\end{picture}} \\
\end{tabular}}
\par\smallskip
\begin{minipage}{\linewidth}\scriptsize
Overall is the source eight-scenario mean for full-coverage configurations. Bold indicates the best and underline the second-highest distinct score (ties included); scenario highlights include all available values. The partial-coverage mean is excluded from full-coverage Overall ranking.
\par $\dagger$ Seedance 2.5 covers six scenarios (300 cases); Daily Life and Human Interaction are unavailable due to face-generation restrictions. Missing values are not zero. Its six-scenario mean is shown separately with an outlined bar. Bars use a shared zero-based 0--35 axis with a reference at 20; scores remain on a 0--100 scale.
\par\smallskip
\textbf{Abbreviations:} Emb. = Embodiment; Drive = Driving; Daily = Daily Life; Web/Game/Human = Web/Game/Human Interaction; Nat./Exp. Sci. = Natural/Expert Science.
\end{minipage}\endgroup
\end{table}

\begin{figure}[!tbp]
\centering
\includegraphics[width=\linewidth,height=.79\textheight,keepaspectratio]{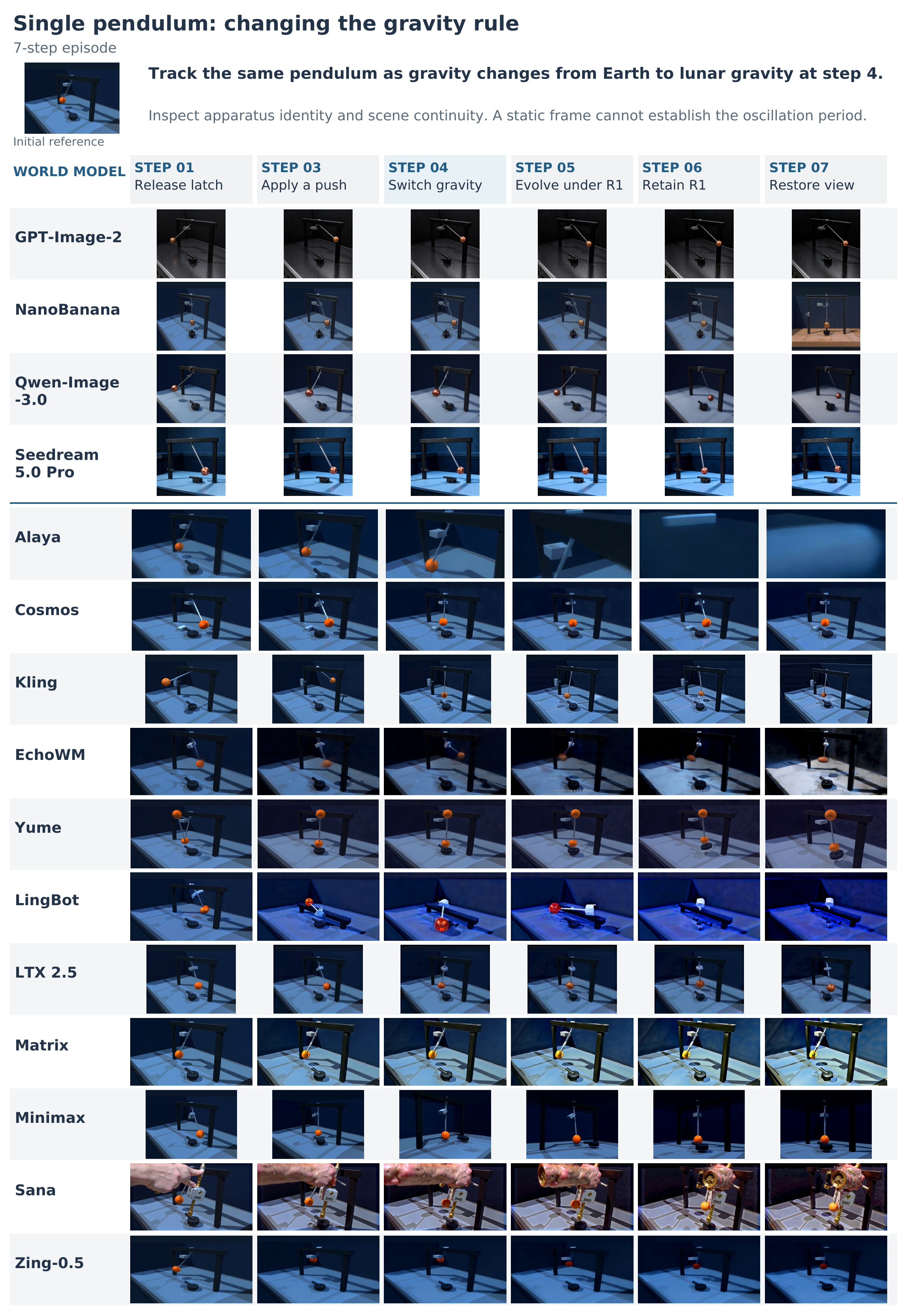}
\caption{\textbf{Single pendulum: rule editing and persistent apparatus.} Steps 1, 3, 4, 5, 6, and 7 cover release, a push, lunar-gravity editing, subsequent evolution, and view restoration. Static checkpoints reveal identity changes; oscillation period requires video. Yume's step-4 checkpoint is carried forward.}
\label{fig:wm-visual-pendulum}
\end{figure}

\begin{figure}[!tbp]
\centering
\includegraphics[width=\linewidth,height=.79\textheight,keepaspectratio]{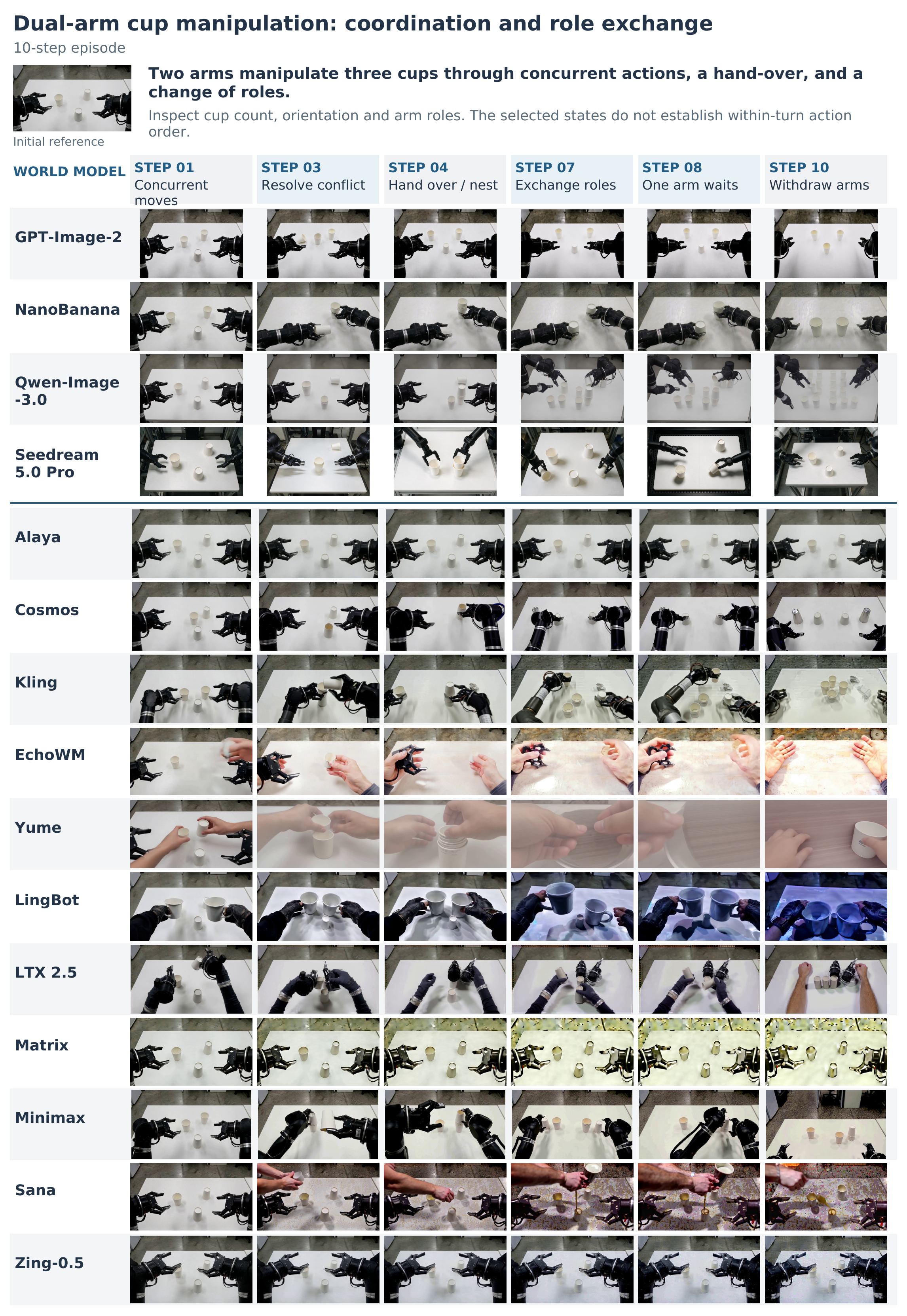}
\caption{\textbf{Dual-arm cup manipulation: coordination with persistent roles.} Steps 1, 3, 4, 7, 8, and 10 sample concurrent actions, contested nesting, hand-over, role exchange, one-arm action, and withdrawal. Images can expose object changes; order and sustained nonmovement require video.}
\label{fig:wm-visual-cupstack}
\end{figure}

\begin{figure}[!tbp]
\centering
\includegraphics[width=\linewidth,height=.79\textheight,keepaspectratio]{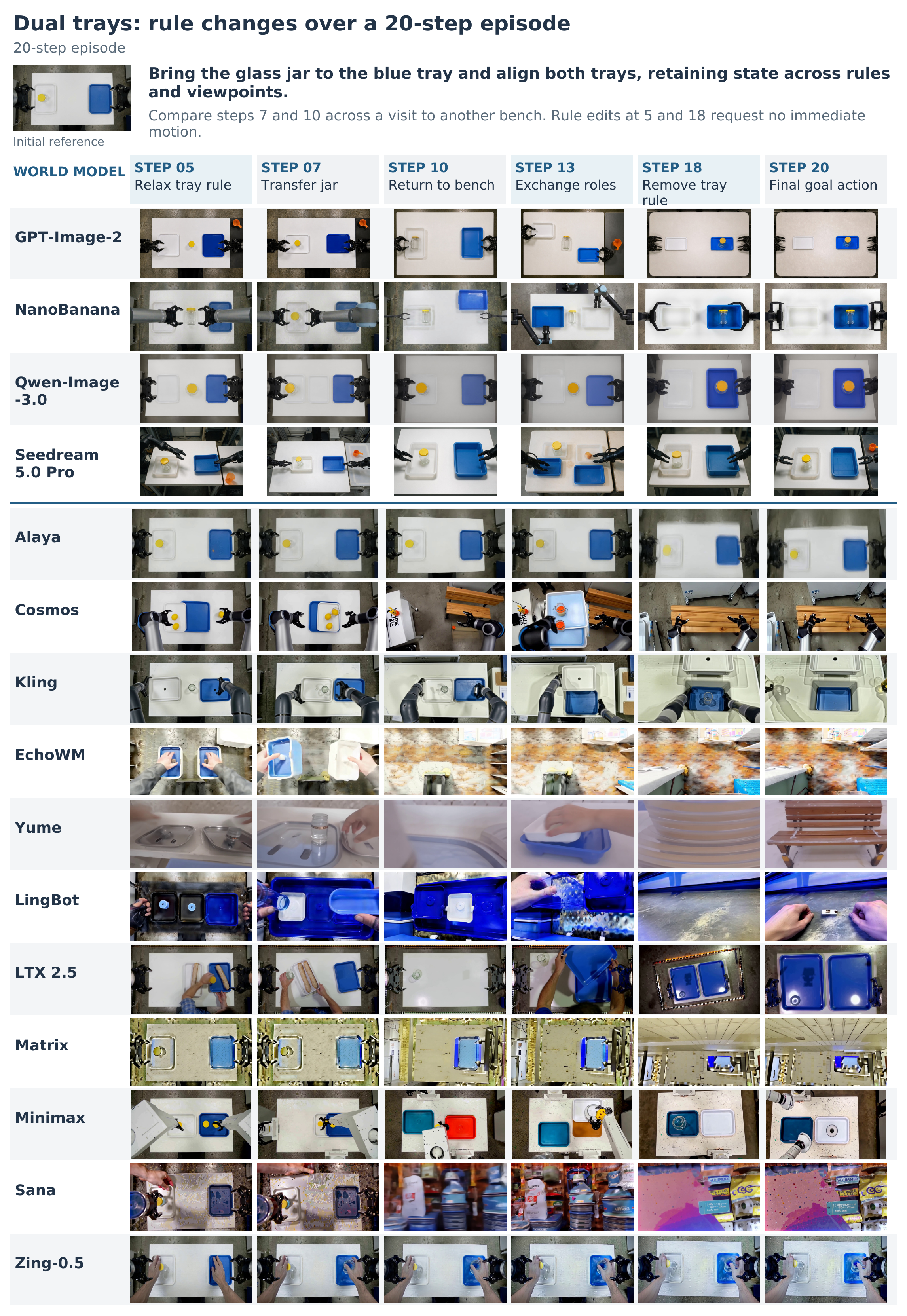}
\caption{\textbf{Dual trays: memory across interventions and revisits.} Steps 5, 7, 10, 13, 18, and 20 cover rule relaxation, jar transfer, return from another bench, role exchange, rule removal, and final action. Rule edits at 5 and 18 do not request immediate object motion.}
\label{fig:wm-visual-dualtray}
\end{figure}

\begin{samepage}
\subsection{World Model Results}

\label{sec:worldmodel_experiments}

\paragraph{Requested changes and preserved state are distinct requirements.}
Qwen-Image-3.0 restores the apparatus view in the pendulum case but adds cups in the stacking case (Figures~\ref{fig:wm-visual-pendulum} and~\ref{fig:wm-visual-cupstack}). MiniMax H3 retains three cups at the final stacking checkpoint (Figure~\ref{fig:wm-visual-cupstack}). In the separate dual-tray case, its step-10 output introduces a red tray, while GPT-Image-2 retains the original white and blue trays (Figure~\ref{fig:wm-visual-dualtray}). These selected cases illustrate state errors that could mislead a partner Agent; they do not explain aggregate ranking differences.\par
\end{samepage}

\paragraph{Digital scenarios have consistently low scores.}
No evaluated configuration exceeds 16.5 on Web or 13.3 on Game. MiniMax H3 scores 53.8 on Embodiment but 3.7 on Web and 8.5 on Game; Qwen-Image-3.0 scores 46.1 on Daily Life but 10.7 on Web and 11.5 on Game. These gaps describe the current tasks and scoring rules, not an isolated domain effect.

\paragraph{Scenario profiles matter for partner selection.}
GPT-Image-2 leads MiniMax H3 in Overall (21.7 versus 17.3), while MiniMax leads on Embodiment (53.8 versus 20.9). Overall weights eight scenarios equally; a partner Agent may encounter a different state and action distribution. Selecting an effective pairing therefore requires evaluation on the interactions that pair produces.

\section{Agent Case Analysis}
\label{app:agent-evaluation}
\label{sec:digital-world-case-studies}
\label{app:task-inventory}
\label{sec:scientific-capability-cases}

The current Agent leaderboard covers 1,088 items and 10 configurations. The following visual cases distinguish reference demonstrations, historical runs, submitted-artifact replays, and illustrative reconstructions. They explain task requirements and observed outcomes; they are not additional leaderboard trials or estimates of failure frequency.

\subsection{Task Classification and Counting Units}
\label{app:task-taxonomy}

The four domains organize task settings, not independent cognitive abilities. Physical categories describe action demands; Digital categories describe interfaces; Social categories describe interaction and resource settings; Scientific categories describe subject matter. Each item has one primary assignment for aggregation, even when its requirements overlap other categories.

\begin{table}[tbp]
\centering
\small
\begin{tabularx}{\linewidth}{lXr}
\hline
Domain & Task category & Items \\
\hline
Physical & Planning & 30 \\
& Manipulation & 42 \\
& Navigation & 282 \\
& Action Sequencing & 170 \\
Digital & Web Interaction & 74 \\
& Desktop Application Interaction & 95 \\
& Mobile Application Interaction & 35 \\
& Game Environment Interaction & 51 \\
Social & Coordination \& Strategic Interaction & 87 \\
& Water, Soil \& Climate Resource Governance & 56 \\
& Terrestrial Ecosystem \& Health Resource Governance & 57 \\
Scientific & Mathematics \& Computation & 15 \\
& Health \& Medicine & 25 \\
& Life Sciences & 32 \\
& Physical Sciences & 37 \\
\hline
Total & 15 reporting categories & 1,088 \\
\hline
\end{tabularx}
\caption{Current agent inventory with descriptive category names. Counts follow the current leaderboard configuration and do not imply equal difficulty or statistical independence. Godot and GameWorld are included within Game Environment Interaction and are not counted again.}
\label{tab:agent-task-taxonomy}
\end{table}

\paragraph{Physical and Digital assignments.}
Planning requires organizing dependent household actions to satisfy goals. Action Sequencing instead requires assigning shuffled clips to two tasks and reconstructing both orders. Navigation checks agent-specific traversability and routes; Manipulation checks object configurations. Digital categories identify Web, Desktop, Mobile, and Game environments, which share perception, planning, and recovery requirements. Browser-based games retain one primary assignment and are not counted twice.

\paragraph{Social assignments.}
The 200 cases instantiate 40 templates. Coordination \& Strategic Interaction has 87 cases from 17 templates, covering bargaining, congestion, public goods, negotiation, and persuasion. Water, Soil \& Climate Resource Governance has 56 cases from 11 templates; Terrestrial Ecosystem \& Health Resource Governance has 57 cases from 12 templates. The latter includes agricultural, forest, and antibiotic resources. Assignments follow the recorded primary resource setting, despite ecological overlap; shared templates also limit statistical independence.

\paragraph{Scientific assignments.}
Scientific categories follow the task objective: mathematical or computational output, health, biological investigation, or physical science. Source assignments are retained for traceability, including the quantum Ising-chain task in Mathematics \& Computation. Physical Sciences includes physics, chemistry, materials, and Earth/space applications, distinct from the action-oriented Physical domain. Its original 13 ALE tasks span molecular calculations, climate simulation, lake calibration, orbital transfer, Hubble-image processing, and chemical kinetics. Reassignment would require recomputing category scores.

\paragraph{Counting units.}
The Scientific inventory combines 99 full workflows and ten ToolChoice decision checkpoints (Table~\ref{tab:scientific-capability-counts}). Checkpoints inherit their parent repair task's subject and are separate evaluation items, not independent full-workflow trials. Six EdgeBench tasks are included; 84 other historical ALE tasks remain excluded.

\subsection{Sensitivity to Domain Weighting}
\label{app:agent-weight-sensitivity}
To test sensitivity to task composition, we first compute a case-weighted score within each domain, then average the four domain scores equally. The primary Index instead weights every item equally. Both use the same rounded baseline means and added-task outcomes. Extra decimal places explain ordering, not improved source precision; reweighting is not an uncertainty estimate.

\begin{table}[tbp]
\centering\small
\setlength{\tabcolsep}{4pt}
\begin{tabular}{lrrrr}
\toprule
Configuration & Index & Rank & Equal-domain & Rank \\
\midrule
GPT 6 Astra & 31.26 & 1 & 32.48 & 1 \\
GPT 5.6 Sol & 20.73 & 2 & 22.74 & 3 \\
Gemini 3.6 Flash & 20.15 & 3 & 20.84 & 5 \\
Qwen 3.8 Max & 20.01 & 4 & 19.67 & 6 \\
Kimi K3 & 19.77 & 5 & 21.53 & 4 \\
DeepSeek V4.1 Flash & 18.35 & 6 & 23.16 & 2 \\
GLM 5.3 Flash & 16.08 & 7 & 19.47 & 7 \\
Seed 2.1 Pro & 12.72 & 8 & 16.83 & 8 \\
MiniMax M3 & 11.74 & 9 & 11.63 & 10 \\
MiMo V2.5 & 11.42 & 10 & 13.53 & 9 \\
\bottomrule
\end{tabular}
\caption{Sensitivity of Agent rankings to domain weights. Within each domain, category scores retain their case-count weights. Values combine the rounded baseline means and added task-level scores; rank changes do not establish statistical significance.}
\label{tab:agent-weight-sensitivity}
\end{table}

GPT 6 Astra ranks first under both mixtures, but several other positions change substantially. DeepSeek V4.1 Flash moves from sixth to second and Kimi K3 from fifth to fourth. The case-weighted Index should therefore be interpreted as performance on the configured inventory rather than a distribution-independent ordering of general Agent ability. Confidence intervals and paired model comparisons require the underlying evaluation records and a resampling unit that accounts for shared templates and the SWE/ToolChoice parent links. In particular, the 200 Social cases instantiate 40 templates and should not automatically be treated as independent observations.

\subsection{Physical Tasks: Goal and Constraint Satisfaction}
\label{app:agent-planning}
\label{app:agent-manipulation}
\label{app:agent-navigation}
\label{app:agent-sequencing}

\begin{figure}[H]
\centering
\includegraphics[width=.94\linewidth,height=.64\textheight,keepaspectratio]{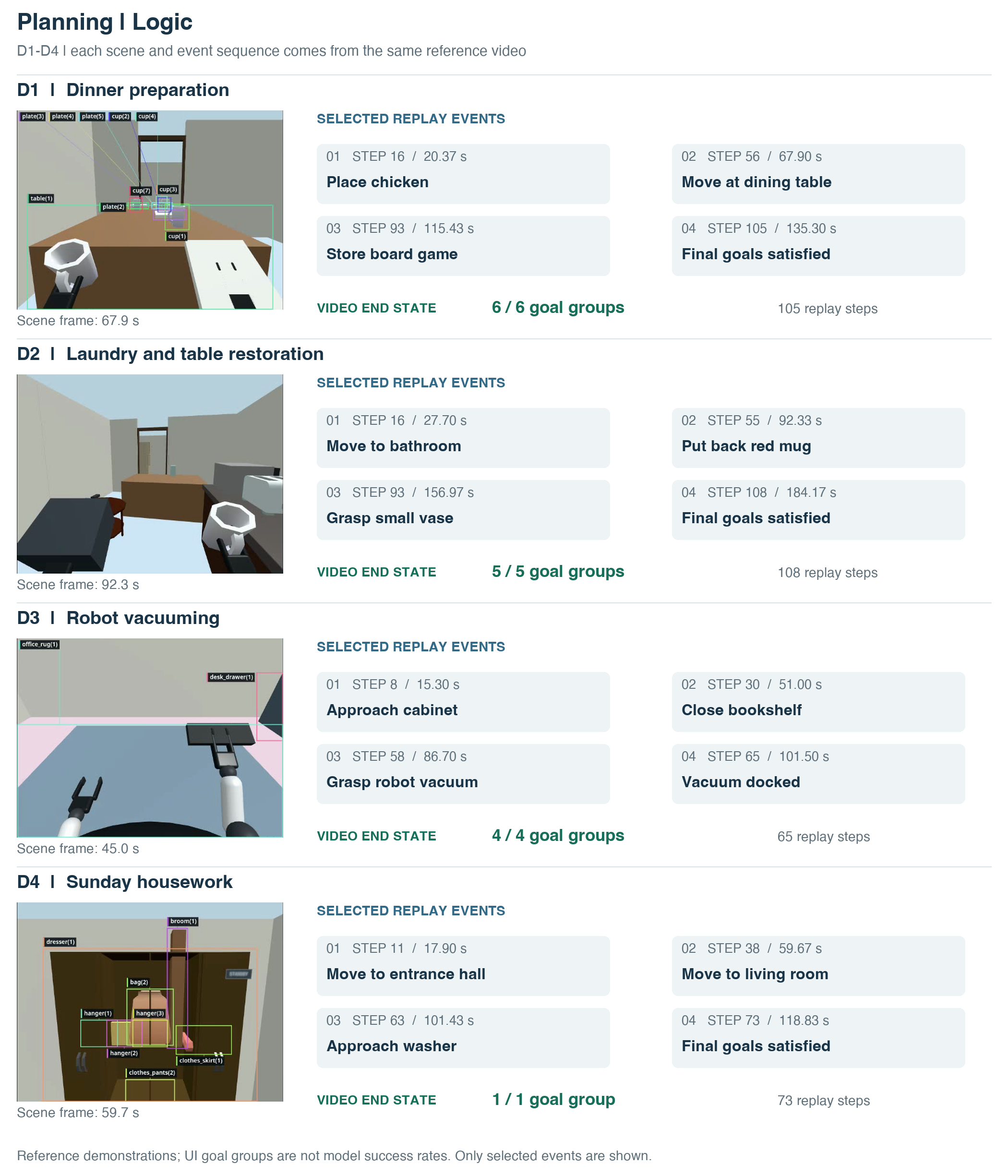}
\caption{\textbf{Planning reference demonstrations.} Four household workflows pair a scene view with selected actions and final goal-group indicators from the same replay. Scene and event timestamps can differ. The counts describe satisfied goal groups, not model success rates; the fourth workflow has one group containing 23 conditions.}
\label{fig:report-planning-demonstrations}
\end{figure}

\paragraph{Planning: satisfy the complete goal.}
Thirty tasks combine household actions with final-state constraints. Figure~\ref{fig:report-planning-demonstrations} shows four reference demonstrations: 105, 108, 65, and 73 steps satisfy six, five, four, and one goal groups, respectively. The last group contains 23 conditions. These demonstrations establish intended behavior, not model success rates.

\begin{figure}[tbp]
\centering
\includegraphics[width=\linewidth,height=.36\textheight,keepaspectratio]{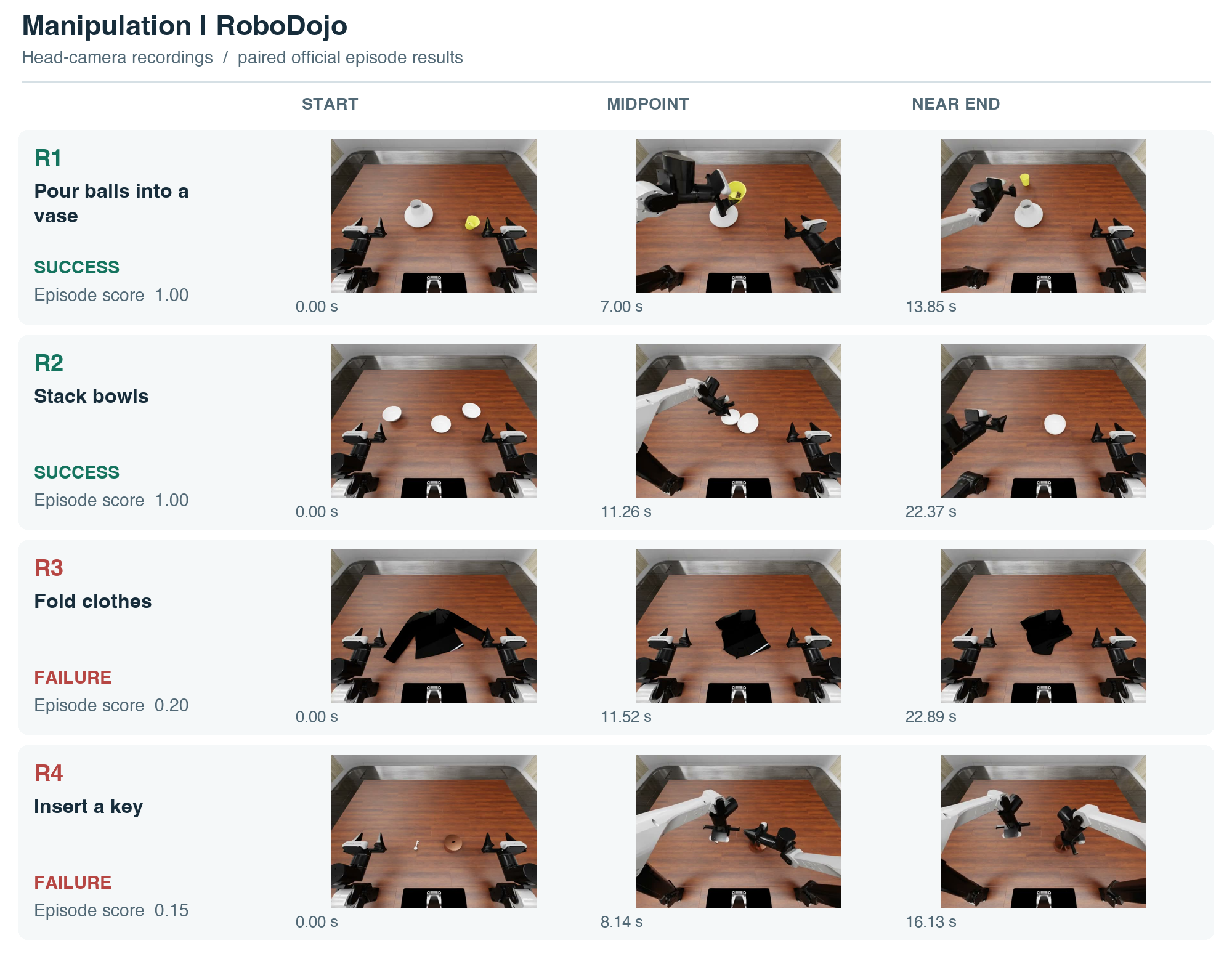}
\caption{Representative manipulation outcomes from the source material. Ball pouring and bowl stacking succeed; cloth folding and key insertion receive partial credit without success. Start, middle, and late frames illustrate behavior, while the recorded evaluator supplies the outcome. Historical controller differences prevent treating this panel as a matched model comparison.}
\label{fig:pw-robodojo}
\end{figure}

\paragraph{Manipulation: progress is not completion.}
Among 42 tasks, selected seed-0 episodes show successful ball pouring and bowl stacking (score 1.0), but failed cloth folding (0.2) and key insertion (0.15). Figure~\ref{fig:pw-robodojo} contrasts these outcomes. The historical InternVLA\_A15 / arx\_x5 controller label does not independently identify its upstream planner.

\begin{figure}[!htbp]
\centering
\includegraphics[width=\linewidth,height=.44\textheight,keepaspectratio]{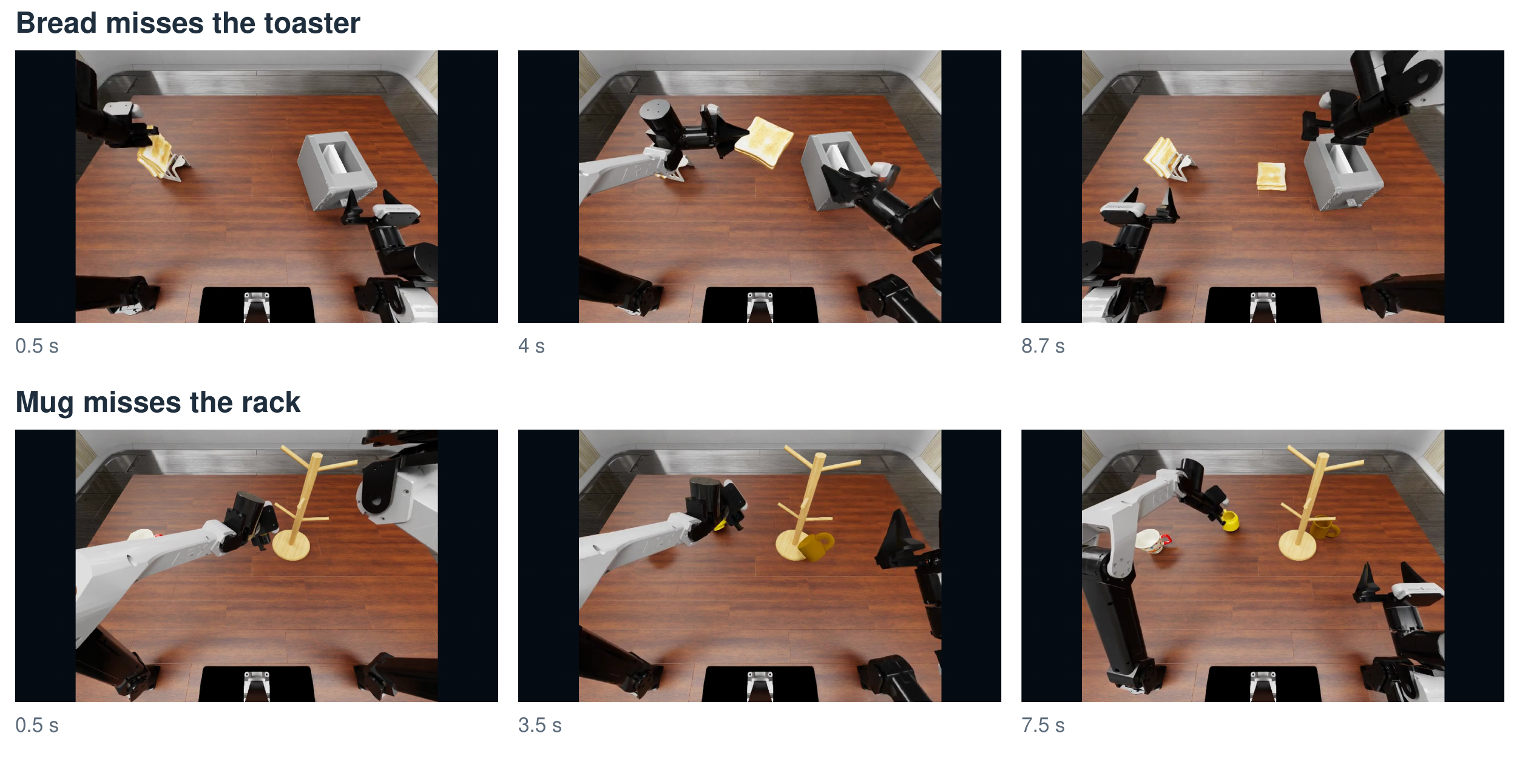}
\caption{\textbf{Two manipulation failures in original robot footage.} Bread lands beside the toaster (top); a mug misses the rack (bottom). Times refer to the supplied clips. The archive labels both InternVLA\_A15 episodes as failures; selected frames do not establish the precise controller fault or a leaderboard failure rate.}
\label{fig:agent-robot-failures}
\end{figure}

\begin{figure}[tbp]
\centering
\includegraphics[width=\linewidth,height=.37\textheight,keepaspectratio]{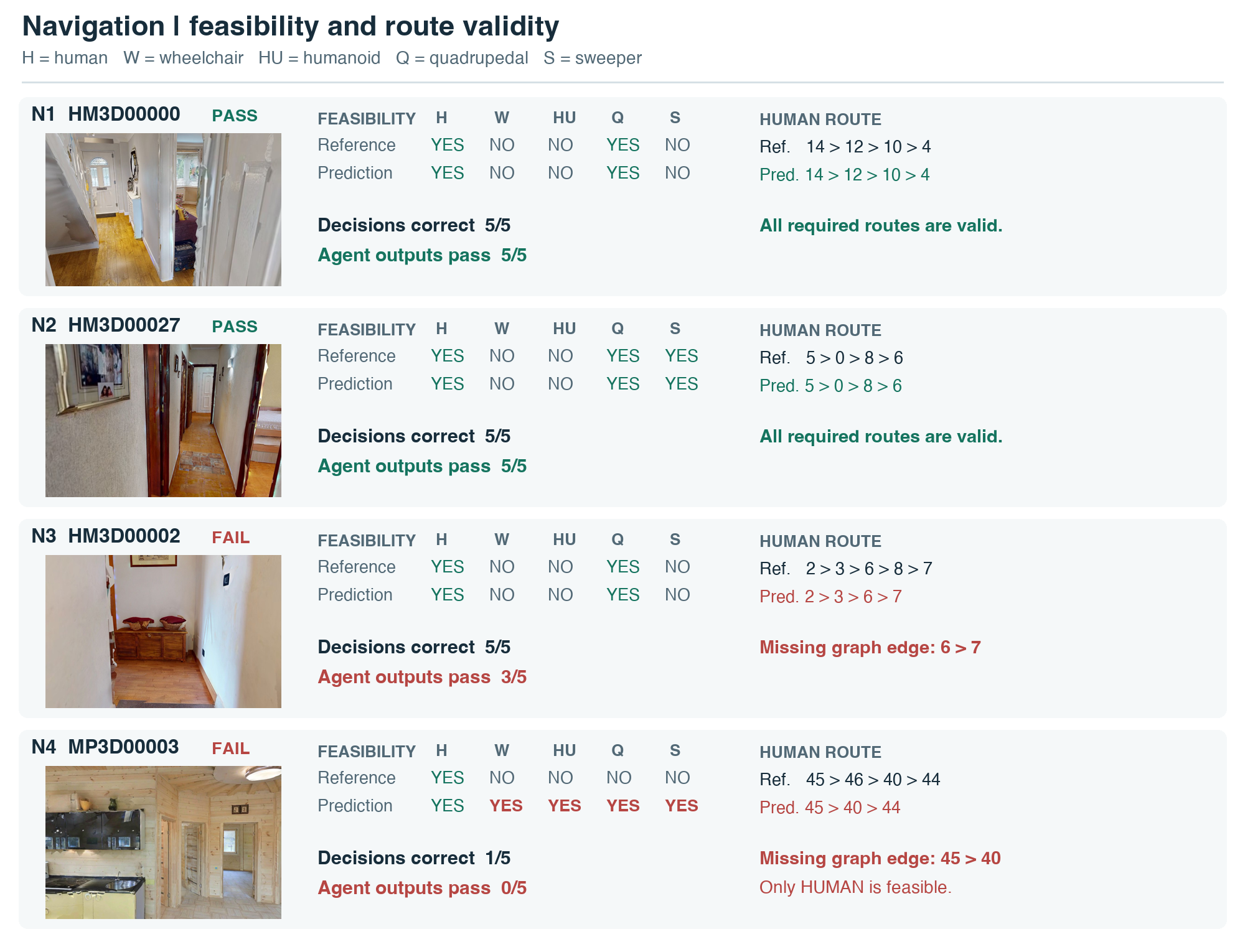}
\caption{Navigation examples distinguish feasibility from valid route construction. A route can fail through a nonexistent edge even when all feasibility judgments are correct. The displayed reference is one valid route, not the only permitted answer; scene images are context rather than exact endpoint views.}
\label{fig:pw-capnav}
\end{figure}

\paragraph{Navigation: provide a valid route.}
Each of 282 questions requires correct feasibility decisions for five agent types and valid traversable routes. In Figure~\ref{fig:pw-capnav}, saved Gemini 3.6 Flash predictions can classify feasibility correctly yet fail through nonexistent edges (6--7 in HM3D00002; 45--40 in MP3D00003). The references are valid examples, not unique routes; scene images provide context. Original API responses are unavailable, so the examples rely on saved predictions checked against task maps.

\begin{figure}[!htbp]
\centering
\includegraphics[width=\linewidth,height=.45\textheight,keepaspectratio]{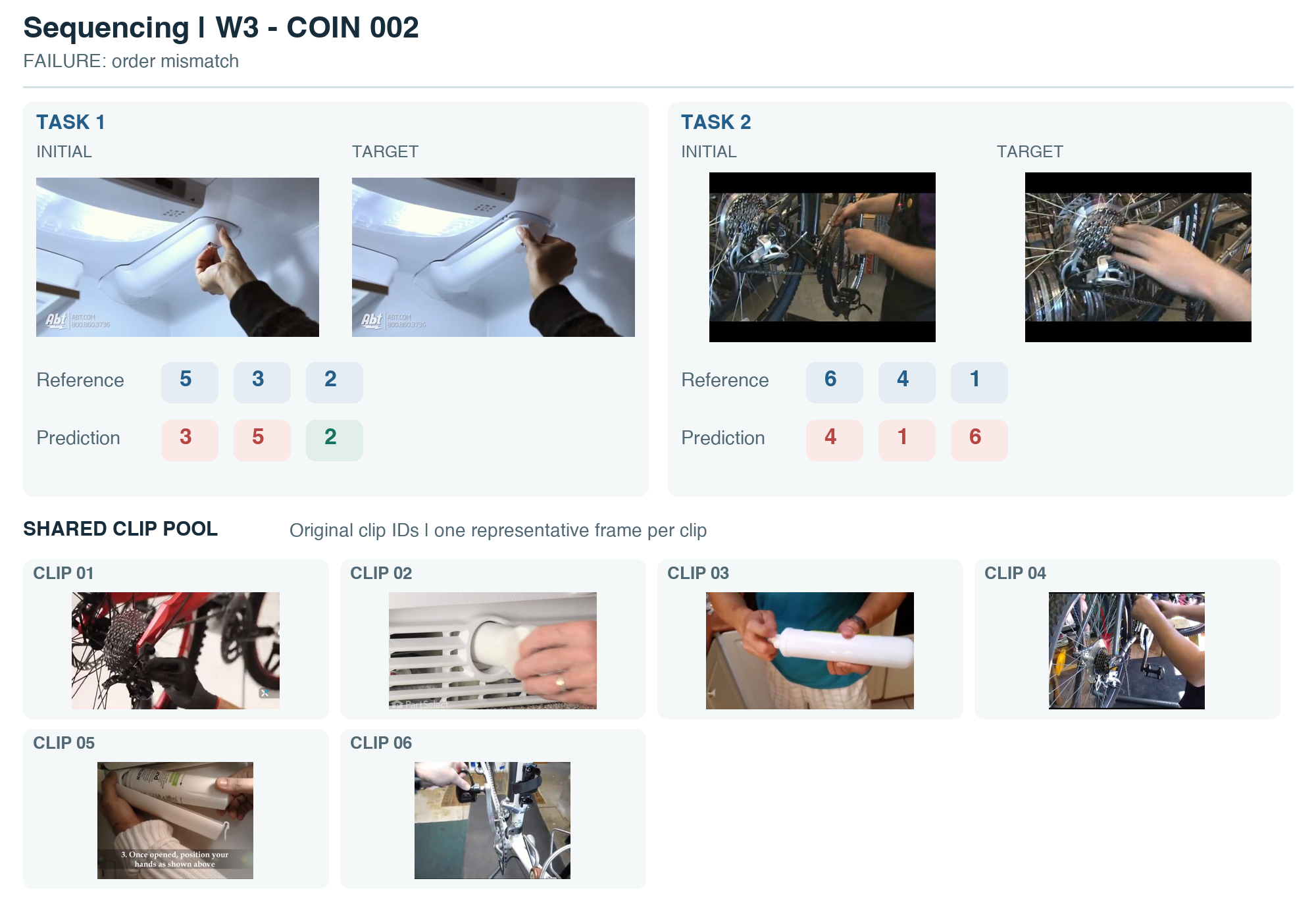}
\caption{\textbf{Correct clip membership, incorrect sequence.} In the historical Qwen3.8-max-preview prediction for COIN pair 002, the predicted orders $[3,5,2]$ and $[4,1,6]$ differ from references $[5,3,2]$ and $[6,4,1]$. The images show task endpoints and one representative frame per clip; they do not measure failure frequency in the main leaderboard.}
\label{fig:report-sequencing-order}
\end{figure}

\paragraph{Sequencing: membership and order must both hold.}
The 170 problems require partitioning shuffled clips into two tasks and reconstructing both reference orders. In the saved Qwen3.8-max-preview COIN 002 case, membership is correct but both orders are wrong (Figure~\ref{fig:report-sequencing-order}). A plausible collection of clips is not an exact reconstruction.

\begin{table}[tbp]
\centering\small
\begin{tabularx}{\linewidth}{@{}lXX@{}}
\toprule
Task family & Locally plausible evidence & Requirement still to verify\\\midrule
Planning & A salient household subgoal is reached & All goal groups and restoration constraints hold together\\
Manipulation & Objects move toward the target & Final arrangement satisfies the official objective\\
Navigation & Destination is judged reachable & Every edge and agent-specific route is valid\\
Sequencing & Relevant clips are recognized & Both assignments and complete orders match\\\bottomrule
\end{tabularx}
\caption{Physical-task lessons grounded in the recorded task rules and selected examples. The rows describe diagnostic distinctions, not measured failure frequencies.}
\label{tab:agent-local-global}
\end{table}

\subsection{Digital Tasks: Verified Interface Outcomes}
\label{app:agent-web}
\label{sec:digital-web-browsing}
\label{app:agent-desktop}
\label{sec:digital-desktop-apps}
\label{app:agent-mobile}
\label{sec:digital-mobile-apps}
\label{app:agent-games}
\label{sec:digital-games}

\paragraph{Web: verify the requested evidence.}
The shopping tasks require 30 qualifying products and three priced purchase pages per product. Their checks separate 30 attribute credits from 90 page credits. A complete-looking list or an accepted evidence snippet does not establish that every requirement is met.

\begin{figure}[!htbp]
\centering
\includegraphics[width=\linewidth,height=.33\textheight,keepaspectratio]{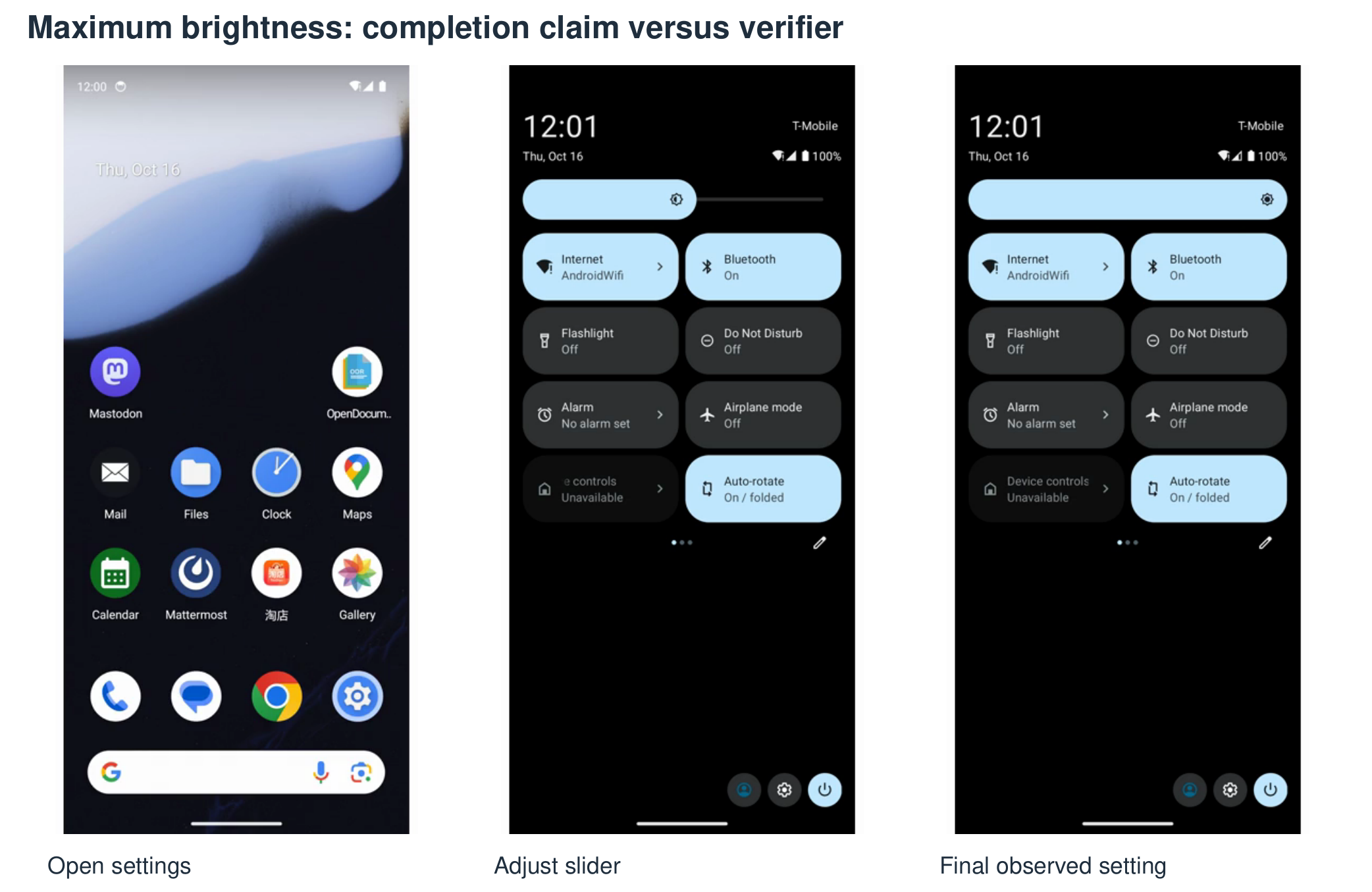}
\caption{\textbf{Claimed maximum brightness, failed verification.} Cropped UI observations from the supplied MobileWorld replay show opening settings, adjusting the slider, and the final screen. The Agent answers that brightness is maximal; the result file records 183/255 and score 0. The supplied evidence does not establish a model identity.}
\label{fig:agent-brightness}
\par\vspace{14pt}
\includegraphics[width=\linewidth,height=.33\textheight,keepaspectratio]{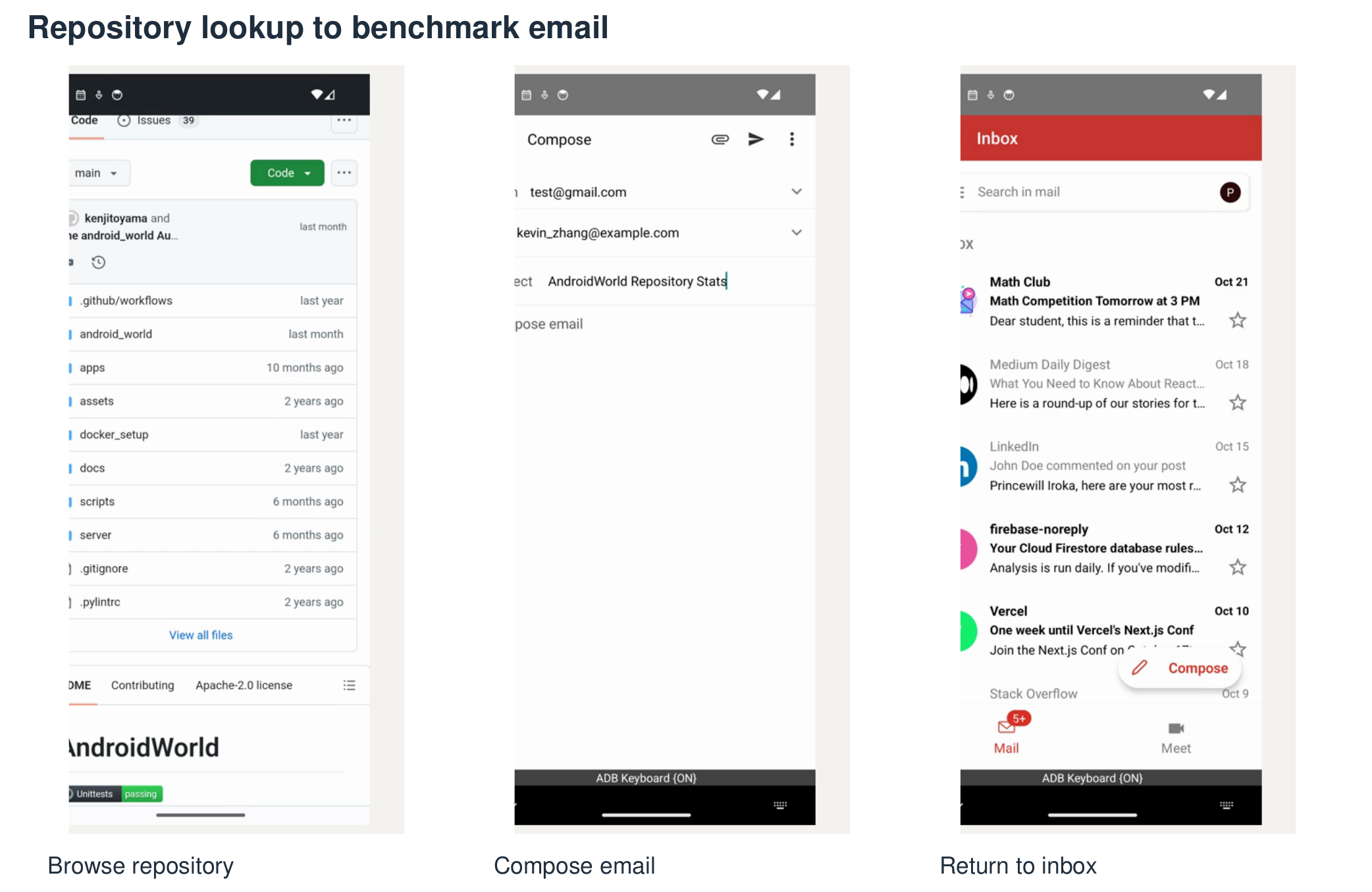}
\caption{\textbf{Repository lookup and benchmark email.} Cropped observations from CheckGithubInfoTask show browsing, composing, and returning to the inbox. The archive labels Gemini 3.6 Flash, contains 40 observations, and records score 1.0. Reported values (854 stars, 15 contributors) belong to that historical run, not current GitHub statistics. This is a saved-observation replay, not continuous screen capture.}
\label{fig:agent-mobile-success}
\end{figure}

\paragraph{Desktop and mobile: verify the final state.}
The inventory retains 95 Desktop and 35 Mobile tasks. A completion message is insufficient: the brightness example claims success but the verifier records 183/255 (Figure~\ref{fig:agent-brightness}). A separate Gemini archive successfully looks up repository statistics and submits a benchmark email (Figure~\ref{fig:agent-mobile-success}).

\begin{figure}[!htbp]
\centering
\begin{tabular}{@{}c@{\hspace{.015\linewidth}}c@{\hspace{.015\linewidth}}c@{}}
\includegraphics[width=.32\linewidth]{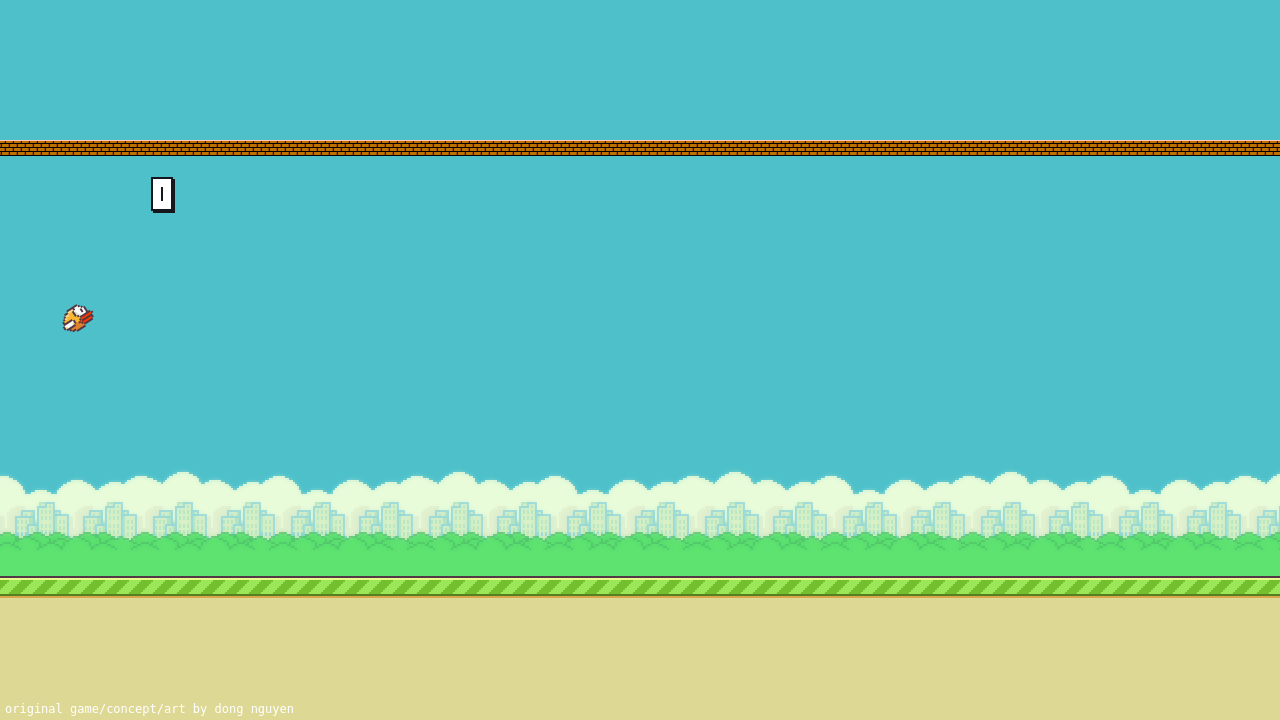} &
\includegraphics[width=.32\linewidth]{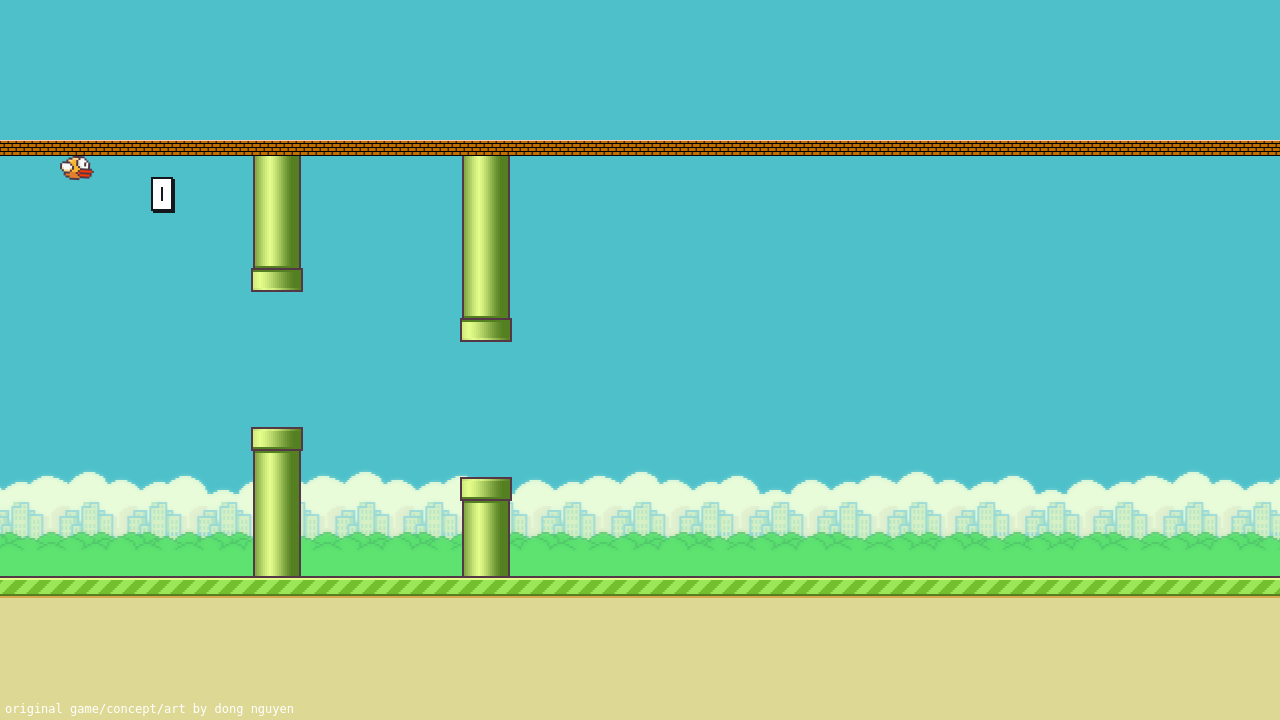} &
\includegraphics[width=.32\linewidth]{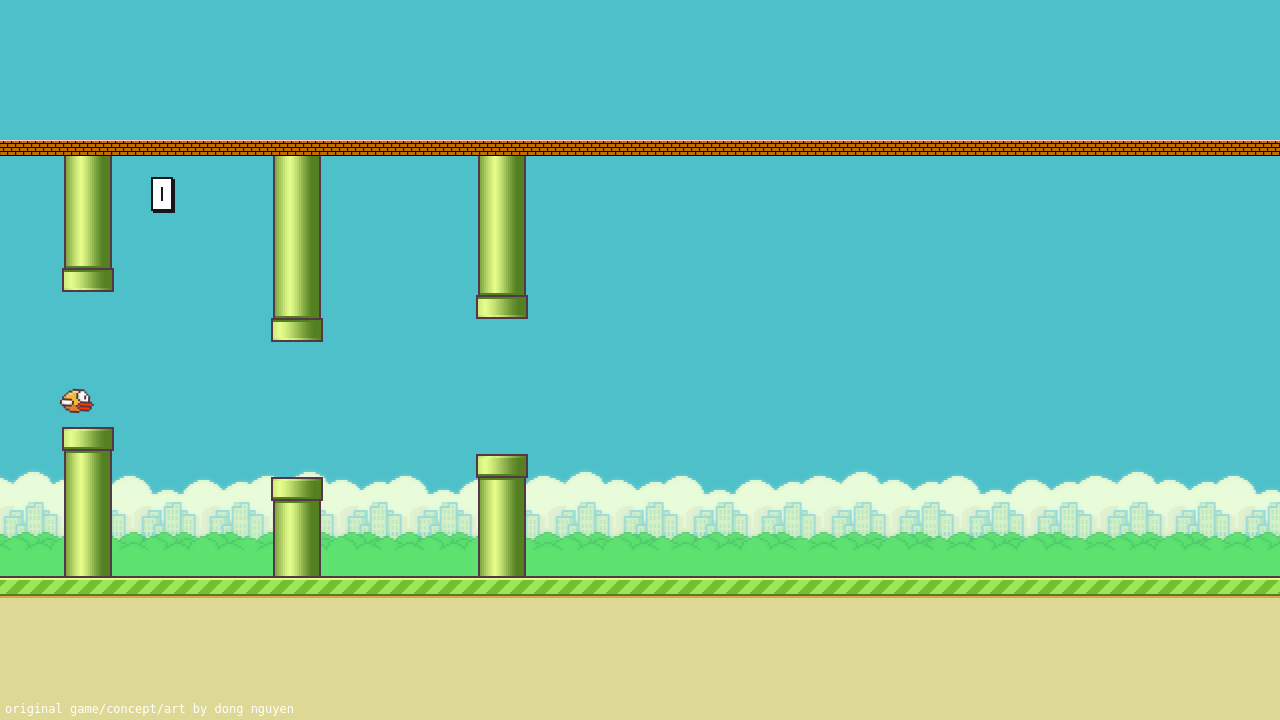}\\
{\small Step 1: flap} & {\small Step 6: wait} & {\small Step 9: flap}
\end{tabular}
\caption{\textbf{Selected observations from the Flappy Bird case.} Frames from the nine-action MiniMax M3 trace show the changing obstacle configuration and the corresponding control choices. These are observations before the labeled actions; completion and score 1 come from the recorded evaluator, not from the last screenshot alone.}
\label{fig:report-game-controls}
\par\vspace{14pt}
\includegraphics[width=\linewidth,height=.30\textheight,keepaspectratio]{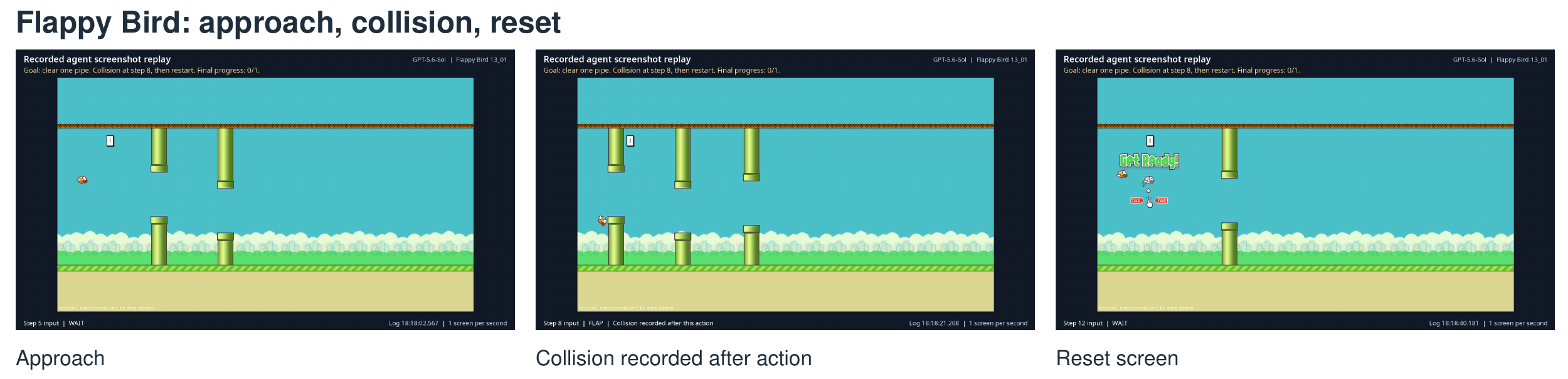}
\caption{\textbf{Approach, collision, and reset in a failed Flappy Bird run.} The supplied GPT-5.6 Sol (high) archive replays saved observations. A collision is logged after the middle action; the later screen returns to Get Ready. The manifest reports 100 actions and no pipe cleared. This is distinct from the successful MiniMax example.}
\label{fig:agent-game-failure}
\end{figure}

\paragraph{Games: inspect the transition as well as the action.}
The MiniMax M3 example clears one pipe after nine valid actions (Figure~\ref{fig:report-game-controls}). A separate GPT-5.6 Sol archive resets after collision and records zero pipes across 100 actions (Figure~\ref{fig:agent-game-failure}). These historical cases illustrate feedback and termination; they are not a matched model comparison.

\begin{figure}[H]
\centering
\includegraphics[width=\linewidth,height=.38\textheight,keepaspectratio]{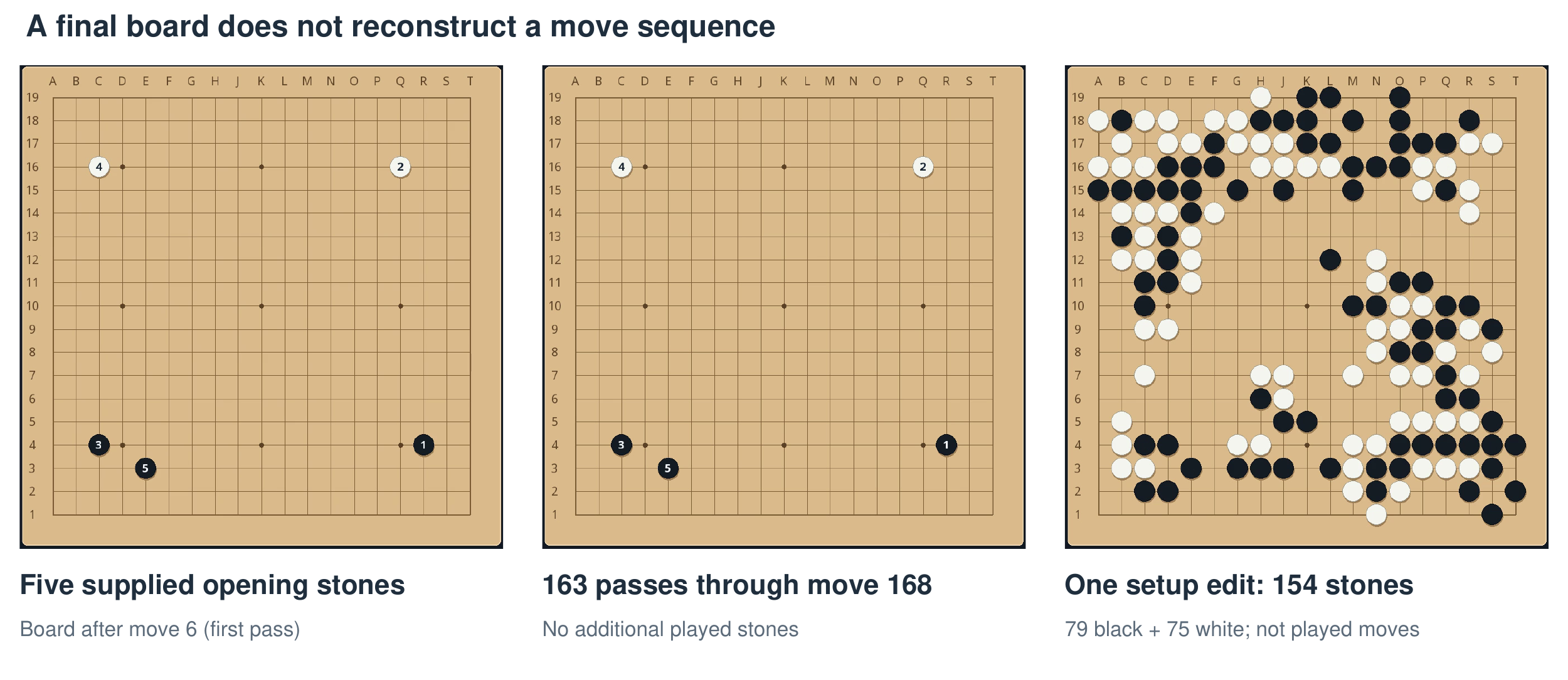}
\caption{\textbf{Submitted SGF omits the move history.} The opening board persists through 163 passes; a separate setup node then clears all 361 points and places 79 black plus 75 white stones. Cropped boards come from deterministic playback of the submitted artifact, not an execution recording or AI reconstruction. This historical example is not an additional leaderboard item.}
\label{fig:agent-go-artifact}
\end{figure}

\FloatBarrier
\paragraph{Reconstruction: an endpoint is not a trajectory.}
The supplied GPT-5.6 Sol Go artifact contains five opening placements, 163 passes, and a setup edit, rather than a reconstructed 168-move game (Figure~\ref{fig:agent-go-artifact}). The source manifest records score 0.

\FloatBarrier
\subsection{Resource-Governance Examples: Groundwater and Forests}
\label{app:social-resource-examples}

\begin{figure}[!htbp]
\centering
\includegraphics[width=\linewidth,height=.43\textheight,keepaspectratio]{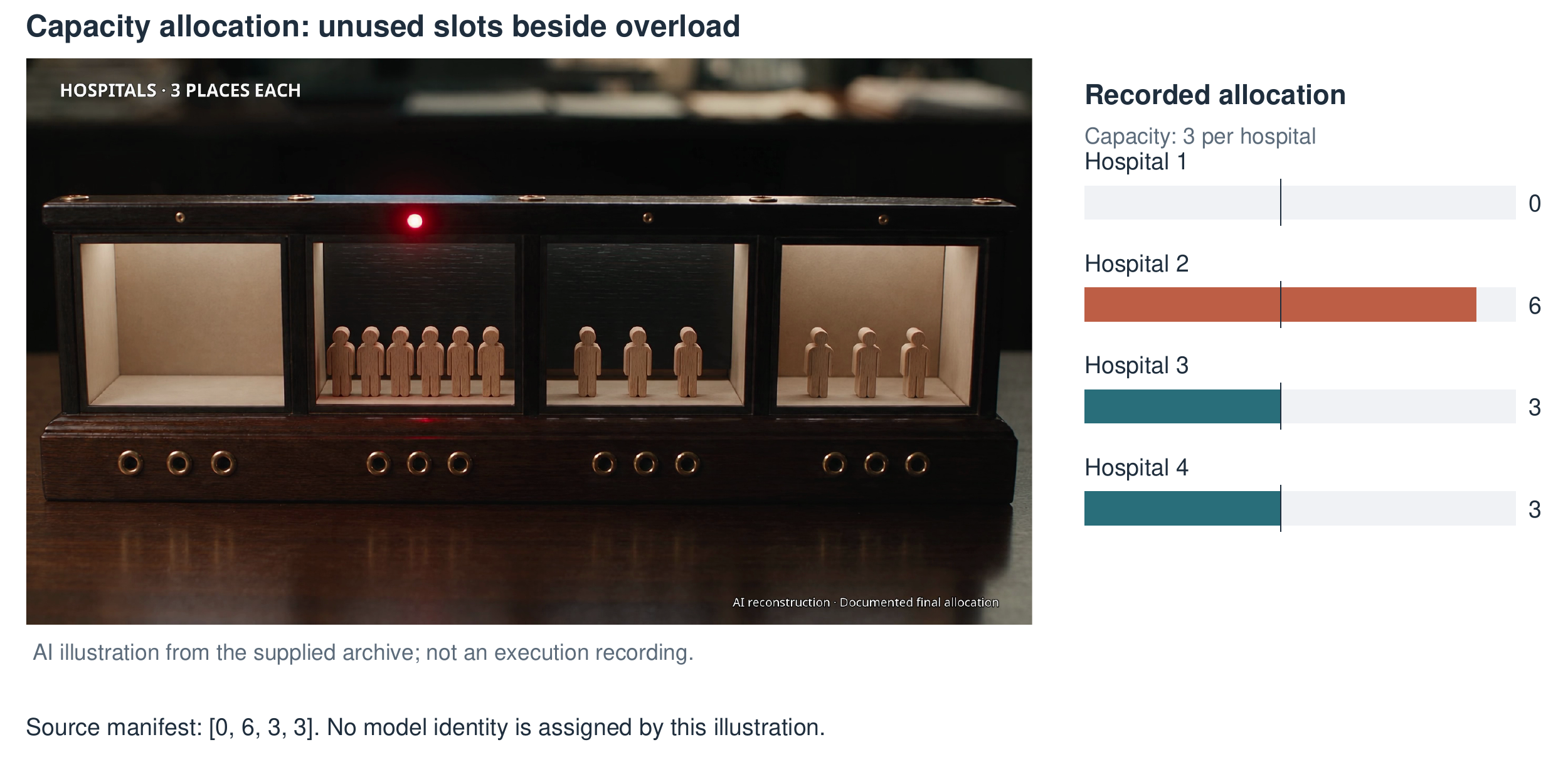}
\caption{\textbf{Unused capacity beside overload.} The source manifest for HOS-F5 records allocation $[0,6,3,3]$ with capacity three per hospital. The left panel is an AI illustration from the supplied package, not an execution recording; the right panel plots the manifest values. No model identity is supplied, and the picture cannot establish the actual action sequence.}
\label{fig:agent-allocation-illustration}
\end{figure}

\paragraph{Inventory and mechanisms.}
Social comprises 200 cases: Coordination \& Strategic Interaction (87), Water, Soil \& Climate Resource Governance (56), and Terrestrial Ecosystem \& Health Resource Governance (57), from 40 templates (17/11/12). Among the 113 resource-labelled cases, 87 use capacity allocation and 26 use resource extraction; scene labels are not independent governance abilities. Figure~\ref{fig:agent-allocation-illustration} visualizes one documented allocation, while Table~\ref{tab:social-resource-traces} records two extraction traces.

\paragraph{Two resource-extraction traces.}
The historical groundwater and forest records retain identifier \texttt{anthropic/claude-fable-5}; no equivalence to a leaderboard configuration is inferred. Each has twelve operators, six rounds, initial stock 100, and sustainable extraction 50 per round. A fixed industrial request of 46 leaves four sustainable units for all participants. First-round participant requests total 129 (groundwater) and 157 (forest), exceeding supply; these are requests, not realized yields. Both logs end round one with zero stock and record no later participant extraction, \texttt{survived=false}, \texttt{constraint\_ok=false}, and raw score 0.05. These simplified games are not validated ecological simulators.

Source identifiers: \texttt{social.commons.aquifer.v0}, strict100 suffix \texttt{018d88e51fdb6e3d}; and \texttt{social.commons.forest.v0}, archive \texttt{case1}. Each contains 72 prompt--response entries.

\begin{table}[H]
\centering\small
\begin{tabularx}{\linewidth}{Xrr}
\toprule
Recorded quantity & Groundwater & Forest \\
\midrule
Initial stock & 100 & 100 \\
Sustainable total per round & 50 & 50 \\
Fixed industrial request per round & 46 & 46 \\
Participant requests in round 1 & 129 & 157 \\
Stock after round 1 & 0 & 0 \\
Participant extraction in each of rounds 2--6 & 0 & 0 \\
Final stock & 0 & 0 \\
Raw recorded score & 0.05 & 0.05 \\
\bottomrule
\end{tabularx}
\caption{Two source-linked resource-extraction failure traces. Groundwater quantities are in megaliters and forest quantities in hectares within the respective game. First-round participant requests are sums of the twelve JSON responses, not realized yields; stock and later extraction follow the recorded state logs. Scores are on the original 0--1 scale and are not category averages.}
\label{tab:social-resource-traces}
\end{table}

\FloatBarrier
\subsection{Scientific Workflows and Verification}
\label{app:agent-tools}
\label{subsec:scientific-tool-selection}
\label{app:agent-long-horizon}
\label{subsec:scientific-long-horizon}

\paragraph{Sources and classification.}
Table~\ref{tab:scientific-capability-counts} maps the 109 Scientific items to their subjects and sources. The local SWE-bench collection~\citep{arxiv231006770} is labeled \texttt{swe-public-30}; its ten derived ToolChoice checkpoints are not an official SWE-bench split. Repair tasks are classified by scientific objective, and each checkpoint inherits its parent assignment. The inventory includes 99 full workflows and ten related decision checkpoints; the latter are not independent workflow trials.

\begin{table}[tbp]
\centering\small
\setlength{\tabcolsep}{5pt}
\begin{tabular}{lrrrrr}
\toprule
Scientific category & ALE & SWE & ToolChoice & Edge & Total \\
\midrule
Mathematics \& Computation & 9 & 2 & 1 & 3 & 15 \\
Health \& Medicine & 21 & 2 & 2 & 0 & 25 \\
Life Sciences & 20 & 9 & 3 & 0 & 32 \\
Physical Sciences & 13 & 17 & 4 & 3 & 37 \\
\midrule
Total & 63 & 30 & 10 & 6 & 109 \\
\bottomrule
\end{tabular}
\caption{Scientific item counts by subject category and source. All 109 items enter the current leaderboard. ToolChoice items are decision checkpoints derived from SWE workflows and inherit the parent subject assignment; source linkage is not an independence claim.}
\label{tab:scientific-capability-counts}
\end{table}

\paragraph{Scoring and execution settings.}
SWE uses recorded binary verifier rewards; ToolChoice requires the prescribed tool, arguments, and call constraints. Each receives 100 for strict success and 0 otherwise. Table~\ref{tab:scientific-source-results} reports accepted outcomes, including valid timeout results; Edge uses the graded mappings below. The retained six configurations use archived Claude Code SWE runs and ToolChoice records. The four-configuration extension uses native Codex for Astra and API execution for GLM, DeepSeek, and MiMo. Context, tools, and budgets are not fully matched, so these are model--harness comparisons.

\paragraph{Specification and provenance limits.}
Six ToolChoice checkpoints (v01, v03, v05, v06, v08, v10) require a \texttt{task\_id} omitted from their prompts. Their archived strict scores remain in the leaderboard; failures cannot be attributed solely to tool selection. The local SWE collection lacks upstream instance IDs, limiting source-level verification. One repair-task title differs from its instruction; both map to Physical Sciences.

The retained six configurations use Claude Code 2.1.240 for SWE and an OpenAI-compatible interface with 11 tools and a 300-second timeout for ToolChoice, with high reasoning, one attempt, and no retries. Infrastructure replacements were used in collecting the 180 accepted SWE records; excluded-attempt totals are unavailable. All 30 Seed runs produced thinking without task tools, and 26 Kimi runs had interrupted tool execution; these outcomes remain scored. The four-configuration extension retains 29 timeouts and one Astra tool-cap termination; numeric SWE caps are not recovered from its summary report. Historical ToolChoice token limits differ from the native Astra interface, and exact provider snapshots are unavailable.

\begin{table}[tbp]
\centering\small
\begin{tabular}{lrr}
\toprule
Configuration & SWE strict passes / 30 & ToolChoice strict passes / 10 \\
\midrule
GPT-6 Astra & 14 & 2 \\
GPT-5.6 Sol & 10 & 1 \\
Gemini 3.6 Flash & 5 & 2 \\
Qwen 3.8 Max & 7 & 2 \\
Kimi K3 & 0 & 2 \\
DeepSeek V4.1 Flash & 8 & 2 \\
GLM 5.3 Flash & 8 & 1 \\
Seed 2.1 Pro & 0 & 2 \\
MiniMax M3 & 3 & 2 \\
MiMo V2.5 & 4 & 2 \\
\bottomrule
\end{tabular}
\caption{Strict outcomes on the 40 added SWE/ToolChoice items. All 10 configurations have 30 SWE and ten ToolChoice records (400 task--configuration observations). Harness and specification limitations described above apply.}
\label{tab:scientific-source-results}
\end{table}

\begin{figure}[!htbp]
\centering
\includegraphics[width=\linewidth,height=.66\textheight,keepaspectratio]{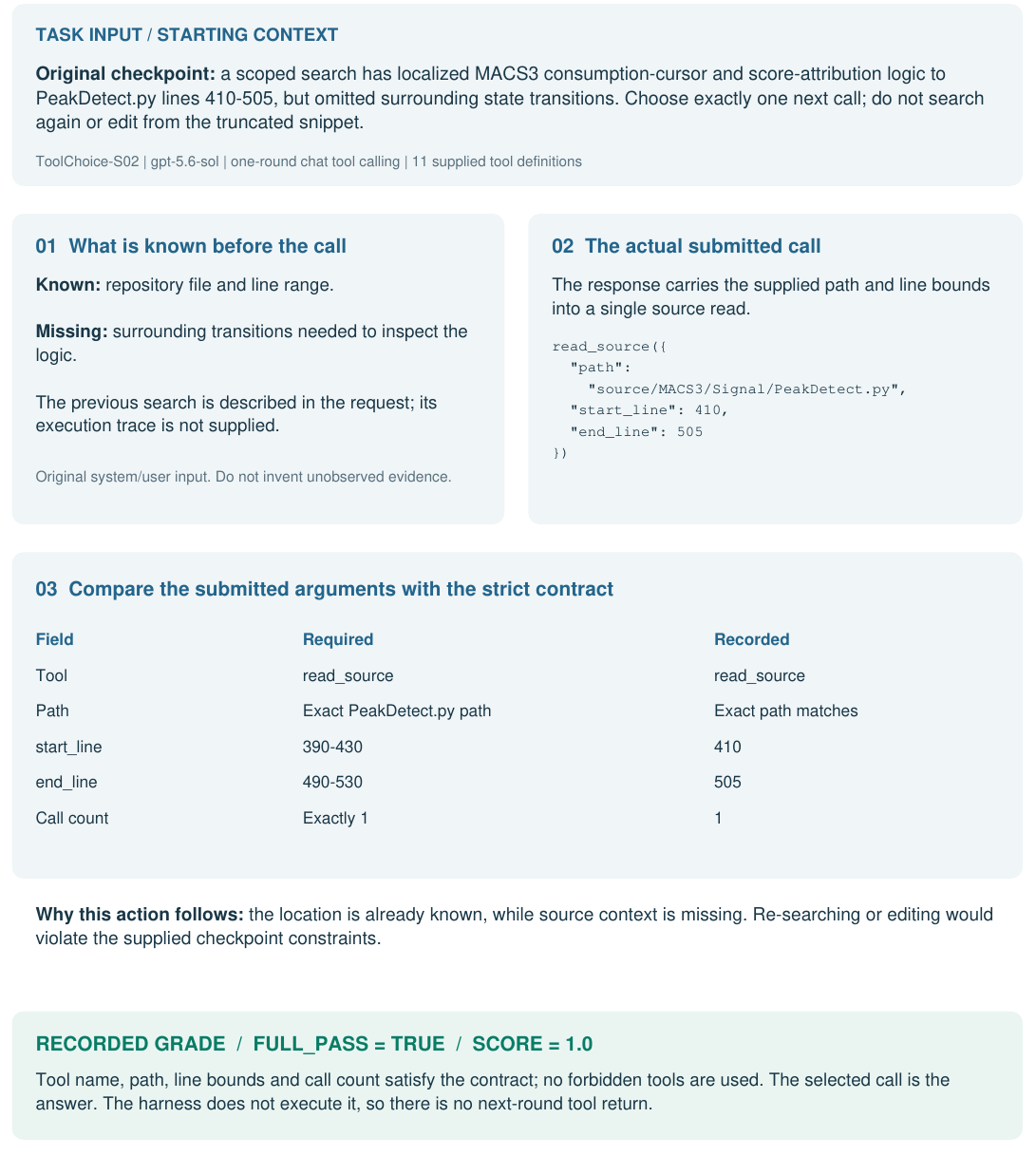}
\caption{\textbf{ToolChoice-S02: verifying a proposed next action.} The source context, GPT-5.6 Sol call, and strict field checks are displayed together. The score of 1.0 applies to this decision checkpoint; the selected tool is not executed and no completed repair is claimed.}
\label{fig:report-toolchoice-evidence}
\end{figure}

\paragraph{Tool choice: selecting is not executing.}
ToolChoice-S02 asks for one \texttt{read\_source} call. GPT-5.6 Sol matches the path, bounds, and call constraints for score 1.0 (Figure~\ref{fig:report-toolchoice-evidence}); the selected call is not executed.

\begin{figure}[!htbp]
\centering
\includegraphics[width=\linewidth,height=.66\textheight,keepaspectratio]{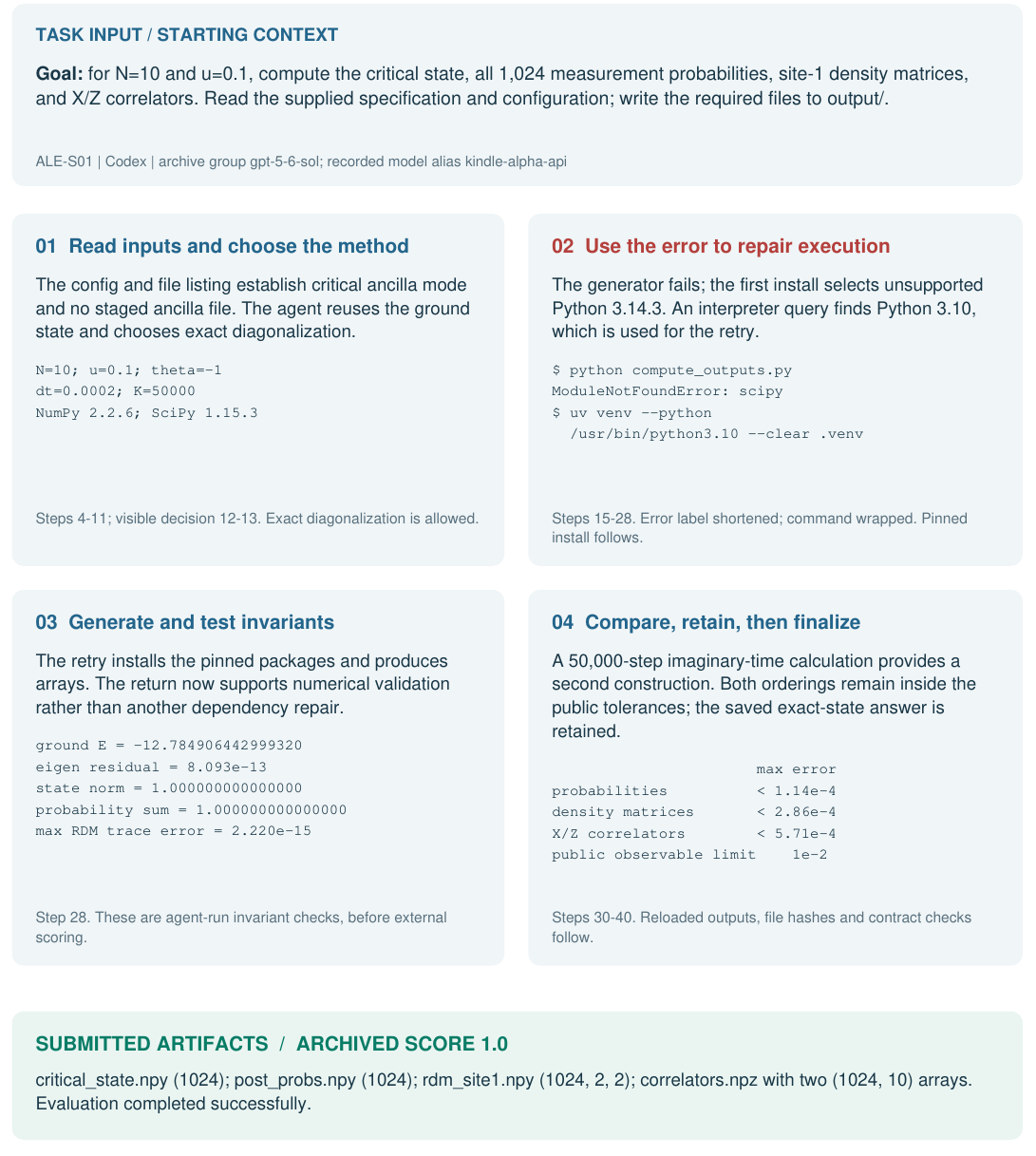}
\caption{\textbf{ALE-S01: numerical workflow and evaluated artifacts.} The historical 41-step trace shows method selection, dependency recovery, numerical checks, and the four retained output files. The archived evaluation assigns 1.0. Shared helper code limits independence of the second calculation; the panel's source identifiers distinguish the execution framework, API alias, and archive model group.}
\label{fig:report-ale-evidence}
\end{figure}

\paragraph{Numerical science: check the submitted artifacts.}
The ten-qubit Ising workflow produces four files and score 1.0 (Figure~\ref{fig:report-ale-evidence}). Its residual is $8.093\times10^{-13}$; a second calculation differs in observables by less than $5.71\times10^{-4}$ against tolerance $10^{-2}$. Shared helper code makes this a consistency check, not independent replication.

\begin{figure}[!htbp]
\centering
\includegraphics[width=\linewidth,height=.66\textheight,keepaspectratio]{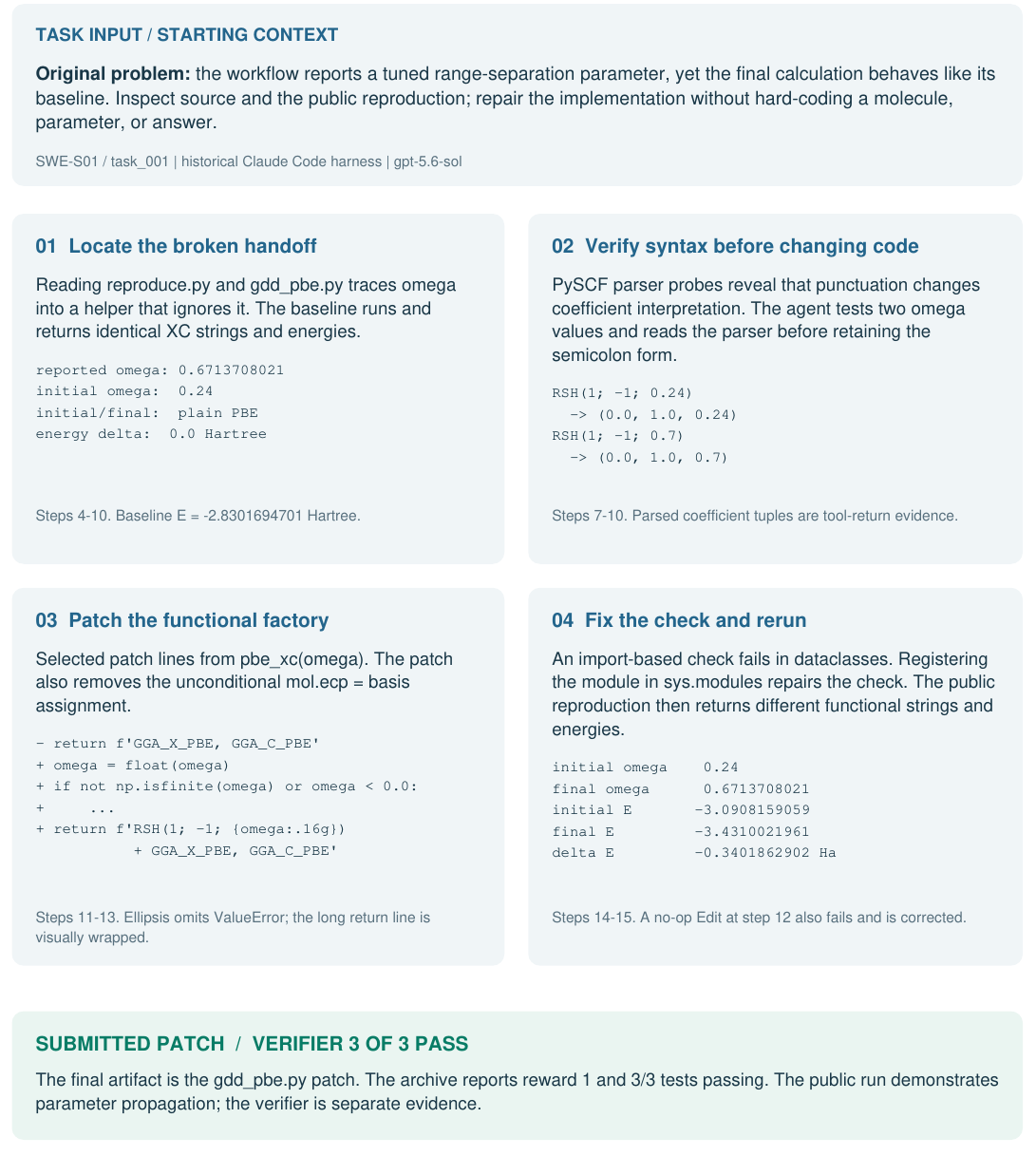}
\caption{\textbf{SWE-S01: making a parameter affect the calculation.} The historical GPT-5.6 Sol / Claude Code trace connects source inspection and parser checks to the patch and changed output. The external verifier reports three passing tests. Omitted code and wrapped lines are marked; passing these tests establishes the specified software behavior, not improved physical accuracy.}
\label{fig:report-swe-evidence}
\end{figure}

\begin{samepage}
\paragraph{Code repair: validate the specified behavior.}
The DFT patch propagates an ignored parameter and rejects invalid values (Figure~\ref{fig:report-swe-evidence}). Three tests pass; energy changes by $-0.34018629$ Hartree. This verifies software behavior, not physical accuracy.
\end{samepage}

\begin{table}[tbp]
\centering\small
\begin{tabularx}{\linewidth}{@{}lXX@{}}
\toprule
Evidence & Supported conclusion & Conclusion not established\\\midrule
Tool-choice fields match & Required next action selected & Tool executed or repair completed\\
Residual and cross-check agree & Numerical consistency & Independent replication\\
Three repair tests pass & Specified behavior corrected & Better physical accuracy\\
\bottomrule
\end{tabularx}
\caption{Scientific examples demonstrate distinct levels of verification. Scores from these workflows are meaningful only with their associated unit and evaluator.}
\label{tab:scientific-evidence-levels}
\end{table}

\FloatBarrier
\subsection{Edge Task Aggregation and Raw Scores}
\label{app:edge-aggregation}
We include six tasks from EdgeBench~\citep{zhu2026edgebench}. Wireless Electricity Layout, BipedalWalker Locomotion RL, and Graph Node Classification are assigned to Mathematics \& Computation; Borden Source Inversion, Molecular Self Assembly, and DABIC Gravity Inversion are assigned to Physical Sciences. These primary reporting assignments do not establish disciplinary exclusivity.

\paragraph{Dual Exam score convention.}
Wireless, Molecular, Graph, and BipedalWalker retain their archived mappings to 0--100. Borden and DABIC have no separate rescaling in the public task configurations. We clip their raw scores below 0 to 0 and above 100 to 100, leaving values within that range unchanged. This is a \textbf{Dual Exam aggregation convention, not an official EdgeBench normalization rule}. It bounds each item's contribution without equating task difficulty. Table~\ref{tab:edge-raw} retains the raw values; Astra's Borden score of 100.493 contributes 100. The reason for the raw value exceeding 100 remains unresolved.

\paragraph{Budget and evidence.}
We use the best recorded valid submission within an allowance of 43,200 seconds in the selected archived run, including valid submissions before early termination. This is our budget-limited aggregation, not a claim that every run completed the full EdgeBench horizon. Confirmed absence of a valid submission receives zero; a missing record is not automatically zero. The retained six configurations use their selected frozen run IDs. The additional four use the supplied result summary, with all four Wireless endpoints additionally checked against their best-submission judge reports. DeepSeek and MiMo Wireless stopped at model context/resume limits, but both passed 50/50 tests; their normalized scores are 19.9172 and 0, respectively. The latter is a mapped quality score, not missing-data imputation. MiniMax Wireless has conflicting checkpoint-validity and final pass-rate fields; both interpretations yield zero under the mapping. The transferred trace archive is truncated, so the other added-task endpoints and evaluator-version equivalence have not all been independently verified. Astra uses a native Codex executor; these results compare model--harness configurations, not matched model-only capability.

\begin{table}[tbp]
\centering\scriptsize
\setlength{\tabcolsep}{4pt}
\begin{tabular}{lrrrrrr}
\toprule
Configuration & Wireless & Borden & Molecular & Bipedal & Graph & DABIC \\
\midrule
GPT-6 Astra & 1.30310e+15 & 100.493 & 3.20779e+08 & 28.34 & 0.85042 & 15.877115 \\
GPT-5.6 Sol & 3.43635e+15 & 34.69 & 2.36472e+08 & 22.76 & 0.704273 & 16.797531 \\
Gemini 3.6 Flash & 3.03795e+16 & 7.224 & 1.73814e+08 & 17.79 & 0.513631 & 17.460083 \\
Qwen 3.8 Max & 3.45418e+15 & 7.298 & 2.56779e+08 & 20.35 & 0.414559 & 15.042482 \\
Kimi K3 & 3.08015e+15 & 75.276 & 1.53309e+08 & 22.33 & 0.669031 & 9.45664 \\
DeepSeek V4.1 Flash & 2.46882e+15 & 50.646 & 2.85375e+08 & 30 & 0.610576 & 17.480643 \\
GLM 5.3 Flash & 1.51011e+16 & 41.463 & 2.13459e+08 & 29.41 & 0.500476 & 15.929143 \\
Seed 2.1 Pro & 6.63101e+17 & 7.432 & 6.92801e+07 & 2.5 & 0.284502 & 8.837336 \\
MiniMax M3 & 1.00000e+17 & 7.048 & 1.86909e+08 & 13.6 & 0.369085 & -- \\
MiMo V2.5 & 6.41351e+16 & 38.967 & 1.48430e+08 & 19.83 & 0.596243 & 8.5838 \\
\bottomrule
\end{tabular}
\caption{Raw Edge objectives before Dual Exam aggregation. Wireless is minimized; the other five objectives are maximized. Large objectives are abbreviated in scientific notation; exact source values are retained in the audit files. A dash denotes no recorded valid score (MiniMax DABIC), assigned zero in our aggregation. This table reports objective values, not pass rates; early-terminated valid results are retained under the stated budget rule.}
\label{tab:edge-raw}
\end{table}

\subsection{Evidence Summary}
Across these examples, local progress can fail a final constraint, a completion claim can disagree with the verifier, and a plausible endpoint can omit the required process. The figures support these concrete distinctions. They do not estimate failure frequencies, isolate model-only ability, or establish which intervention improves leaderboard performance.

\section{World Model Analysis: Action Effects and State Preservation}
\label{app:world-model-evaluation}
\label{app:world-model-visual}

Six case panels examine requested action effects, preserved state, and rule adherence under fixed interventions. They illustrate errors that aggregate scenario scores cannot distinguish; their source and observation limits are specified below.

\subsection{Sensitivity to the Scenario Mixture}

The full-coverage ordering in Table~\ref{tab:world-model-results} depends on the scenario mixture. MiniMax H3 outperforms GPT-Image-2 on Embodiment (53.8 versus 20.9), but has a lower Overall (17.3 versus 21.7). A partner Agent concentrated on embodied tasks may therefore require a different weighting from the benchmark's eight-scenario mean.

If Embodiment receives more than about 22.9\% of the weight and the other seven scenarios share the remainder equally, MiniMax H3 overtakes GPT-Image-2. This calculation uses rounded scenario scores and unchanged within-scenario mixtures; it is a weighting sensitivity result, not a threshold for partner suitability. Digital scores remain low (Web at most 16.5; Game at most 13.3), subject to task and rubric differences.

\subsection{Case Evidence and Observation Limits}

Each panel shows six checkpoints from the complete instructed history. Columns align instructed steps; rows track four image and eleven video configurations. The reference is an initial scene, not a ground-truth trajectory. Historical labels are preserved: the first four cases use Seedream 5.0 Pro, the interface cases use Seedream 5.0; sources also include physical Yume \texttt{runs\_hard\_v2}, pendulum/mentorship \texttt{yume-5b}, and September 2026 Kling reruns. These batches need not match the main leaderboard settings.

Selected checkpoints can reveal visible discrepancies but cannot establish continuous-time behavior, failure prevalence, or a controlled model comparison. Table~\ref{tab:wm-transition-checks} separates each case's requested effect, invariant, and evidence requirement. Similarity alone is insufficient: a copied frame may omit an action, while an altered frame may realize it and corrupt unrelated state.

\begin{table}[tbp]
\centering\small
\begin{tabularx}{\linewidth}{@{}lXXX@{}}
\toprule
Case & Requested effect & State to preserve & Evidence limit\\\midrule
Pendulum & Evolve under edited gravity & Bob, rod, support & Period needs video\\
Mentorship & Named person's pose/gaze & Identity and other attributes & Timing needs video\\
Cup manipulation & Assigned arm actions & Cup count and roles & Concurrency needs video\\
Dual trays & Transfer and rule changes & Jar, trays, prior state & Final image is not completion\\
Minesweeper & Edit specified cells & Other cells and flags & Layout alone is insufficient\\
Shopping & Query and filter operations & Earlier selections & Imagery is not UI state\\\bottomrule
\end{tabularx}
\caption{A common reading of the World Model cases: desired change, invariant, and observation needed for verification. These are case-specific checks rather than independent quantitative metrics.}
\label{tab:wm-transition-checks}
\end{table}

\subsection{Rule Changes and Delayed Effects}

The pendulum case changes Earth gravity to lunar gravity at step 4 without requesting immediate motion. Steps 5--6 require evolution under the new rule, and step 7 restores the view. The copper bob, rod, and support must persist. Alaya's final checkpoint no longer shows the original apparatus, whereas Qwen-Image-3.0 restores its view. Neither static observation verifies the oscillation period.

Immediate stability after a rule edit can be valid; compliance must also be checked during later motion. Yume's step-4 checkpoint is carried forward and provides no new evidence of a response.

\subsection{Participant-Specific State}

In the painting-studio case, A is the standing mentor and B the seated student. A establishes gaze toward B and raises a hand; B later nods. At step 9, A must lower the hand while retaining the gaze. The new instruction changes one attribute without resetting the others.

Changes in framing or appearance make participant-specific checks ambiguous in several rows. The displayed states support checks of visible identity and pose, but response timing requires video. A socially plausible image can still assign the action to the wrong person; uncertain gaze or identity cannot be treated as verified compliance.

\begin{figure}[!tbp]
\centering
\includegraphics[width=\linewidth,height=.79\textheight,keepaspectratio]{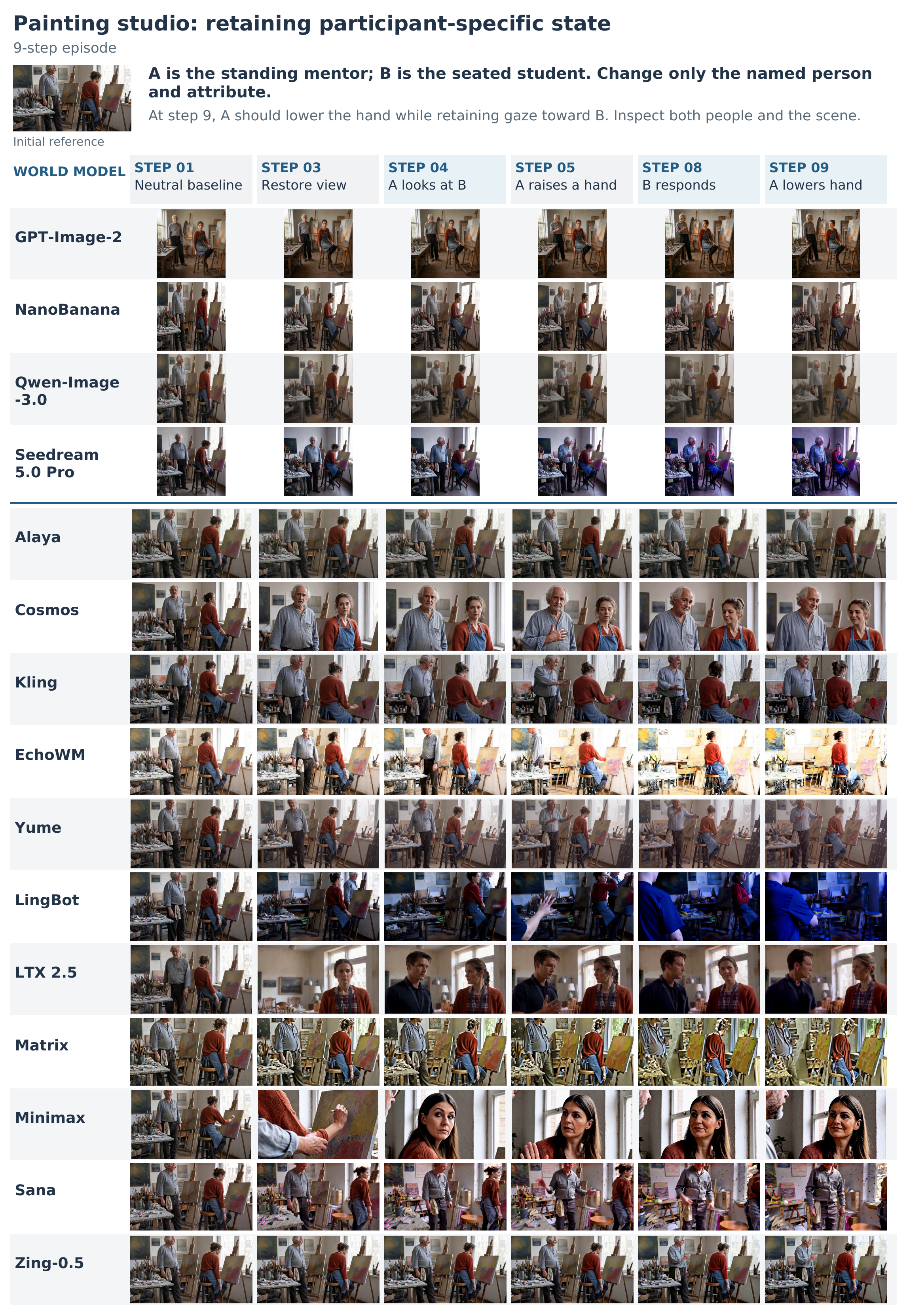}
\caption{\textbf{Painting-studio mentorship: entity-specific updates.} Steps 1, 3, 4, 5, 8, and 9 show baseline, restored view, mentor gaze/hand updates, student response, and lowering of the mentor's hand while gaze persists. The panel supports state checks, not within-video timing claims.}
\label{fig:wm-visual-mentorship}
\end{figure}

\subsection{Coordination and State Across Revisits}

\paragraph{Cup manipulation.}
Two arms manipulate three cups through concurrent motion, nesting, hand-over, and role exchange. The step-7 role change must persist into step 8, when the right arm waits. EchoWM and Yume introduce human hands, and Qwen-Image-3.0 adds cups; MiniMax H3 retains three cups at step 10. The first two observations illustrate entity and inventory changes, while retained cup count verifies only one task constraint. A correct final arrangement would still not verify arm assignment or within-turn concurrency, which require temporal evidence.

\paragraph{Dual trays.}
The 20-step task requires moving a glass jar to the blue tray and aligning the trays while retaining state through rule, role, and viewpoint changes. Steps 7 and 10 bracket a visit to another bench; steps 5 and 18 edit rules without requesting immediate motion. At step 10, MiniMax H3 introduces a red tray, whereas GPT-Image-2 retains the original white and blue trays. Yume's final checkpoint shows a wooden bench. Tray appearance alone does not verify jar placement, alignment, or rule compliance.

Revisits test whether a workspace agrees with its earlier state plus authorized changes. Adjacent-frame similarity cannot establish this persistence. The observed drift does not, however, identify memory as its sole cause; generation and instruction interpretation may also contribute.

\subsection{Exact Updates in Digital Interfaces}

\paragraph{Minesweeper.}
Actions flag g1 and i1, reveal e5, then flag g2, h2, and i2. The omitted second step flags h1, which must also persist in later checkpoints. Checks therefore use the complete seven-step history rather than only the six displayed actions. LingBot replaces the grid with a different graphic. Less conspicuous errors can preserve the layout while omitting a requested flag or erasing a previous one. Validity requires both the target edit and unchanged non-target cells.

\begin{figure}[!tbp]
\centering
\includegraphics[width=\linewidth,height=.79\textheight,keepaspectratio]{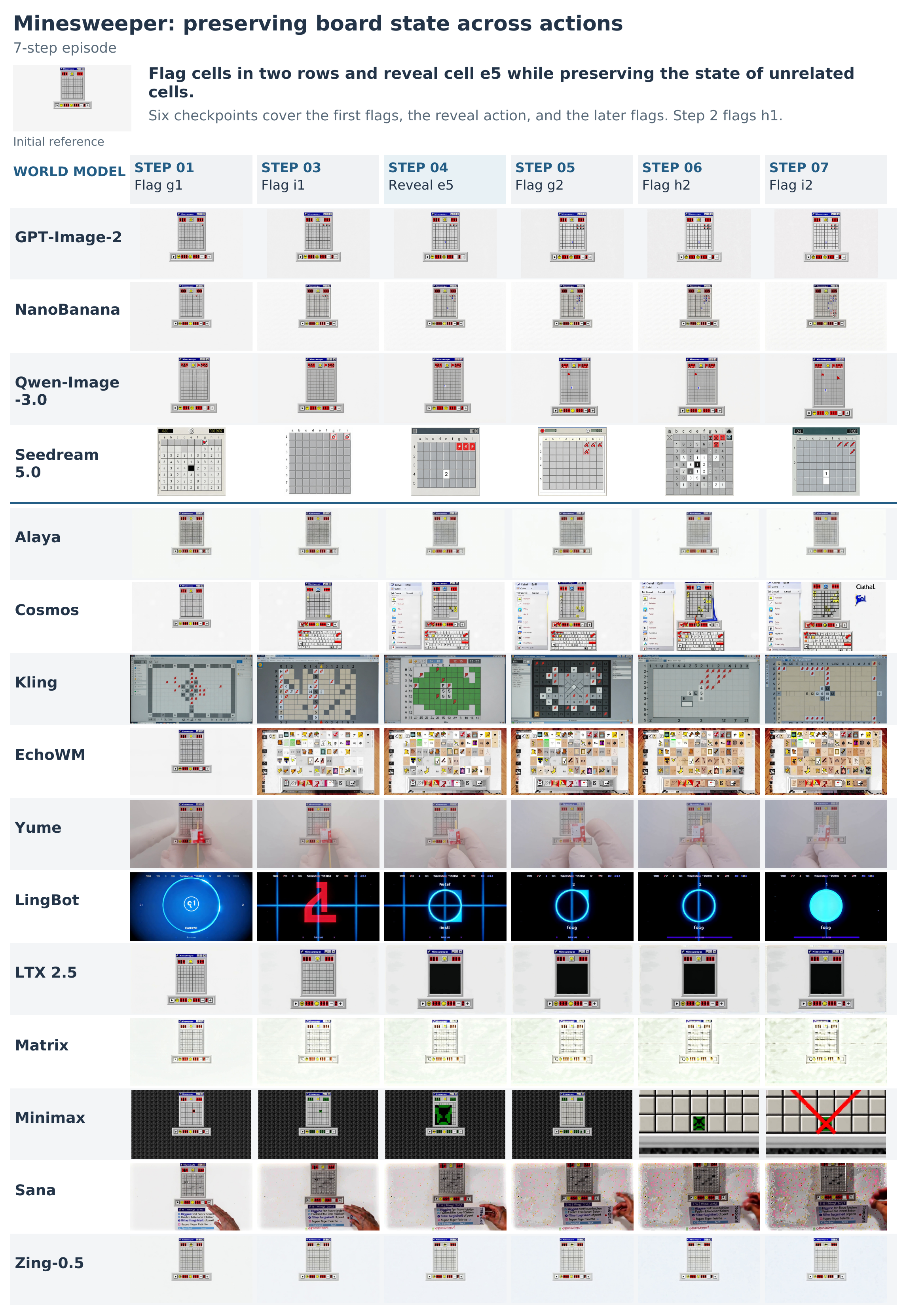}
\caption{\textbf{Minesweeper: exact cell edits and persistent board state.} Steps 1, 3, 4, 5, 6, and 7 show first-row flags, revealing e5, and later second-row flags. Omitted step 2 flags h1 and remains in the history. Geometry preservation alone does not establish cell-level correctness.}
\label{fig:wm-visual-minesweeper}
\end{figure}

\paragraph{Shopping.}
The six instructions enter and select the query Google Pixel 7, enable Buy on Google, choose 128 GB and Snow, and open the sorting menu. Each action must preserve previous selections and use a visible, actionable control. MiniMax ultimately replaces the page with a dialog. Plausible product imagery cannot verify query contents, filters, or menu state, although these small regions determine the meaning of subsequent actions.

\begin{figure}[!tbp]
\centering
\includegraphics[width=\linewidth,height=.79\textheight,keepaspectratio]{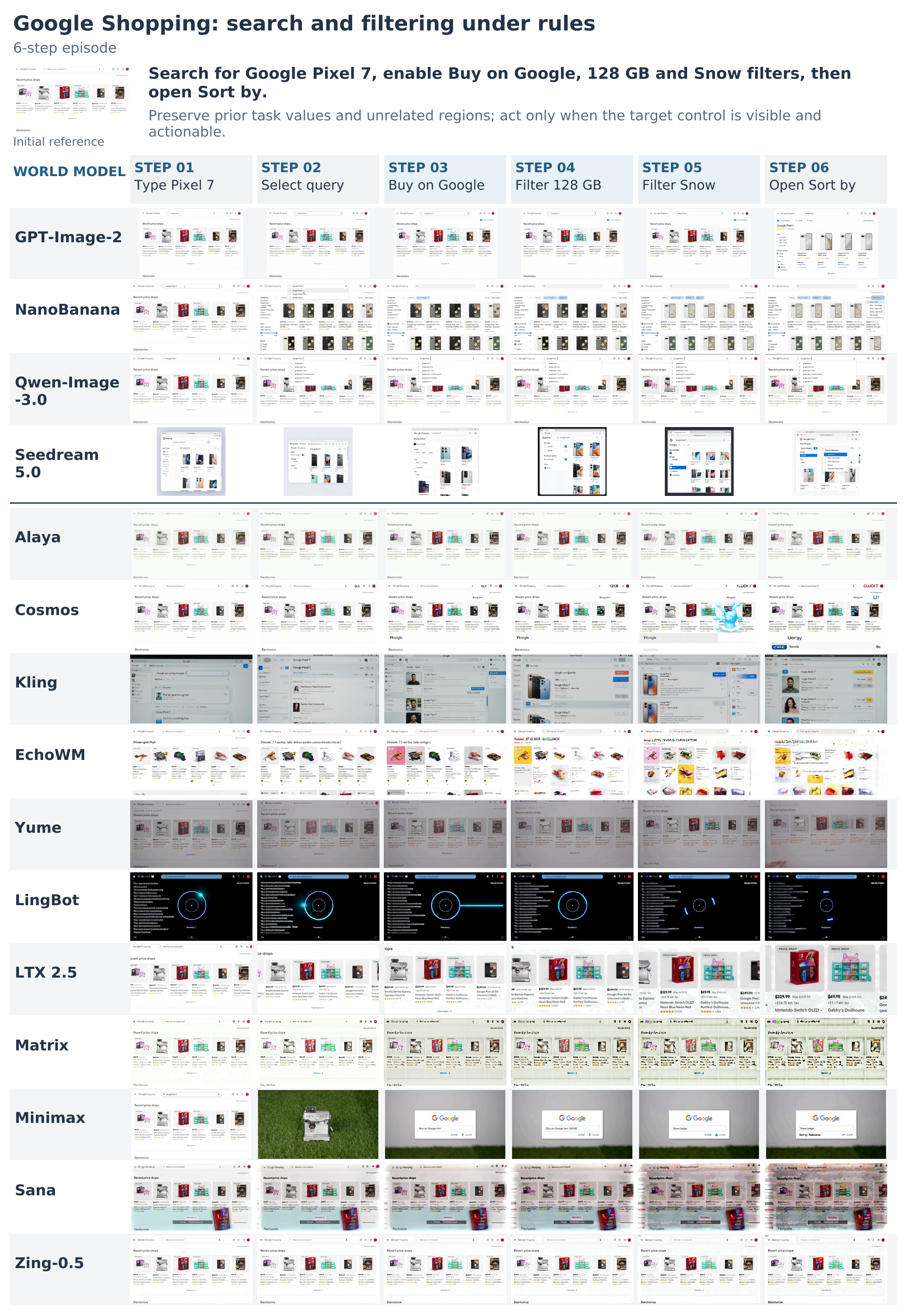}
\caption{\textbf{Google Shopping: persistent query and filter state.} Steps 1--6 cover query entry and selection, the Buy on Google, 128 GB, and Snow filters, and opening the sorting menu. Every operation must preserve previous selections and unaffected interface state.}
\label{fig:wm-visual-shopping}
\end{figure}

\subsection{Implications for Evaluation}

A valid WM response realizes the requested change while preserving unaffected entities, attributes, and earlier commitments. The cases distinguish three requirements: immediate action effects, persistence across the instructed history, and relevance to states encountered by a partner Agent. The controlled benchmark measures the first two; partner-specific utility requires matched interaction tests. Historical and modality differences remain material: static frames cannot establish timing, and selected examples cannot estimate error frequencies. Section~\ref{sec:comparison-limits} discusses the remaining validation and comparison limits.

\par

\section{Discussion}
\label{sec:discussion}

\subsection{Related Work}
\label{sec:related_work}

\textbf{Agent benchmarks.}
Agent evaluation spans general interactive reasoning and domain-specific task
execution. AgentBench tests reasoning and decision-making across eight
interactive environments~\citep{liu2024agentbench}. $\tau$-bench tests reliable
tool use and policy compliance~\citep{yao2024taubench}; $\tau^2$-Bench adds
coordination with users who also act in the shared
environment~\citep{barres2025tau2bench}. In digital environments,
WebArena evaluates task completion on functional websites, VisualWebArena adds
visually grounded web tasks, and OSWorld tests computer use across real desktop
applications with execution-based
verification~\citep{zhou2024webarena,koh2024visualwebarena,xie2024osworld}.
TheAgentCompany and Agents' Last Exam extend evaluation to workplace activities
and sustained professional workflows~\citep{xu2024agentcompany,sun2026agentslastexam},
whereas SkillsBench isolates the benefit of supplied procedural knowledge through
matched evaluations with and without curated skills~\citep{li2026skillsbench}.
Beyond digital work, ALFWorld connects abstract planning in text environments
with visually grounded household task execution~\citep{shridhar2020alfworld_tr}.
ScienceWorld tests scientific reasoning through interaction and experimentation
in a text environment~\citep{wang2022scienceworld_tr}, and SOTOPIA evaluates
social goal achievement through scenarios involving cooperation, competition,
and strategic communication~\citep{arxiv231011667}. These benchmarks cover
complementary aspects of Agent competence across physical, digital, social, and
scientific settings, although their observation modalities and evaluation
protocols differ. 

\textbf{World Model benchmarks.}
World Model evaluation covers predictive structure, physical plausibility, and
utility for downstream decisions. WorldPrediction tests whether models can
identify actions or action sequences connecting observed
states~\citep{chen2025worldprediction}, while
\citet{vafa2024implicitworld} assess recovery of underlying world structure
through automata-based diagnostics.
For video generation, WorldModelBench evaluates instruction following and
physical adherence~\citep{li2025worldmodelbench}; VideoPhy and VideoPhy-2 examine
physical commonsense in material interactions and diverse
activities~\citep{bansal2024videophy_tr,bansal2025videophy2_tr}; and PhyGenBench
uses prompts spanning 27 physical laws to assess physical
correctness~\citep{meng2025phygenbench}. These tests of generated videos provide
evidence about simulation fidelity, but do not by themselves establish reliable
interactive control. On the modeling side, GameGen-X generates open-world game
videos and controls their continuation through user inputs, including character
actions and scene changes~\citep{che2025gamegenx}.
WorldSimBench connects perceptual assessment with embodied
utility by evaluating whether generated videos can be translated into suitable
control signals~\citep{qin2024worldsimbench_tr}. WorldArena 2.0 extends embodied
evaluation to visual and tactile prediction, planning, and policy optimization
using World Models as interactive environments across simulated and real robotic
settings~\citep{shang2026worldarena2}. MIND evaluates memory consistency and
action control, including generalization across action spaces~\citep{ye2026mind}.
For text-based environments, \citet{huang2026behavior} evaluate and train World
Models using changes in a frozen reference Agent's next-action likelihood between
real and predicted states. These studies provide complementary measures of structural, physical, and temporal fidelity and downstream utility.

{\clubpenalty=10000
\textbf{Reciprocal learning.}
Dyna, learned World Models, and synthetic environments use simulated experience to improve planning and policies~\citep{sutton1990dyna,ha2018worldmodels,hafner2025dreamerv3,gu2025webdreamer,wang2026agentworldmodel}. Short model rollouts can limit the effect of model bias~\citep{janner2019mbpo}. In the other direction, directed exploration, policy rollouts, and interaction feedback support World Model or simulator updates~\citep{kim2020active,chebotar2019simopt,worldevolver,yu2026reinforcementworldmodel}. WebEvolver, GenEnv, and COMAP connect these directions within bounded task distributions~\citep{fang2025webevolver,guo2025genenv,liu2026comap}. Dual Exam complements this work with separate component assessments; it does not test a learning loop.\par}

\subsection{Implications for Coupled Evaluation}

The case analyses identify complementary discrepancies: Agents may reach plausible endpoints without satisfying route or evidence requirements, while World Models may realize an instruction but alter protected state. These observations come from different tasks, not matched Agent--World Model episodes. They motivate interface checks that verify the state an Agent acts on and the action effects a World Model generates.

\paragraph{Measure repair outcomes separately from responses.}
The pilot counts targeted corrective actions, not successful repairs. A repair claim requires subsequent evidence that the discrepancy was resolved while earlier valid state and task constraints were preserved. Retries, reflection text, or memory alone do not establish recovery.

\paragraph{Evaluate pairings on a specified task distribution.}
Agent ranks change under domain reweighting, and World Model preferences can reverse with the scenario mixture. Separate scores can guide candidate selection, but their sum or product has no validated interpretation as coupled success. Pairing studies should use matched goals, initialization, budgets, and failure accounting, with component controls and task-family-aware uncertainty estimates.

\paragraph{Test reciprocal improvement across training rounds.}
The present failures identify execution and state-consistency problems, not limits on learnability. In particular, the pilot's 0\% success rate applies to GPT-5.6 Sol with Seedance 2.5 under the stated protocol. A co-evolution study would compare joint updates with Agent-only, World-Model-only, and no-update controls on fresh held-out tasks across rounds and seeds, measuring both component performance and coupled task success. Neither a competence threshold nor sustained reciprocal improvement is established here.

\subsection{Scientific Evidence and Comparison Limits}
\label{sec:comparison-limits}
\paragraph{Scientific aggregation and evidence coverage.}
Scientific scores combine binary and graded outcomes under differing harnesses. Our Borden/DABIC clipping to $[0,100]$ is a Dual Exam convention, not official EdgeBench normalization or difficulty calibration. Edge aggregation accepts valid submissions within a 43,200-second allowance, including early-terminated runs. The four added configurations' Wireless endpoints were checked against best-submission judge reports. The truncated archive leaves other added-task endpoints and evaluator-version equivalence incompletely verified. Section~\ref{app:edge-aggregation} preserves raw objectives and evidence details; Section~\ref{app:agent-tools} describes specification and provenance limits.

\paragraph{Limits of the comparisons.}
The leaderboard reports descriptive point scores, without repeated-run uncertainty estimates or significance tests. Differences between configurations therefore do not establish statistically significant superiority. Original category means are available only to one decimal place, so derived aggregates retain this baseline precision limit. The separate tracks also do not establish how component scores predict coupled performance. Action-sensitive validation would benefit from controls such as unchanged-frame, no-op, and shuffled-intervention outputs, with scoring targets defined in advance for each control. Matched interaction experiments and repeated evaluations are needed to test robustness, partner suitability, and the effect of corrective feedback. These questions are distinct from the descriptive rankings reported here.

Reference environments use task-specific constraints and verifiers, but no unified independent error audit is available. Construction screening and expert review of specifications do not establish agreement on scores assigned to generated outputs. Independent WM scoring agreement and a full invalid-run audit are not reported, and the original Agent inventory lacks a consolidated per-model record of budgets, histories, retries, and run dates. Future uncertainty estimates must account for shared Social templates and SWE/ToolChoice parent links; domain-weight sensitivity alone is not an uncertainty estimate.

\section{Conclusion}
Dual Exam evaluates Multimodal Agents under fixed environment response mechanisms and World Models under fixed action interventions. The Agent track compares 10 configurations on 1,088 items in 15 reporting categories; GPT 6 Astra leads the case-weighted Index at 31.3. The World Model track covers 400 cases across eight scenarios; GPT-Image-2 leads full-coverage Overall at 21.7. These aggregates use different tasks and metrics and are not directly comparable. Scenario profiles and case evidence reveal discrepancies between local progress and verified completion, and between requested changes and preserved state. The results describe the evaluated configurations under the documented weighting, provenance, and precision limits. Predicting coupled success and testing reciprocal learning require matched interaction and training experiments.

\iftechnicalreport
\clearpage
\section{Authors}
\label{app:authors}
\begingroup
\noindent\reportauthors\par
\endgroup
\fi

\clearpage
\bibliographystyle{assets/fonts/meta/plainnat}
\bibliography{benchmark,benchmark_extended,benchmark_extended_tab3,benchmark_techreport}
\end{document}